\pdfoutput=1
\documentclass[11pt]{article}
\usepackage{acl}

\usepackage[aboveskip=1.5pt]{subcaption}
\usepackage[skip=1.5pt]{caption}
\usepackage{times}
\usepackage{latexsym}
\usepackage[T1]{fontenc}
\usepackage[utf8]{inputenc}
\usepackage{microtype}
\usepackage{inconsolata}
\usepackage{latexsym}
\usepackage{pgfgantt}
\usepackage{booktabs}
\usepackage{textcomp}
\usepackage{enumitem}
\usepackage{eucal}
\usepackage{amssymb}
\usepackage{amsmath}
\usepackage{multirow}
\usepackage{xcolor}
\usepackage[outline]{contour}
\usepackage{MnSymbol,bbding,pifont}
\definecolor{wic}{rgb}{0.78, 0.18, 0.297}
\newcommand{\wic}{\textcolor{black!10!wic}{\bf\texttt{WiC}\textsubscript{\includegraphics[width=.2cm]{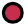}}}}

\usepackage{xfakebold}
\newcommand{\fbseries}{\unskip\setBold\aftergroup\unsetBold\aftergroup\ignorespaces}
\makeatletter
\newcommand{\setBoldness}[1]{\def\fake@bold{#1}}
\makeatother
\setBoldness{10}

\definecolor{rte}{rgb}{0.899, 0.341, 0.285}
\newcommand{\rte}{\textcolor{black!10!rte}{\bf\texttt{RTE}\textsubscript{\includegraphics[width=.2cm]{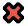}}}}
\definecolor{mrpc}{rgb}{0.97, 0.51, 0.30}
\newcommand{\mrpc}{\textcolor{black!10!mrpc}{\bf\texttt{MRPC}\textsubscript{\includegraphics[width=.2cm]{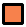}}}}
\definecolor{cola}{rgb}{0.97, 0.51, 0.30}
\newcommand{\cola}{\textcolor{black!10!cola}{\bf\texttt{CoLA}\textsubscript{\includegraphics[width=.2cm]{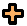}}}}
\definecolor{boolq}{rgb}{0.995, 0.85, 0.52}
\newcommand{\boolq}{\textcolor{black!10!boolq}{\bf\texttt{BoolQ}\textsubscript{\includegraphics[width=.2cm]{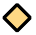}}}}
\definecolor{sstt}{rgb}{0.999, 0.955, 0.67}
\newcommand{\sstt}{\textcolor{black!25!sstt}{\bf\texttt{SST-2}\textsubscript{\includegraphics[width=.2cm]{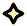}}}}
\definecolor{sstf}{rgb}{0.963, 0.985, 0.69}
\newcommand{\sstf}{\textcolor{black!25!sstf}{\bf\texttt{SST-5}\textsubscript{\includegraphics[width=.2cm]{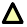}}}}
\definecolor{emotion}{rgb}{0.866, 0.946, 0.603}
\newcommand{\emotion}{\textcolor{black!25!emotion}{\bf\texttt{Emotion}\textsubscript{\includegraphics[width=.2cm]{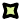}}}}
\definecolor{ih}{rgb}{0.684, 0.872, 0.64}
\newcommand{\ih}{\textcolor{black!10!ih}{\bf\texttt{ImplicitHate}\textsubscript{\includegraphics[width=.2cm]{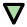}}}}
\definecolor{stormfront}{rgb}{0.485, 0.794, 0.646}
\newcommand{\stormfront}{\textcolor{black!10!stormfront}{\bf\texttt{Stormfront}\textsubscript{\includegraphics[width=.2cm]{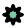}}}}
\usepackage{wasysym}
\definecolor{reuters}{rgb}{0.304, 0.654, 0.69}
\newcommand{\reuters}{\textcolor{black!10!reuters}{\bf\texttt{Reuters}\textsubscript{\includegraphics[width=.2cm]{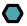}}}}
\definecolor{trec}{rgb}{0.24, 0.476, 0.71}
\newcommand{\trec}{\textcolor{black!10!trec}{\bf\texttt{TREC}\textsubscript{\includegraphics[width=.2cm]{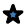}}}}

\newcommand{\wicb}{\texttt{WiC}}
\newcommand{\rteb}{\texttt{RTE}}
\newcommand{\mrpcb}{\texttt{MRPC}}

\newcommand{\ssttb}{\texttt{SST-2}}
\newcommand{\sstfb}{\texttt{SST-5}}
\newcommand{\emotionb}{\texttt{Emotion}}
\newcommand{\ihb}{\texttt{ImplicitHate}}
\newcommand{\stormfrontb}{\texttt{Stormfront}}
\newcommand{\reutersb}{\texttt{Reuters}}
\newcommand{\trecb}{\texttt{TREC}}

\title{Generalisation First, Memorisation Second? Memorisation \\ Localisation for Natural Language Classification Tasks}
\author{Verna Dankers$^1$ \textnormal{and} Ivan Titov$^{1,2}$ \\
$^1$ILCC, University of Edinburgh \\
$^2$ILLC, University of Amsterdam \\
\texttt{vernadankers@gmail.com}, \texttt{ititov@inf.ed.ac.uk}}
\begin{document}
\maketitle

\begin{abstract}
Memorisation is a natural part of learning from real-world data: neural models pick up on atypical input-output combinations and store those training examples in their parameter space. \textit{That} this happens is well-known, but \textit{how} and \textit{where} are questions that remain largely unanswered. Given a multi-layered neural model, where does memorisation occur in the millions of parameters?
Related work reports conflicting findings: a dominant hypothesis based on image classification is that lower layers learn generalisable features and that deeper layers specialise and memorise. Work from NLP suggests this does not apply to language models, but has been mainly focused on memorisation of facts.
We expand the scope of the localisation question to 12 natural language classification tasks and apply 4 memorisation localisation techniques.
Our results indicate that memorisation is a gradual process rather than a localised one, establish that memorisation is task-dependent, and give nuance to the generalisation first, memorisation second hypothesis.
\end{abstract}

\section{Introduction}
\label{sec:introduction}

Memorisation in neural models is both concerning due to overfitting and privacy concerns, and desired because of information that needs to be stored, such as facts.
With the recent surge in the number of models trained on very large, closed-access training corpora, the NLP community has seen an increased interest in research questions that aim to improve our understanding of memorisation, such as:
Which data points are memorised, and can we extract those examples from models \citep[e.g.][]{chang2023speak,shi2023detecting,nasr2023scalable}? How dependent is memorisation on model scale, architecture and training procedures \citep[e.g.][]{carlini2022quantifying}? Can we localise memorised information (i.e.\ pinpoint which weights, subcomponents or layers are most associated with storing that information) and edit models' memories \citep[e.g.][]{meng2022locating,hase2023does}?

\begin{figure}
    \centering\small
    \includegraphics[width=\columnwidth]{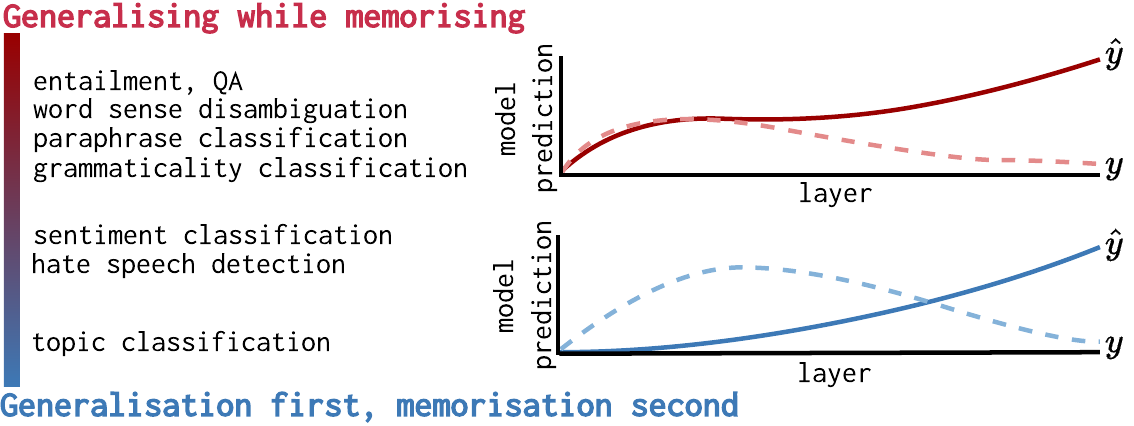}
    \caption{If we train transformer to memorise incorrect label $\hat{y}$, the implementation of that memorisation is task-dependent. We demonstrate this for 12 NLP classification tasks. The visualisation is for illustrative purposes.}
    \label{fig:introduction_illustration}
    \vspace{-0.5cm}
\end{figure}

Memorisation localisation is the central theme in this work, specifically localisation at the level of layers.
There is a lack of consensus about which layers are particularly involved in memorisation in deep neural models, which may stem from the varying experimental setups and varying definitions of memorisation employed by different studies.
Work from \textit{computer vision} (CV) mostly focused on memorisation of perfectly memorised mislabelled examples, positing that lower layers capture generalisable features while deeper layers memorise \citep[e.g.][]{baldock2021deep,stephenson2021geometry} (although that has been recently challenged by \citet{maini2023can}).
Work from NLP often discusses memorisation of facts, for which lower \citep{geva2023dissecting}, middle \citep{meng2022locating} and final layers \citep{dai2022knowledge} have all been mentioned as playing crucial roles in \textit{pre-trained language models} (PLMs).
And lately, with the availability of some open-access pre-training data, memorisation localisation for sequences memorised verbatim has gained traction, and initial results primarily point to lower layers \citep{stoehr2024localizing}.
Different articles investigating different types of memorisation arrive at different answers, and even work focused on a specific type (such as facts) does not always agree.

We contribute an important piece of the puzzle in the memorisation localisation landscape, by employing a setup that is similar to \citeauthor{stephenson2021geometry}\ and \citeauthor{maini2023can}: we perform layer-wise localisation in fine-tuned models for 12 NLP classification tasks, enforcing memorisation by applying label perturbation to a data subset.
We use four memorisation localisation methods (\S\ref{sec:methods}), and first examine their accuracy in a control setup.
Afterwards (\S\ref{sec:results}), we address the main research question: \textbf{In which layers does memorisation occur?} The results do not always align -- which underscores that we should not overly rely on one localisation method -- but do tell a coherent story. In \S\ref{sec:centroid_analysis}, we introduce a visualisation technique (centroid analysis) to make this story more interpretable: memorisation is not sudden but gradual.
Together, layers gradually shift mislabelled examples to their newly assigned class, but \textit{when} that happens is task-dependent: the better a model generalises to new data for a particular task, the more relevant deeper layers are for memorisation. Figure~\ref{fig:introduction_illustration} illustrates this.
Our findings beg a nuance of the generalisation first, memorisation second hypothesis, and we end with a discussion (\S\ref{sec:discussion}) of what our findings mean for localisation and model editing going forward. 
\section{Related work}
\label{sec:related_work}
In this section, we discuss a broad range of related work on memorisation localisation from CV and NLP, focusing specifically on studies that discuss the role of different \textit{layers} in deep neural networks.

\paragraph{Noise memorisation in CV}
Multiple image classification studies concluded that deeper layers are more involved in memorisation than earlier layers, based on analyses of the memorisation of mislabelled or hard examples.
This has been reported by work contrasting entire models (regular models and ones trained with randomised labels) \citep[][i.a.]{morcos2018insights,cohen2018dnn,ansuini2019intrinsic},
and work discussing how some examples are handled differently within one model: \citet{baldock2021deep} establish a positive correlation between prediction depth in image classification (the earliest layer that predicts the label) and example-level learning difficulty metrics.
\citet{stephenson2021geometry} analyse hidden representations for image classification and report that memorisation of mislabelled examples occurs abruptly in late layers and late training epochs. Rewinding models' final layers to earlier checkpoints reversed memorisation.

Contrary to previous work, \citet{maini2023can} report that for image classification datasets that were partially mislabelled, instead of memorisation being confined to specific layers, there are small sets of neurons dispersed over the full architecture involved in memorisation.

\paragraph{Memorisation of factual knowledge}
NLP memorisation localisation studies have primarily focused on factual knowledge, although only a subset of work in this direction discusses the roles of different layers.
\citet{de2021editing} first connected work from CV to fact memorisation in transformer, and train a hypernetwork to edit facts. Their hypernetwork mostly edits the bottom layer of a six-layer transformer, and \citeauthor{de2021editing}\ suggest this difference might be due to the change in modality. 

Later work operated under the assumption of the \textit{knowledge neuron thesis},\footnote{The term was coined by \citet{Niu2024WhatDT} to summarise the hypothesis underlying multiple related studies. \citeauthor{Niu2024WhatDT} criticise the thesis since it oversimplifies knowledge storage. Instead, they suggest to focus on network-wide circuits.} assuming that facts are recalled from the training corpus through transformer's MLP weights, that act as a key-value memory \citep{geva2021transformer}, and that one may thus be able to identify MLP knowledge neurons \citep{dai2022knowledge}:
(1) \citet{meng2022locating,meng2022mass} edit factual memories in transformer-based PLMs' MLPs by first localising memorisation to specific layers and only updating those layers, in which case
early/mid-layers are most often selected.\footnote{E.g.\ generalisation to paraphrased prompts of modified facts peaks when editing layer 18 out of 48 for \texttt{GPT-2 XL}. This result is not specific to transformer: \citet{sharma2024locating} find early/mid-layers to be important when editing facts in \texttt{Mamba}.}
Note that \citet{hase2023does} find success in model editing to be unrelated to the layers selected by \citeauthor{meng2022locating}'s localisation method, which means that model editing might be an unreliable way to check where facts are stored.
(2) \citet{dai2022knowledge} identify knowledge neurons for factual information, and mostly find neurons in the \textit{top} layers of \texttt{BERT}, with similar findings being reported in later work on knowledge neurons by \citet{zhao2024tracing} and \citet{chen2024journey}.
\citeauthor{dai2022knowledge} then succesfully use those neurons to update facts.

Although the articles above disagree in terms of whether higher or lower layers store factual information, work beyond model editing and knowledge neurons indicates factual information is already present in the lower layers:
\citet{haviv2023understanding} show that facts and idioms are stored and retrieved in early layers in \texttt{BERT} and \texttt{GPT-2}. Deeper layers perform confidence boosting \textit{after} retrieval.
\citet{geva2023dissecting} artificially block parts of the computation in fact retrieval for \texttt{GPT-J} and \texttt{GPT-2 XL}, and find that factual information is stored in the lower MLP sublayers, which is echoed by recent work from \citet{ortu2024competition}.

\paragraph{Verbatim memorisation}
While memorisation localisation has predominantly focused on facts, recent open science initiatives publishing PLM pre-training corpora enable research on memorisation of pre-training data.
\citet{chang2023localization} present a dataset to evaluate different localisation methods by relying on text memorised verbatim by three PLMs, but they do not elaborate on the roles of layers.
\citet{stoehr2024localizing} examine for one specific 12-layer architecture (\texttt{GPT-Neo-125M}, that we will also analyse) which model components are responsible for memorising sequences of 50 tokens verbatim, finding lower layers to be the most relevant.
Localisation of verbatim memorisation is hard to standardise across different tasks and models due to the required access to pre-training data, and because all PLMs memorise different sequences.

\paragraph{Memorisation beyond localisation}
Other work on memorisation in NLP examines the conditions under which memorisation occurs during fine-tuning \citep{tanzer2022memorisation,mireshghallah2022empirical}, which fine-tuning tasks lead to the most memorisation \citep{zeng2023exploring}, which examples are memorised \citep{biderman2024emergent} and when memorisation is beneficial for generalisation \citep{zhang2023counterfactual,zheng2022empirical}.

\vspace{0.2cm}
\noindent 
Summarising, many different conclusions have been drawn in memorisation localisation studies. 
It is unclear whether the different conclusions from different articles may be reduced to a difference between the vision and language modalities, to a difference between the different types of memorisation investigated, to localisation techniques employed or even to the varying models and model sizes used in related work.\footnote{For instance, \citeauthor{de2021editing} investigate 6-layer transformers, whereas \citeauthor{dai2022knowledge} and \citeauthor{stoehr2024localizing} use 12-layered networks, and \citeauthor{meng2022locating} mostly focus on \texttt{GPT-2 XL} with 48 layers.}
We take away some of that confusion by (1) employing a setup previously used for the vision modality and (2) investigating a type of memorisation (`noise memorisation') understudied in NLP, using widely studied models.
This allows us to conclude whether the `deeper layers' answer from CV really stands in contrast with the `lower layers' answer from the majority of NLP studies (or whether that was simply unique to noise memorisation), and allows us to investigate whether the `lower layers' answer is unique to fact memorisation and verbatim memorisation.

\begin{table*}[t]
    \setlength{\tabcolsep}{1.5pt}
    \centering\small
    \resizebox{\textwidth}{!}{\begin{tabular}{llllcc}
    \toprule
    \textbf{Category} & \textbf{Dataset} & \textbf{Task} & \textbf{Domain} & \textbf{Size} & \textbf{Labels} \\\midrule
    \multirow{5}{*}{NLU} & \wic\ by \citet{pilehvar2019wic}         & word sense disambiguation & Word-/Verb-net, Wiktionary & 5.4k & 2 \\ 
    &\rte\ by \citet{rte1}         & textual entailment & news, Wikipedia & 2.5k & 2\\
    &\mrpc\ by \citet{dolan2005automatically}        & paraphrase classification & news & 3.7k & 2 \\
    &\cola\ by \citet{warstadt2019neural}        & labelling grammaticality & linguistic theory books & 8.5k & 2 \\
    &\boolq\ by \citet{clark2019boolq}       & question answering given a context & Google queries, Wikipedia & 9.4k & 2 \\\midrule
    \multirow{3}{*}{Sentiment} &\sstt\ by \citet{socher2013recursive}        & sentiment classification & movie reviews & 6.9k & 2 \\
    &\sstf\ by \citet{socher2013recursive}        & sentiment classification & movie reviews & 8.5k & 5\\
    &\emotion\ by \citet{saravia2018carer}     & emotion classification & tweets & 16k & 6 \\\midrule
    \multirow{2}{*}{Hate speech} &\ih\ by \citet{elsherief2021latent}          & hate speech classification & tweets & 5.1k & 7\\
    &\stormfront\ by \citet{de2018hate}  & hate speech classification & social media & 8.6k & 2\\\midrule
    Topic&\reuters\ by \citet{apte1994towards}     & topic classification & news & 5k & 8\\
    Classification&\trec\ by \citet{li-roth-2002-learning,hovy-etal-2001-toward}        & topic classification & news, misc. & 5.5k & 6 \\
    \midrule
    \end{tabular}}
    \caption{Datasets with their domain, label set size and training set size. In \S\ref{sec:results} and \S\ref{sec:centroid_analysis}, datasets are marked consistently using the same colours and symbols.}
    \label{tab:datasets}
    \vspace{-0.4cm}
\end{table*}

\section{Methods}
\label{sec:methods}
To gain a good understanding of how memorisation is task-dependent, we combine binary classification tasks from (Super)GLUE \citep{wang2018glue,wang2019superglue} with tasks from more diverse domains and label set sizes.
The tasks fall into four categories: generic \textit{natural language understanding} (NLU), sentiment-related tasks, hate speech detection and topic classification.
Table~\ref{tab:datasets} enumerates the tasks, the datasets' domains, and the training set and label set sizes.
For each dataset, we perturb the labels of 15\% of the training examples (`noisy' examples, $x\in\mathcal{X}_n$, $y\in\mathcal{Y}_n$), with the new label randomly drawn from all labels but the original one. The remaining 85\% is unperturbed (`clean' examples, $x\in\mathcal{X}_c$, $y\in\mathcal{Y}_c$).

We analyse four PLMs: \texttt{BERT-base} \citep{devlin-etal-2019-bert}, \texttt{OPT-125m} \citep{zhang2022opt}, \texttt{Pythia-160m} \citep{biderman2023pythia} and \texttt{GPT-Neo-125m} \citep{gpt-neo,gao2020pile}.
We fine-tune each model separately for the 12 tasks.
Appendix~\ref{ap:setup} describes the models and our technical setup.\footnote{The codebase is available at: \url{https://github.com/vernadankers/memorisation_localisation}.} These 4 architectures are similar in size and have 12 layers each. In Appendix~\ref{ap:opt_big}, we repeat a subset of the experiments with \texttt{OPT-1.3B}.

The PLMs ($\theta_{P}$) are fine-tuned for 50 epochs, and checkpoints are stored when the training accuracy is near-ceiling ($\theta_{M_1}$), and at the end of training ($\theta_{M_2}$).
We also train models using the original labels ($\theta_O$), using the same random seeds as $\theta_{M_1}$ and $\theta_{M_2}$.
During fine-tuning, we freeze the input embeddings.
Results reported in \S\ref{sec:control_setup} are based on one fine-tuning seed, and the remainder of the main paper computes results using three seeds. Seeds control the data order and classification heads.

\subsection{Localisation techniques}
\label{subsec:techniques}
We apply four localisation methods that are detailed in this subsection and further evaluated in \S\ref{sec:control_setup}.

\paragraph{Layer retraining and layer swapping}
First, we perform layer retraining, similar to \citet{maini2023can}.
We reset layers of interest using weights from $\theta_{P}$, freeze the remaining layers using weights from $\theta_{M_2}$,
and retrain using clean examples for five epochs.
If the resulting model maintains its performance on noisy data, the retrained layers are redundant in terms of memorisation.
If the performance on noisy data decreases, that does not guarantee that memorisation can be localised to the retrained layers since the retraining objective may have multiple minima, of which only some maintain the memorisation performance.
We retrain consecutive layers of window sizes ranging from 1 to 12.

Alternatively, we swap layers between $\theta_{M_2}$ and $\theta_{O}$, using the same window sizes. If swapping layers leads to a drop in performance on noisy examples while maintaining performance on clean ones, it becomes more likely that the layers were vital for memorisation (although this is again not guaranteed).
We indicate layer relevance using the \textbf{memorisation error}: the ratio of incorrect predictions for noisy examples.
The lower the error rate for noisy examples when retraining or swapping a layer, the less likely it is that this layer was crucial for memorisation.

Retraining or swapping all 12 layers means modifying the full model, and provides a baseline for the maximum error we can expect for the noisy data. In the results section, we will use this to normalise the results, such that the memorisation error is 1.0 when modifying all 12 layers.

\paragraph{Forgetting gradients}
We also inspect gradients, computed by back-propagating $-\mathcal{L}(\mathcal{X}_n, \mathcal{Y}_n, \theta_{M_1})$
and computing the $L_1$-norm per layer. We use $\theta_{M_1}$ due to gradient saturation in $\theta_{M_2}$.\footnote{See \citet{akyurek2022tracing} for a discussion of issues with gradient-based methods when tracing knowledge in a model.}
The assumption is that memorisation is localised in the layers requiring the largest updates when `forgetting' noisy labels.
Because gradient magnitudes do not reliably pinpoint layers, we used two tasks to decide on the norm to use and whether or not to normalise gradients using gradients for clean examples and gradients for a frozen model (Appendix~\ref{ap:hypestimation}).

\paragraph{Probing}
Lastly, we train probing classifiers \citep{conneau-etal-2018-cram} to predict whether, for a hidden state encoding $x$ in layer $l$ ($h^x_l$), $x\in\mathcal{X}_n$ or $x\in\mathcal{X}_c$.
The classifier is an MLP with one hidden layer, trained for 100 epochs maximum with a learning rate of $2e{-}4$. The hidden states come from \textit{training} examples that are redistributed into a training, validation and test set for the probe.
The classifiers are trained separately per layer, using five random seeds per layer.
We extract the $F_1$-score on the test partitions and use the increase from $l{-}1$ to $l$ as an indication of $l$'s involvement in memorisation (except for layer 1, which we compare to the $F_1$-score from a probe trained on $\theta_P$).

\begin{figure}[t]
    \includegraphics[height=2.35cm]{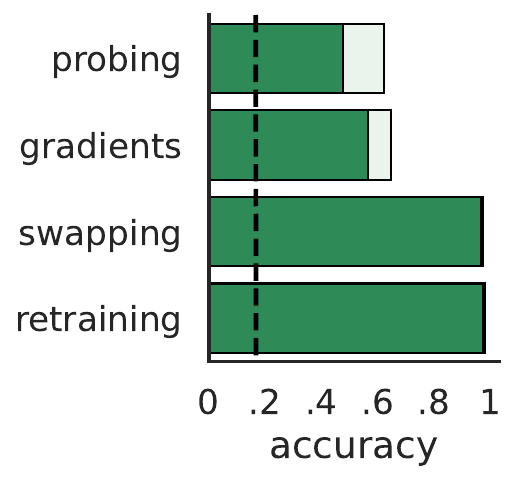}
    \includegraphics[height=2.35cm]{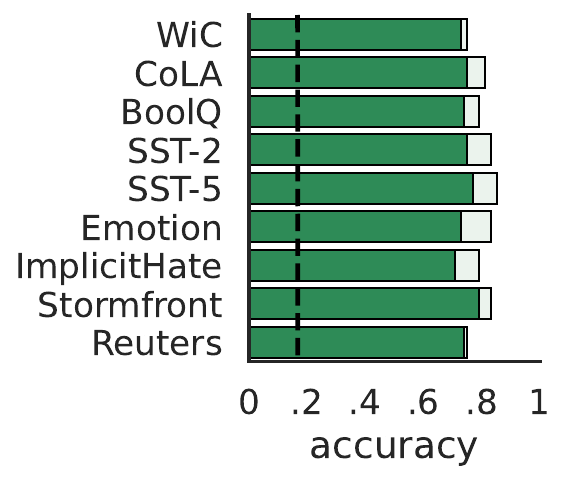}
    \includegraphics[height=2.35cm]{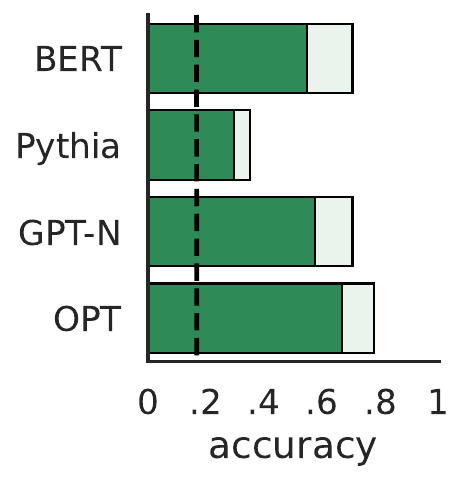}
    \caption{Control setup accuracy@1 (light) and accuracy@2 (dark) per localisation method (left), dataset (middle) or model (computed using probing and gradients, right), and a random guessing baseline (dashed).}
    \label{fig:control_setup}
    \vspace{-0.4cm}
\end{figure}

\subsection{Control setup: does localisation succeed?}
\label{sec:control_setup}
We now evaluate the localisation techniques by enforcing memorisation in prespecified layers and examining whether the techniques pinpoint those layers (i.e.\ whether localisation succeeds).

\paragraph{Experimental setup}
We approach this as a multitask learning setup, to ensure all layers are fine-tuned, but only two are modified by the task with noisy labels: the entire model is fine-tuned using \rteb, while the remaining task can only modify two layers at a time (layers 1 and 2, 6 and 7 or 11 and 12).
We train the model separately for the remaining 11 tasks and these 3 different choices of layer combinations.
Afterwards, we first use \mrpcb\ and \trecb\ to validate the postprocessing steps for the forgetting gradients (see Appendix~\ref{ap:hypestimation}), after which all localisation techniques were applied to the remaining nine tasks.
We evaluate the techniques using \textbf{accuracy@k}, indicating the percentage of the $k$ highest-scoring layers that were among the correct ones for that setup, computed for $k\in\{1, 2\}$.

\paragraph{Results}
Figure~\ref{fig:control_setup} (left) summarises the accuracies per localisation technique.
Swapping and retraining are very accurate, but gradients and probing are less reliable, with accuracy@1 just over 60\%.
Note that the near-perfect accuracy for retraining and swapping here does not guarantee perfect accuracy in the uncontrolled setup; the per-layer freezing is just very well-aligned with the per-layer approach of those techniques.
The accuracy per dataset (Figure~\ref{fig:control_setup}, middle) only shows slight variations.
For the two lowest-scoring localisation techniques (probing, gradients), Figure~\ref{fig:control_setup} (right) details the accuracies per model.
\texttt{Pythia} scores particularly badly for the gradient analysis, for which the accuracies barely exceed the baseline.
Postprocessing (Appendix~\ref{ap:hypestimation}) did not help, which underscores gradients' unreliability.
\begin{figure}[t]
\begin{subfigure}[b]{0.49\columnwidth}
    \centering
    \includegraphics[width=\textwidth]{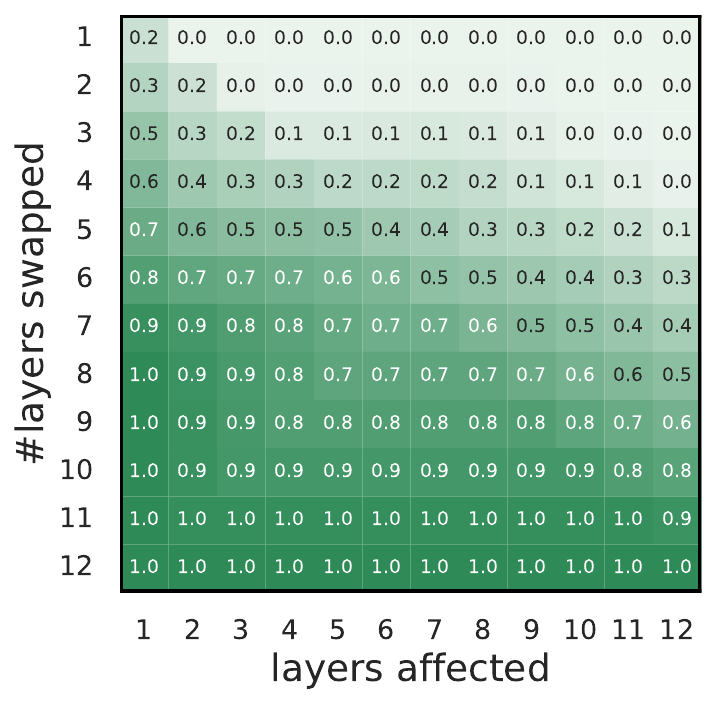}
    \caption{\texttt{RTE}, swapping}
\end{subfigure}
\begin{subfigure}[b]{0.49\columnwidth}
    \centering
    \includegraphics[width=\textwidth]{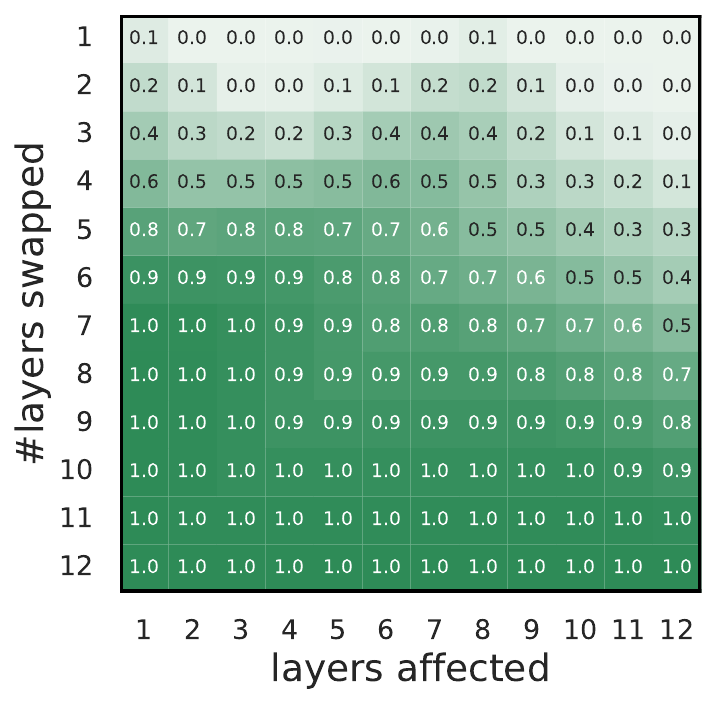}
    \caption{\texttt{SST-2}, swapping}
\end{subfigure}
\begin{subfigure}[b]{0.49\columnwidth}
    \centering
    \includegraphics[width=\textwidth]{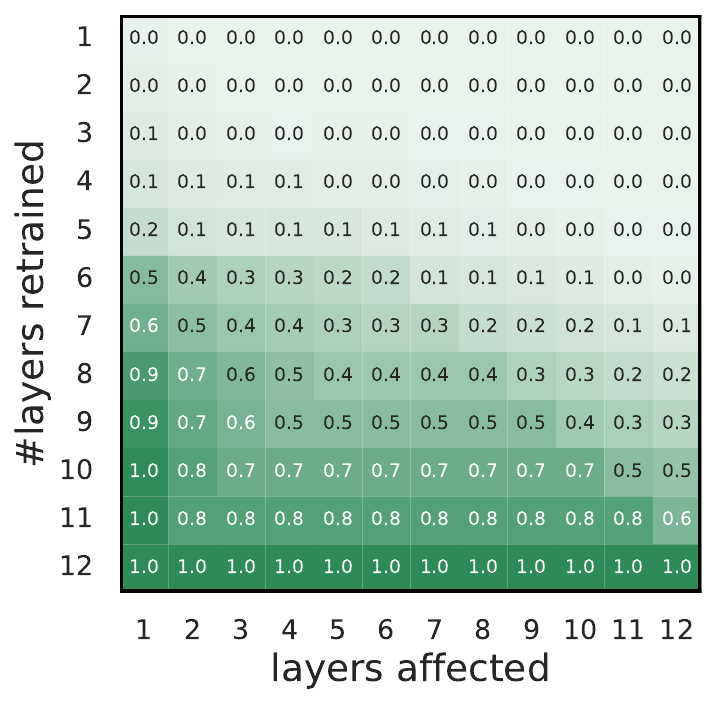}
    \caption{\texttt{RTE}, retraining}
\end{subfigure}\hfill\begin{subfigure}[b]{0.49\columnwidth}
    \centering
    \includegraphics[width=\textwidth]{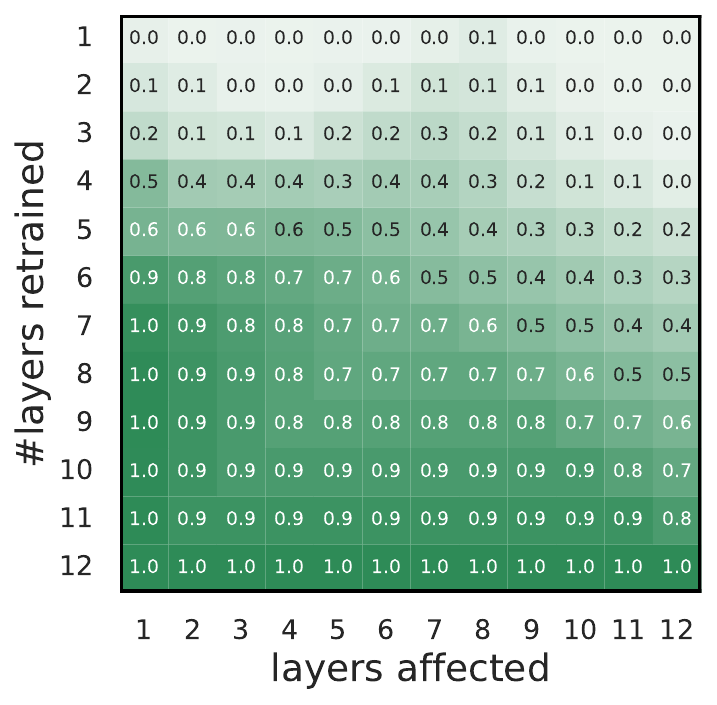}
    \caption{\texttt{SST-2}, retraining}
\end{subfigure}
\caption{Memorisation error for layer swapping and retraining for two datasets, for the \texttt{OPT} model.}
\label{fig:swapping_results_full}
\vspace{-0.5cm}
\end{figure}

\section{Results for memorisation localisation}
\label{sec:results}

We now apply the localisation techniques to models in which all layers have been fine-tuned for one task at a time. The results indicate how important each layer is for memorisation, per dataset, per model. We cannot simply aggregate over all results (12 layers $\times$ 12 datasets $\times$ 4 localisation techniques $\times$ 4 models), because the absolute scores returned by different techniques are not directly comparable. We discuss the results per localisation technique.

\begin{figure}[t]
\begin{subfigure}[b]{0.49\columnwidth}
    \centering
    \includegraphics[width=\textwidth]{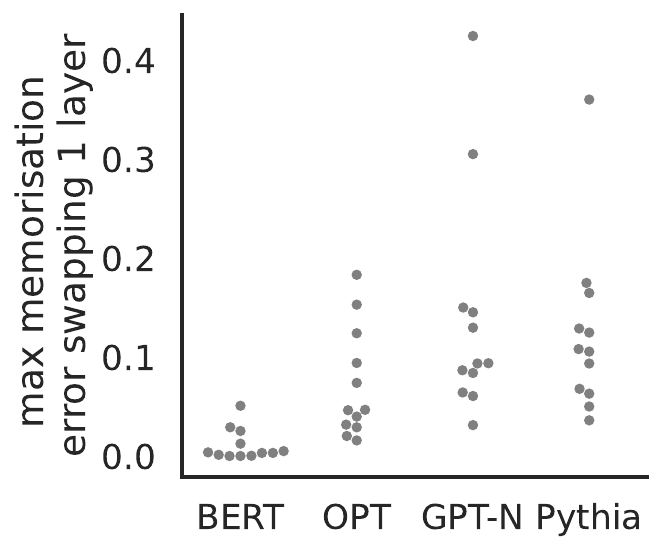}
    \caption{Layer swapping}
\end{subfigure}
\begin{subfigure}[b]{0.49\columnwidth}
    \centering
    \includegraphics[width=\textwidth]{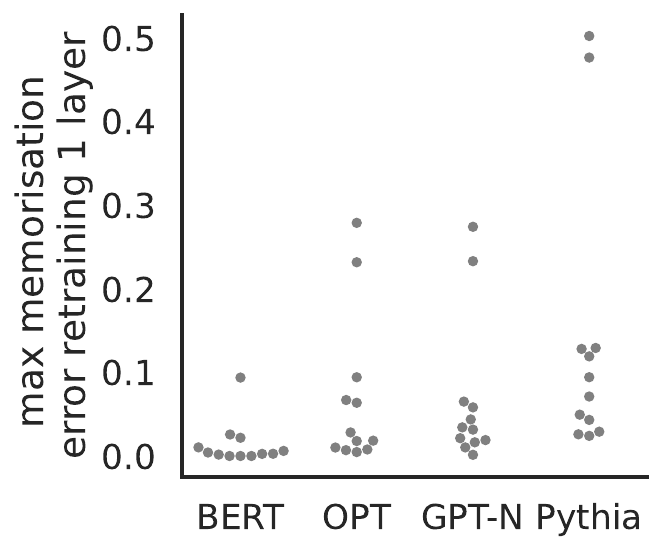}
    \caption{Layer retraining}
\end{subfigure}
\caption{Maximum memorisation error over 12 layers when modifying 1 layer; dots represent datasets.
Jitter along the $x$-axis was added to improve visibility.}
\label{fig:one_layer}
\vspace{-0.5cm}
\end{figure}

\subsection{Layer swapping and retraining}
\label{sec:swapping_retraining}
When swapping or retraining layers, we gradually modify more and more layers in $\theta_{M_2}$, either using weights from $\theta_O$, or by retraining layers using clean examples.\footnote{When swapping layers, we monitor errors on clean examples to ensure that the mixture of models $\theta_O$ and $\theta_{M_2}$ differs only in terms of predictions for noisy examples. The mean error for clean examples over all windows was 0.3\%.}
We modify 1 to 12 layers at a time, and measure the effect via the memorisation error.

\begin{figure*}

\begin{subfigure}[b]{0.7\textwidth}
    \includegraphics[width=\textwidth]{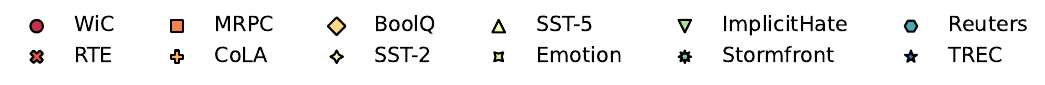}
\end{subfigure}
\begin{subfigure}[b]{\textwidth}
    \includegraphics[width=.285\textwidth]{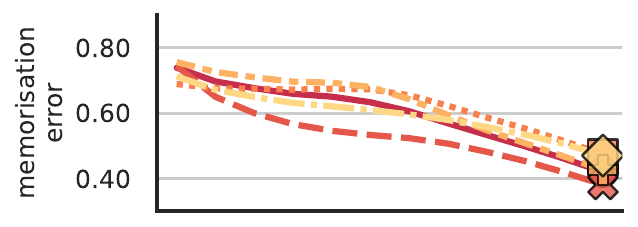}
    \includegraphics[width=.23\textwidth]{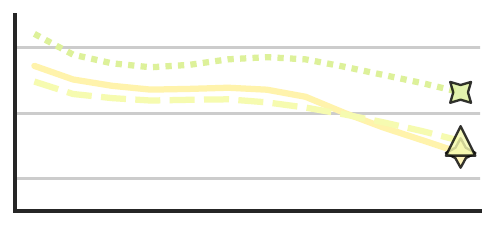}
    \includegraphics[width=.23\textwidth]{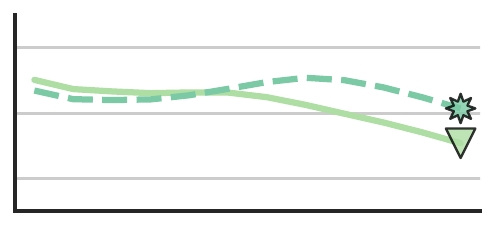}
    \includegraphics[width=.23\textwidth]{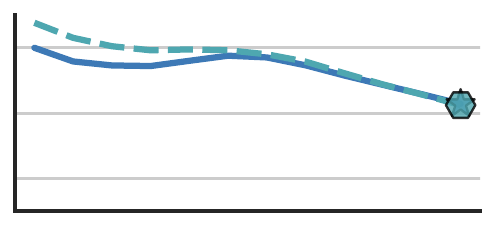}

    \includegraphics[width=.285\textwidth]{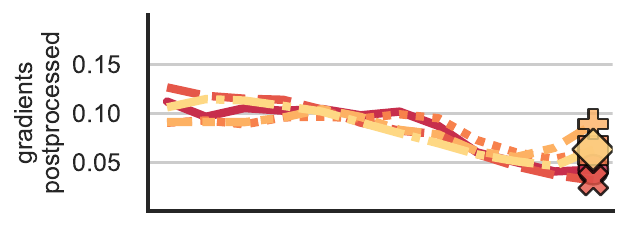}
    \includegraphics[width=.23\textwidth]{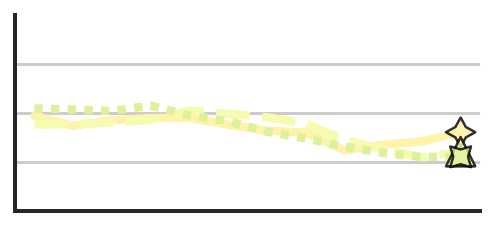}
    \includegraphics[width=.23\textwidth]{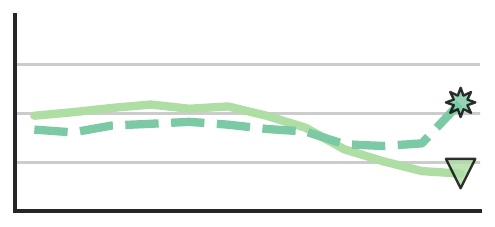}
    \includegraphics[width=.23\textwidth]{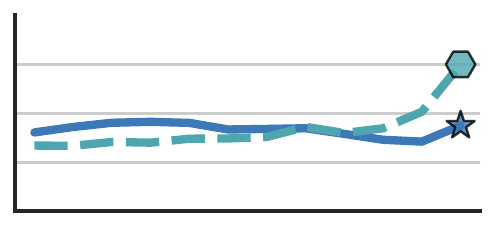}

    \hspace{0.15cm}\begin{subfigure}[b]{0.275\textwidth}
        \includegraphics[width=\textwidth]{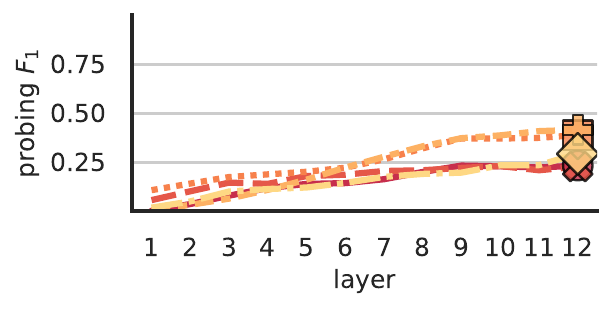}
    \caption{NLU tasks}
    \end{subfigure}
    \begin{subfigure}[b]{0.23\textwidth}
        \includegraphics[width=\textwidth]{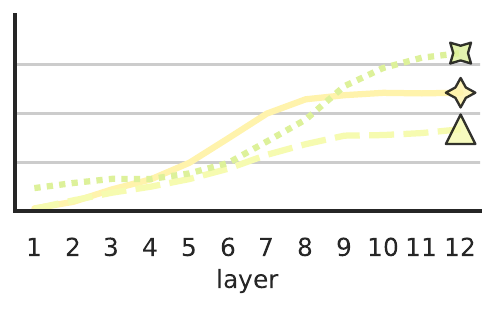}
    \caption{Sentiment tasks}
    \end{subfigure}
    \begin{subfigure}[b]{0.23\textwidth}
        \includegraphics[width=\textwidth]{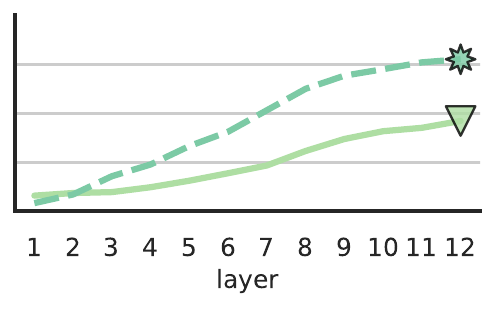}
    \caption{Hate speech tasks}
    \end{subfigure}
    \begin{subfigure}[b]{0.23\textwidth}
        \includegraphics[width=\textwidth]{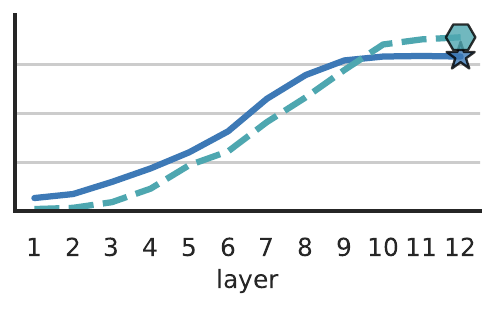}
    \caption{Topic classification}
    \end{subfigure}
\end{subfigure}

\caption{Memorisation localisation for \texttt{OPT}: (1) layer swapping error rates, higher numbers suggest higher relevance. (2) gradient norms, higher numbers suggest higher relevance. (3) probing $F_1$-scores when training probes to predict whether an example is noisy. The increase between layers suggests layer relevance.}
\vspace{-0.5cm}
\label{fig:taskxtechniques}
\end{figure*}

\begin{figure}[t]
    \centering
    \begin{subfigure}[b]{0.55\columnwidth}
        \includegraphics[width=\textwidth]{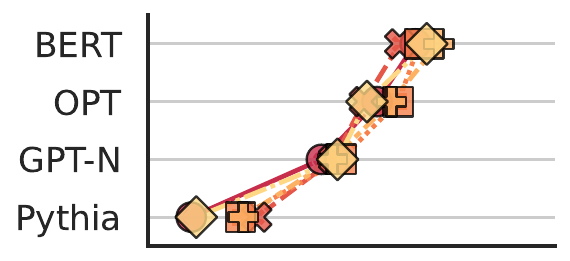}
        \caption{NLU tasks}
    \end{subfigure}
    \begin{subfigure}[b]{0.42\columnwidth}
        \includegraphics[width=\textwidth]{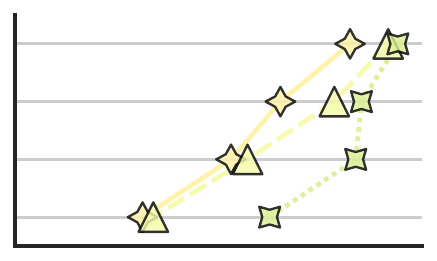}
        \caption{Sentiment tasks}
    \end{subfigure}
    \begin{subfigure}[b]{0.55\columnwidth}
        \includegraphics[width=\textwidth]{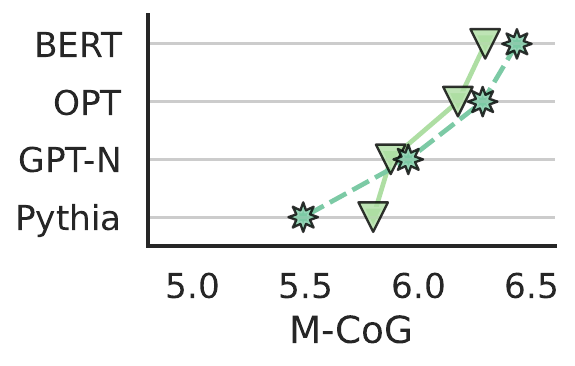}
        \caption{Hate speech tasks}
    \end{subfigure}
    \begin{subfigure}[b]{0.42\columnwidth}
        \includegraphics[width=\textwidth]{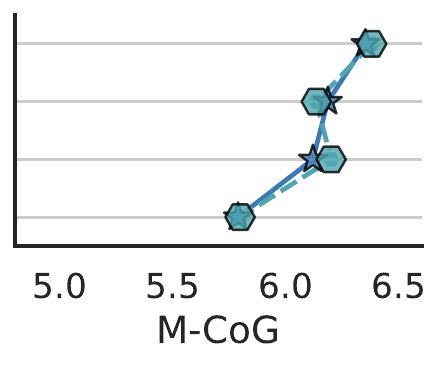}
        \caption{Topic classification}
    \end{subfigure}
    \caption{M-CoG coefficients for layer retraining, that give a coarse indication of whether lower or higher layers matter more for memorisation.}
    \label{fig:retraining_cog}
    \vspace{-0.5cm}
\end{figure}

\paragraph{Case study: \rteb\ vs \ssttb} Before discussing trends across all 12 datasets, we inspect two specific sets of results to gain a deeper understanding of the data.
Figure~\ref{fig:swapping_results_full} details memorisation error rates for \rteb\ and \ssttb\ (for \texttt{OPT}): in these matrices, value $z$ in row $x$, column $y$, indicates that for all layer combinations of $x$ consecutive layers including $y$, $z$ was the mean error rate. We show the results separately for swapping and retraining.

What commonalities and differences do we observe? For both datasets, modifying a few layers only yields low error rates (see the top few light green rows), and fully reverting memorisation requires modifying 7 to 10 layers. Memorisation is thus not limited to a few layers, but, instead, dispersed over the model.
Despite these similarities, the datasets differ in which layers appear the most crucial for memorisation: for \rteb, modifying early layers leads to the largest increase in memorisation error, whereas for \ssttb, both the very first layers and layers in the middle appear most relevant.

\paragraph{Aggregating results}
The findings for these two tasks are echoed in the overall swapping and retraining results. Firstly, \textbf{memorisation is not confined to individual layers}: modifying individual layers barely affects the memorisation error. This is shown in Figure~\ref{fig:one_layer}, which provides the memorisation error when modifying one layer only, taking the \textit{maximum} over layers (i.e.\ highlighting the largest error increase), showing datasets as dots. For most model-dataset combinations, the memorisation error rate is below 15\% when modifying one layer.
This agrees with findings from \citet{maini2023can}, who similarly employed layer retraining to identify that memorisation in image classification is not confined to individual layers.

Secondly, \textbf{the importance of layers does appear task-dependent}. To investigate this more systematically, we express layer relevance using the mean memorisation error (averaging over rows in the result matrices similar to the ones in Figure~\ref{fig:swapping_results_full}). Figure~\ref{fig:taskxtechniques} (top row demonstrating layer swapping) details this error per dataset for \texttt{OPT}. Across the board, \textbf{early layers matter more for memorisation}, but that is more prominent for the NLU tasks than for the other tasks, and for \stormfront\, there is even a slight increase in the error for \textit{deeper} layers. For the remaining models, the same visualisation is shown in Appendix~\ref{ap:additional_results}, and we can summarise the per-layer weights by computing a \textit{Memorisation Centre-of-Gravity} (M-CoG), which is a weighted sum of all layers with weights summing to 1: $\sum_{i=1}^{12}\alpha_i\cdot i$. For layer swapping and retraining, $\alpha_i$ is the memorisation error for layer $i$, as depicted in Figure~\ref{fig:taskxtechniques}. Figure~\ref{fig:retraining_cog} displays the M-CoG coefficients for layer retraining, per model, and Figure~\ref{fig:summary_cogs} provides M-CoG coefficients per dataset by averaging over models (left) and over localisation techniques (right).
The results show \textbf{strong agreement between models in terms of the relative ordering of tasks} -- the average pairwise correlation of the data from Figure~\ref{fig:retraining_cog} is $0.85$ (Spearman's $\rho$) -- and between layer swapping and layer retraining -- Figure~\ref{fig:correlations} (left) includes rank correlations for the M-CoG coefficients, and Figure~\ref{fig:correlations2} (left) includes rank correlations for raw layer weights. Both indicate strong agreement between the two techniques.

\subsection{Probing}
Figure~\ref{fig:taskxtechniques} (bottom row) displays the probing performance for \texttt{OPT}, and the increase from layer to layer indicates layers' relevance.
The first observation is that the performance typically does not decrease for deeper layers, i.e.\ representations do not `lose' information about the fact that some examples are noisy.
Secondly, the performance is quite low for NLU tasks, especially, which could mean that 
clean and noisy examples are more alike for these tasks than for the remaining tasks. 
Lastly, in accordance with the previous results, probing performance does not change suddenly (i.e.\ \textbf{memorisation is not local to individual layers}), and \textbf{tasks differ in how the probing performance changes over layers}: performance flattens early for some tasks (e.g.\ \rte) but gradually improves over all layers for others (e.g.\  \emotion, \ih).
Appendix~\ref{ap:additional_results} provides results for the other models; for \texttt{Pythia}, probing performance peaks earlier than for the other models, indicating that the lower layers are extra important for this model.

\begin{figure}[t]
    \includegraphics[width=.57\columnwidth]{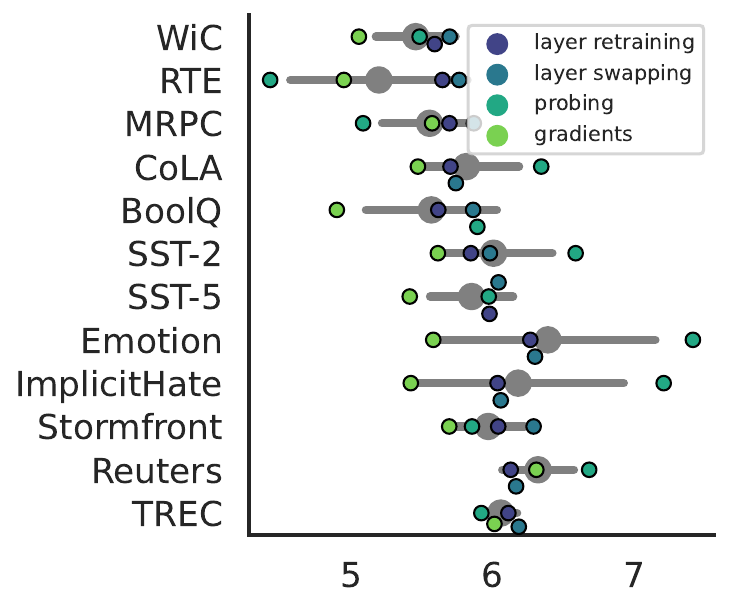}
    \includegraphics[width=.38\columnwidth]{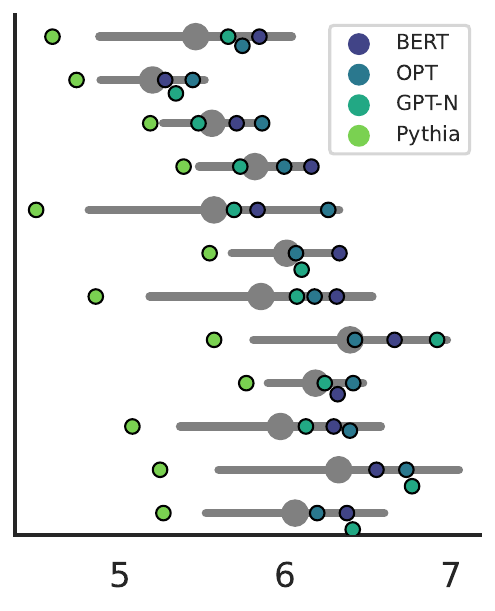}
    \caption{M-CoG coefficients averaged over models (left) and averaged over localisation techniques (right). Error bars show standard deviations.}
    \label{fig:summary_cogs}
   \vspace{-0.5cm}
\end{figure}

To draw more generic conclusions, we compute the M-CoG coefficients by using the per-layer increase in probing performance as weights. Figure~\ref{fig:summary_cogs} (left) includes the M-CoG averaged over models and demonstrates that probing puts a larger emphasis on deeper layers compared to layer swapping and retraining. The M-CoG of probing have a moderately positive correlation to the swapping and retraining coefficients (see Figure~\ref{fig:correlations}), and raw weights per layer have a weakly positive correlation to swapping and retraining (see Figure~\ref{fig:correlations2}).

\subsection{Gradient analysis}

\begin{figure}[t]
    \centering
    \includegraphics[width=.4\columnwidth]{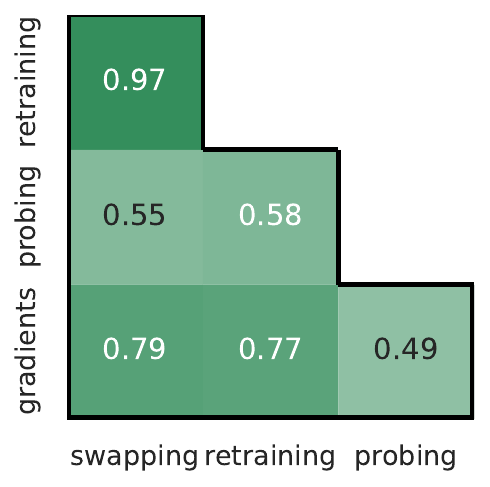}\hspace{.8cm}
    \includegraphics[width=.4\columnwidth]{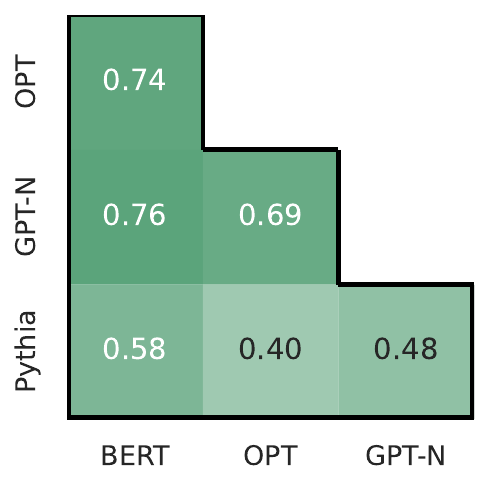}
    \caption{Spearman's $\rho$ for M-CoG from different localisation techniques (left), and different models (right).}
    \label{fig:correlations}
    \vspace{-0.4cm}
\end{figure}

\begin{figure}[t]
    \centering
    \includegraphics[width=.4\columnwidth]{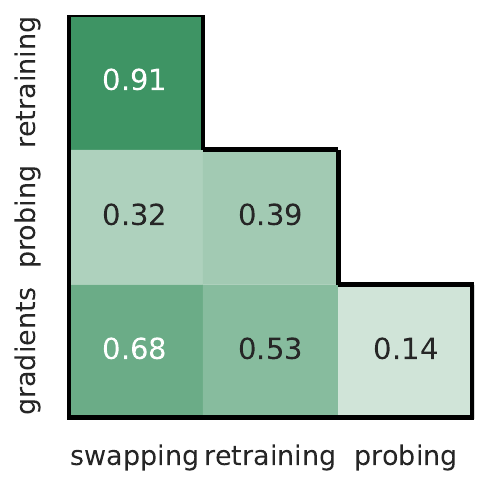}\hspace{.8cm}
    \includegraphics[width=.4\columnwidth]{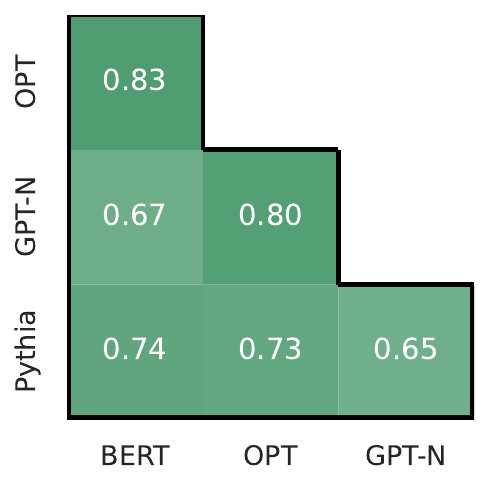}
    \caption{Spearman's $\rho$ for layer-wise scores from different localisation techniques (left), and models (right). When comparing models, we collect weights from 4 techniques. Those are not directly comparable, so we apply min-max normalisation per technique.}
    \label{fig:correlations2}
    \vspace{-0.5cm}
\end{figure}

Finally, we inspect the gradient norms, post-processed as described in Appendix~\ref{ap:hypestimation}.
Results obtained using the forgetting gradients correlate quite strongly with layer swapping and retraining (Figure~\ref{fig:correlations}, Figure~\ref{fig:correlations2}).
That agreement can also be seen when visually inspecting the norms per layer for \texttt{OPT} (middle row of Figure~\ref{fig:taskxtechniques}, see Appendix~\ref{ap:additional_results} for the remaining models): 
NLU tasks have higher scores in earlier layers, \sstt\ and \trec\ have a more uniform distribution, and \stormfront\ and \reuters\ point to deeper layers (although the gradient norms show a slight increase for the final layer for multiple tasks).
At the same time, the gradient analysis weakly correlates to the probing results, potentially because \textit{both} methods have much lower accuracies than swapping/retraining.
The forgetting gradients failed to pinpoint one model's correct layers in the control setup (\S\ref{sec:control_setup}). 
\textit{That} gradients agree with swapping/retraining supports our overall findings, but we recommend against relying solely on gradients for memorisation localisation.


\subsection{Intermediate conclusion}
In this section, we have taken a closer look at the localisation results for \texttt{OPT}, and inspected aggregated results for all techniques and models via M-CoG coefficients.
Because memorisation is not strictly localised to individual layers, these coefficients lie close to the middle layer, but they do generally skew towards earlier layers and provide us with an ordering of tasks.
The most notable pattern in that ordering that the earlier layers are the most important for the NLU tasks.
This is somewhat surprising since the NLP community would typically consider NLU tasks to be more complex than topic classification or sentiment detection, and assumes higher-level tasks to be processed in higher layers.\footnote{E.g.\ \citet{muller2023subspace} show that for topic classification in \texttt{BERT} (using unperturbed datasets), the centre-of-gravity as defined by \citet{tenney2019bert} lies around layer 4/5 for topic classification, whereas for NLI it is layer 11.}
If that is the case, it seems natural for memorisation to also happen in higher layers, but this is contradicted by the experiments.

While this section has concentrated primarily on the comparison of localisation methods, we finally note that when computing correlations between models (Figure~\ref{fig:correlations}-\ref{fig:correlations2}, right), these are strongly positive, except for \texttt{Pythia}, yielding more moderate correlations. That suggests that our results are not specific to one training setup, but somewhat generic to 12-layer transformer-based PLMs.
\begin{figure}[t]\centering
        \begin{subfigure}[b]{0.95\columnwidth}
            \centering
            \includegraphics[width=\textwidth]{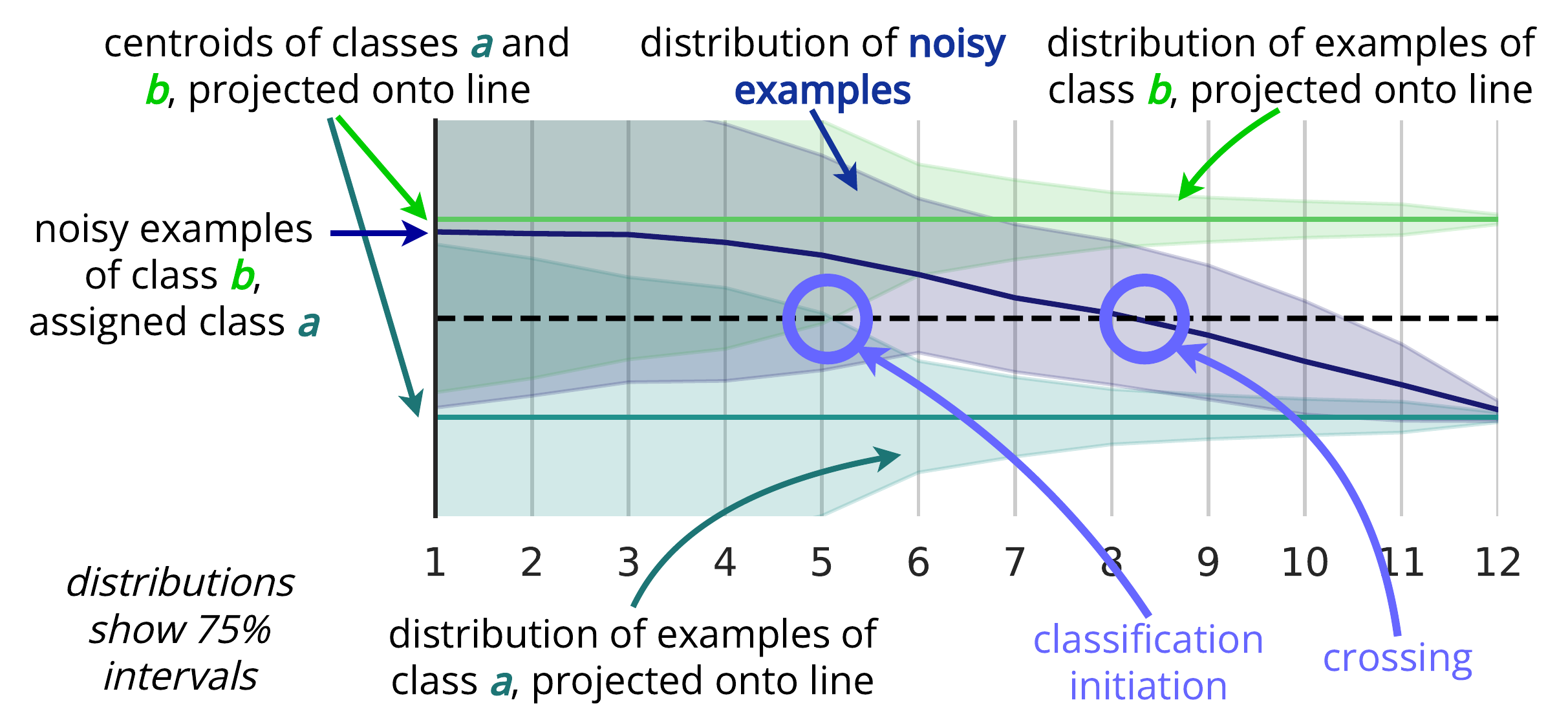}
            \caption{\texttt{TREC}}
            \label{fig:centroid_illustration}
        \end{subfigure}
        \begin{subfigure}[b]{0.46\columnwidth}
            \centering
            \includegraphics[width=\textwidth]{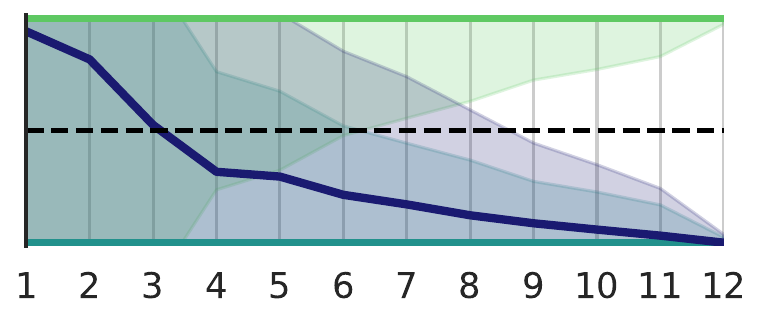}
            \caption{\texttt{RTE}}
            \label{fig:centroid_a}
        \end{subfigure}\hfill
        \begin{subfigure}[b]{0.46\columnwidth}
            \centering
            \includegraphics[width=\textwidth]{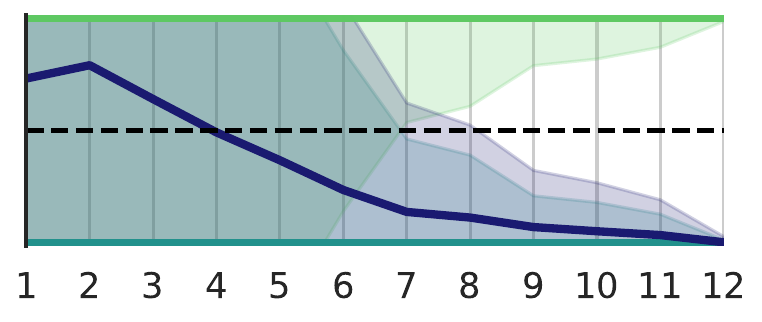}
            \caption{\texttt{WiC}}
            \label{fig:centroid_b}
        \end{subfigure}
        \begin{subfigure}[b]{0.46\columnwidth}
            \centering
            \includegraphics[width=\textwidth]{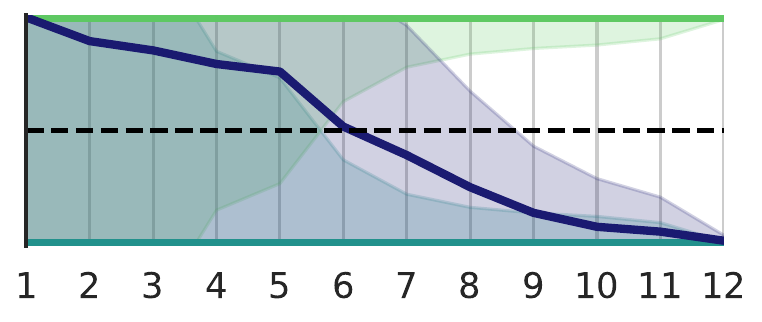}
            \caption{\texttt{SST-2}}
            \label{fig:centroid_c}
        \end{subfigure}\hfill
        \begin{subfigure}[b]{0.46\columnwidth}
            \centering
            \includegraphics[width=\textwidth]{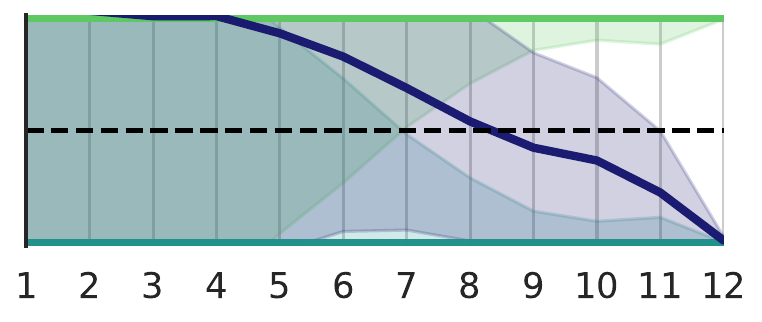}
            \caption{\texttt{Stormfront}}
            \label{fig:centroid_d}
        \end{subfigure}
    \caption{Centroid analysis visualises how noisy examples gradually change, for five datasets, for \texttt{OPT}.}
    \vspace{-0.4cm}
\end{figure}

\begin{figure}[t]\centering
    \begin{subfigure}[b]{0.46\columnwidth}
            \centering
            \includegraphics[width=\textwidth]{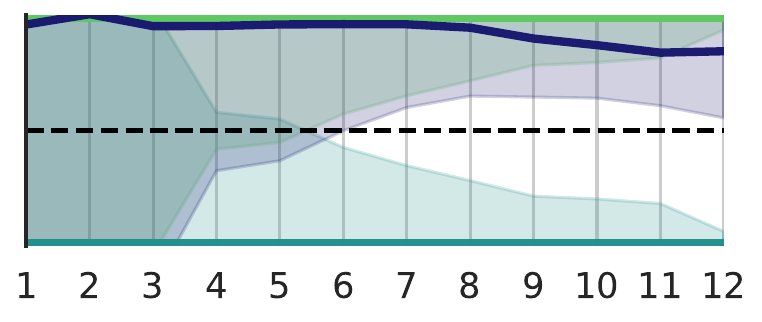}
            \caption{\texttt{RTE}, regular bottom}
    \end{subfigure}\hfill
    \begin{subfigure}[b]{0.46\columnwidth}
            \centering
            \includegraphics[width=\textwidth]{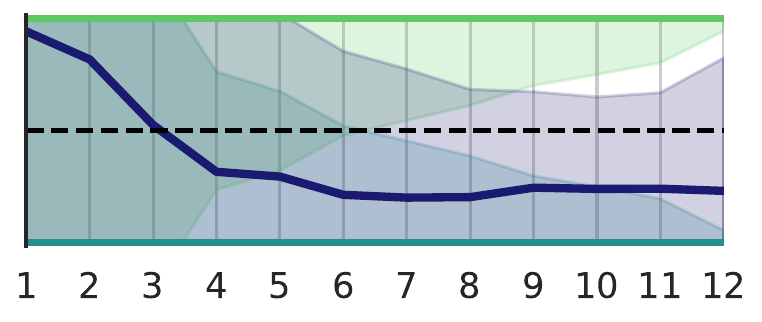}
            \caption{\texttt{RTE}, regular top}
    \end{subfigure}
        \begin{subfigure}[b]{0.46\columnwidth}
            \centering
            \includegraphics[width=\textwidth]{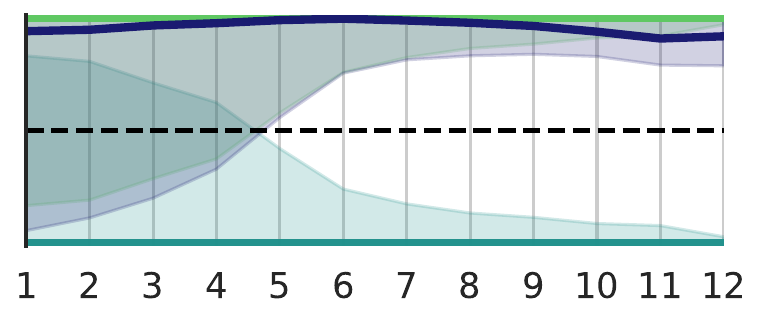}
            \caption{\texttt{TREC}, regular bottom}
    \end{subfigure}\hfill
    \hfill
    \begin{subfigure}[b]{0.46\columnwidth}
            \centering
            \includegraphics[width=\textwidth]{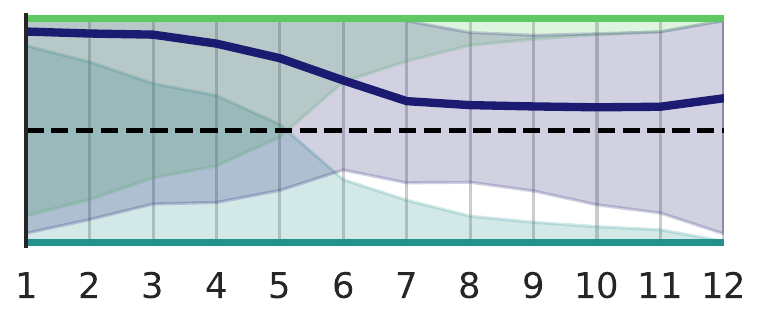}
            \caption{\texttt{TREC}, regular top}
    \end{subfigure}
    \caption{Illustration of the effect replacing 6 layers has, for \texttt{RTE} and \texttt{TREC} data, for \texttt{OPT}. We insert 6 layers from $\theta_O$ at the lower (`bottom') or upper half (`top').}
    \label{fig:centroid_replacement}
    \vspace{-0.3cm}
\end{figure}

\section{Making memorisation interpretable via centroid analysis and probing}
\label{sec:centroid_analysis}

The results from \S\ref{sec:results} suggested that earlier layers are the most relevant for memorisation. To better understand why, we make models' internal processing of memorised examples more interpretable through a \textbf{centroid analysis}: we examine pairs of classes, monitoring examples with original class $y_b$ and noisy class $y_a$, for all pairs of $a$ and $b$.
We compute the centroids of the hidden representations from the two classes and project all data points from those two classes onto the line through the centroids. We measure the distance to centroid $a$ for every data point, normalised by the distance between the two centroids. This is performed separately per layer. In layer 1, points belonging to $y_a$ and $y_b$ largely overlap. Towards layer 12, the two classes are fully separated, and in between, the memorised examples move away from centroid $b$ and move towards centroid $a$. Figure~\ref{fig:centroid_illustration} explains this via annotations for \trecb, and Figures~\ref{fig:centroid_a}-\ref{fig:centroid_d} do so for four additional tasks that are illustrative of the variety that we observe.
We include figures for all tasks and models in Appendix~\ref{ap:centroid_analysis}.

This visualisation indicates that memorisation occurs through \textit{gradual} changes from the first layers onward. This explains the results from the previous section, where we found that memorisation is not confined to individual layers and that lower layers were more successful in reverting memorisation (in layer swapping/retraining) than deeper layers: memorisation \textit{starts} early, and interventions are more successful when conducted before the hidden state has moved too far away from class $y_b$. We can demonstrate this using the centroid analysis, applied while swapping six layers at the bottom or top with layers from $\theta_O$. For all tasks (see Appendix~\ref{ap:centroid_analysis}), we can prevent the noisy examples from moving to centroid $a$ by replacing the bottom six, but for only a few tasks, replacing the top six has a similar effect.
Figure~\ref{fig:centroid_replacement} shows this using \rteb\ and \trecb. Swapping the bottom six prevents emitting the noisy class for \rteb\ \textit{and} \trecb, but swapping the top six is more successful for \trecb\ than \rteb.

\begin{figure}[t]
    \centering
    \begin{subfigure}[b]{\columnwidth}
        \includegraphics[width=\textwidth]{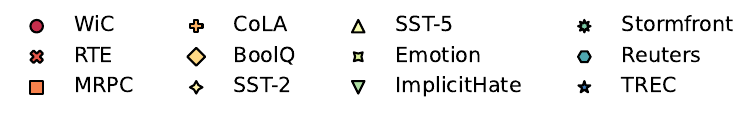}
    \end{subfigure}
    \begin{subfigure}[b]{0.49\columnwidth}
        \includegraphics[width=\textwidth]{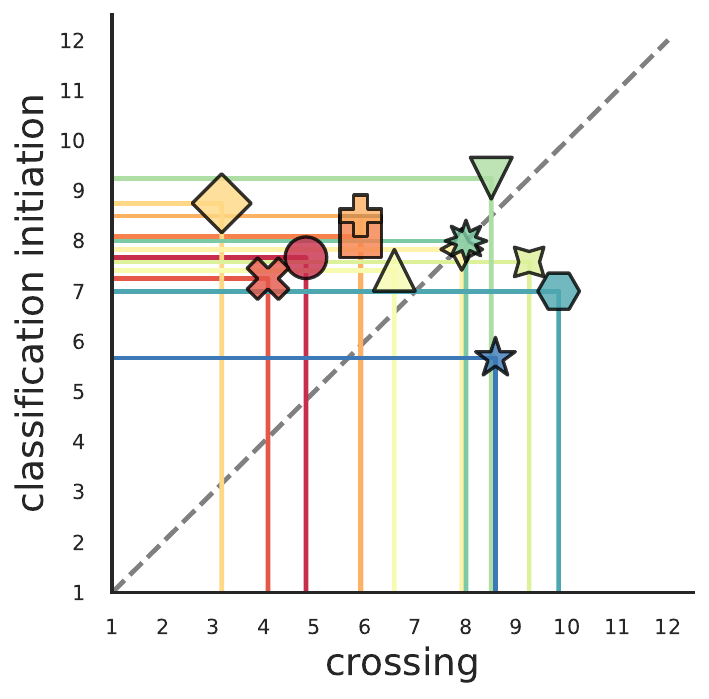}
        \caption{Centroid analysis}
        \label{fig:events_averaged_centroid}
    \end{subfigure}
    \begin{subfigure}[b]{0.49\columnwidth}
        \includegraphics[width=\textwidth]{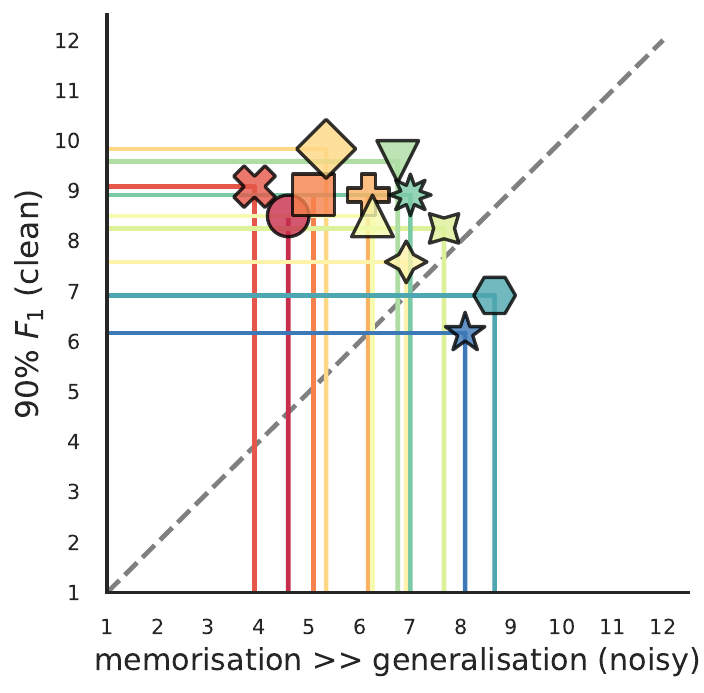}
        \caption{Probing}
        \label{fig:events_averaged_probing}
    \end{subfigure}
    \caption{Summary of the memorisation and classification onset for all datasets, averaged over models, computed using the centroid analysis or via probing.}
    \label{fig:events_averaged}
\end{figure}

\paragraph{Task differences}
To summarise task differences, we compute two statistics: the \textbf{crossing} (the first layer in which the noisy mean is closer to $a$ than to $b$) and \textbf{classification initiation} (the first layer without overlapping distributions for the two classes), shown in Figure~\ref{fig:events_averaged_centroid}.
Many NLU tasks have an early crossing and a late classification initiation (e.g. \rte\ and \wic). The two events are closer together for hate speech and sentiment tasks, and topic classification tasks (\trec\ and \reuters) start classification early but have a late crossing. This confirms findings from \S\ref{sec:results}: lower-level tasks (early classification initiation) rely more heavily on deeper layers for memorisation than higher-level tasks (late classification initiation). Yet, tasks with similar classification initiations (e.g. \sstf\ and \emotion) can still have different crossings.

\paragraph{Consolidation via probing}The centroid analysis merely visualises representations. To consolidate that we reach similar conclusions using different methods, we train probes to predict an example's class from the hidden state, using (i) original or (ii) noisy labels. In Appendix~\ref{ap:probing}, we include the probes' performances. We apply the probes to noisy examples and compute a statistic similar to the crossing: the layer at which the $F_1$ of probe ii exceeds the $F_1$ of probe i with ten percentage points, referred to as `memorisation>>generalisation' in Figure~\ref{fig:events_averaged_probing}. The timing of this event strongly correlates with the crossings (Spearman's $\rho=0.84$). 
We apply the probes to clean examples to compute a statistic similar to the classification initiation: the layer at which the probes' $F_1$ for clean examples (normalised by random guessing performance) reaches 90\%. 
The depth of this event strongly correlates with classification initiation ($\rho=0.73$). Together, these two events thus tell a story similar to that of the centroid analysis (Figure~\ref{fig:events_averaged_probing}).

\begin{table}[t]
    \centering\small
    \begin{tabular}{lccccc}
    \toprule
    \textbf{Correlates} & \texttt{all} & \texttt{BERT} & \texttt{OPT} & \texttt{GPT-N} & \texttt{Pythia} \\\midrule
    \multicolumn{3}{l}{\textit{- Generalisation score}} \\
    crossing     & 0.75 & 0.88 & 0.94 & 0.94 & 0.72 \\
    mem. >> gen. & 0.63 & 0.86 & 0.88 & 0.94 & 0.69 \\
    M-CoG        & 0.56 & 0.78 & 0.69 & 0.92 & 0.69 \\ 
    \multicolumn{3}{l}{\textit{- Validation score}} \\
    crossing     & 0.70 & 0.90 & 0.84 & 0.83 & 0.70 \\
    mem. >> gen. & 0.61 & 0.90 & 0.77 & 0.76 & 0.77 \\
    M-CoG        & 0.54 & 0.80 & 0.52* & 0.72 & 0.69 \\
    \bottomrule
    \end{tabular}
    \caption{Spearman's $\rho$ relating memorisation for the 12 tasks to models' generalisation performances. *: $p{>}0.05$}
    \label{tab:memvsgen}
    \vspace{-0.3cm}
\end{table}

\paragraph{Memorisation's connection to generalisation}
When inspecting model internals, we have seen that the depth of memorisation (quantified as M-CoG coefficients, the `crossing' and `memorisation>>generalisation') appears anti-correlated with the difficulty of a task. However, we have yet to have a proper way of quantifying that difficulty.
We now take models' \textbf{validation accuracy} at the end of training (on data unseen during training, normalised by random guessing performance) and compute a \textbf{generalisation score} (percentage of \textit{training examples} for which the probability of the target exceeds random guessing when that example is held out from training).\footnote{Computed by training on a randomly selected 50\% of the data, and testing on the held-out 50\%, repeated with 30 random seeds. This approximates metrics reported by \citet{feldman2020does,feldman2020neural,jiang2021characterizing}. We adopt the naming of the metric from \citet{dankers2023memorisation}.}
As indicated in Table~\ref{tab:memvsgen}, these two metrics correlate moderately with the memorisation depth when combining data from all models (Spearman's $\rho>0.54$), with most correlations being stronger when examining results per model. 
All in all, this suggests that the better a model generalises a task to new data, the more later layers are involved in memorisation.\footnote{Tasks with late crossings are mostly multi-class tasks. To ensure that this is not a confound here, Appendix~\ref{ap:binarised_tasks} repeats some of the analyses with binary versions of these tasks.}

\section{Discussion}
\label{sec:discussion}

We set out to perform memorisation localisation for natural language classification tasks by perturbing a subset of the labels and tracing those `noisy' examples over layers. Applying four localisation techniques to four models crystallised that memorisation is not local to specific layers but a cooperative process of weights from many layers.
Nonetheless, not all layers appear equally important. Overall, early layers are more important than later ones: the model's manipulation of memorised examples \textit{starts} in lower layers, and to prevent memorisation, early intervention is more successful than late intervention.
We discussed results for 12-layer models in the main paper, but further experimentation with a 24-layer model (in Appendix~\ref{ap:opt_big}) leads to similar findings of memorisation being gradual, with a similar ordering of tasks, but \textit{mid-range} layers being more important than late layers.

This is not in accordance with the generalisation-first, memorisation-second hypothesis from CV (see \S\ref{sec:related_work}), but does agree with more recent work on image classification by \citet{maini2023can}. 
It also aligns with related work on PLMs for fact memorisation and verbatim memorisation pointing to the lower layers \citep{geva2023dissecting,stoehr2024localizing}, while also describing cooperative roles for earlier and deeper layers \citep{haviv2023understanding}.

The fact that memorisation is not local implies that editing model weights locally does not necessarily erase memorised information, even if a flipped label suggests this at the level of the output layer. This might be harmless when editing facts about named entities like cities \citep[e.g.][]{meng2022locating}, yet is more worrisome when regarding memorisation of personal information \citep[e.g.][]{carlini2021extracting}, and may be a reason why safety measures can easily be reversed in PLMs modified to reduce harmful outputs \citep[e.g.][]{zhan2023removing}. 

Can we, due to the importance of early layers, conclude that our results falsify the generalisation first, memorisation second hypothesis? The results from \S\ref{sec:centroid_analysis} suggest that this question requires a nuanced answer due to the variation observed among tasks. The depth of memorisation is positively correlated with a model's generalisation capabilities, i.e.\ we do observe a generalisation first, memorisation second tendency but the better the model performs at a certain task, the stronger that tendency is.
We consider better understanding what properties of a task direct memorisation to lower or higher layers an exciting avenue for future work.

\section*{Limitations}
\label{sec:limitations}

We identify four main limitations of our work:
\begin{itemize}
    \item \textbf{Simplified data}: To trace memorised examples over the transformer's many layers, we resorted to label flipping to create `noisy' examples. This situation is somewhat unnatural when considering real-world examples that require memorisation from the model. For example, in the case of sentiment analysis, that might be a sarcastic phrase whose sentiment is the opposite of what is expected based on a literal interpretation. We cannot guarantee that our noisy examples behave in the same way as real-world examples would. Similarly, memorisation of noisy examples need not affect models in the same way as the memorisation of factual information or verbatim memorisation of long strings. As laid out in the introduction (\S\ref{sec:introduction}), we opted for this type of data manipulation to create an experimental setup that more closely resembles that of related work from CV. 
    \item \textbf{Localisation techniques are imperfect:} As identified in \S\ref{sec:control_setup} the localisation techniques applied are themselves flawed: in a control setup where only two layers were modified during fine-tuning, probing and gradient analyses could not accurately pinpoint those two layers, and the techniques that could pinpoint them (layer swapping and layer retraining) are more reliable at determining which layers are \textit{not} crucial for memorisation than at determining which ones are. Because of the general agreement between the techniques and the results from \S\ref{sec:results}-\ref{sec:centroid_analysis} we do think our conclusions are robust, but the absolute numbers of layer relevance should be taken with a grain of salt.
    \item \textbf{Visualisation $\neq$ localisation:} In \S\ref{sec:centroid_analysis} we introduced the centroid analysis as a way of visualising what is happening to examples over the different layers. This visualisation is a one-dimensional projection of hidden representations and thus an extreme simplification of the intricate process of memorisation. We do not mean to use it as a localisation technique, but as a way to explain the outcomes of other experiments in the paper.
    \item \textbf{Lack of evidence for individual examples:} We analysed the group of noisy examples as a whole, and concluded that many layers work together to gradually shift examples from their original class to the newly assigned class. However, we have not examined individual examples; it can still be the case that for individual examples, memorisation is more localised to specific layers. We only have preliminary results suggesting that individual examples, too, are memorised over multiple layers, which is the fact that in \S\ref{sec:results}, swapping/retraining individual layers was unsuccessful in increasing the memorisation error rate.
\end{itemize}
\section*{Acknowledgements}

VD is supported by the UKRI Centre for Doctoral Training in Natural Language Processing, funded by the UKRI (grant EP/S022481/1) and the University of Edinburgh, School of Informatics and School of Philosophy, Psychology \& Language Sciences.
IT is supported by the Dutch National Science Foundation (NWO Vici VI.C.212.053).

\bibliographystyle{acl_natbib}
\bibliography{memorisation}

\begin{thebibliography}{60}
\expandafter\ifx\csname natexlab\endcsname\relax\def\natexlab#1{#1}\fi

\bibitem[{Akyürek et~al.(2022)Akyürek, Bolukbasi, Liu, Xiong, Tenney,
  Andreas, and Guu}]{akyurek2022tracing}
Ekin Akyürek, Tolga Bolukbasi, Frederick Liu, Binbin Xiong, Ian Tenney, Jacob
  Andreas, and Kelvin Guu. 2022.
\newblock \href {https://doi.org/10.18653/v1/2022.findings-emnlp.180} {Towards
  tracing knowledge in language models back to the training data}.
\newblock In \emph{Findings of the Association for Computational Linguistics:
  EMNLP 2022}, pages 2429--2446.

\bibitem[{Ansuini et~al.(2019)Ansuini, Laio, Macke, and
  Zoccolan}]{ansuini2019intrinsic}
Alessio Ansuini, Alessandro Laio, Jakob~H Macke, and Davide Zoccolan. 2019.
\newblock \href
  {https://proceedings.neurips.cc/paper/2019/hash/cfcce0621b49c983991ead4c3d4d3b6b-Abstract.html}
  {Intrinsic dimension of data representations in deep neural networks}.
\newblock \emph{Advances in Neural Information Processing Systems},
  32:6111--6122.

\bibitem[{Apte et~al.(1994)Apte, Damerau, and Weiss}]{apte1994towards}
Chidanand Apte, Fred Damerau, and Sholom~M Weiss. 1994.
\newblock \href {https://link.springer.com/chapter/10.1007/978-1-4471-2099-5_3}
  {Towards language independent automated learning of text categorization
  models}.
\newblock In \emph{SIGIR’94: Proceedings of the Seventeenth Annual
  International ACM-SIGIR Conference on Research and Development in Information
  Retrieval, organised by Dublin City University}, pages 23--30. Springer.

\bibitem[{Baldock et~al.(2021)Baldock, Maennel, and
  Neyshabur}]{baldock2021deep}
Robert Baldock, Hartmut Maennel, and Behnam Neyshabur. 2021.
\newblock \href
  {https://proceedings.neurips.cc/paper/2021/file/5a4b25aaed25c2ee1b74de72dc03c14e-Paper.pdf}
  {Deep learning through the lens of example difficulty}.
\newblock \emph{Advances in Neural Information Processing Systems},
  34:10876--10889.

\bibitem[{Biderman et~al.(2024)Biderman, Prashanth, Sutawika, Schoelkopf,
  Anthony, Purohit, and Raff}]{biderman2024emergent}
Stella Biderman, USVSN~Sai Prashanth, Lintang Sutawika, Hailey Schoelkopf,
  Quentin Anthony, Shivanshu Purohit, and Edward Raff. 2024.
\newblock \href
  {https://proceedings.neurips.cc/paper_files/paper/2023/file/59404fb89d6194641c69ae99ecdf8f6d-Paper-Conference.pdf}
  {Emergent and predictable memorization in large language models}.
\newblock \emph{Advances in Neural Information Processing Systems},
  36:28072--28090.

\bibitem[{Biderman et~al.(2023)Biderman, Schoelkopf, Anthony, Bradley,
  O’Brien, Hallahan, Khan, Purohit, Prashanth, Raff
  et~al.}]{biderman2023pythia}
Stella Biderman, Hailey Schoelkopf, Quentin~Gregory Anthony, Herbie Bradley,
  Kyle O’Brien, Eric Hallahan, Mohammad~Aflah Khan, Shivanshu Purohit,
  USVSN~Sai Prashanth, Edward Raff, et~al. 2023.
\newblock \href
  {https://proceedings.mlr.press/v202/biderman23a/biderman23a.pdf} {Pythia: A
  suite for analyzing large language models across training and scaling}.
\newblock In \emph{International Conference on Machine Learning}, pages
  2397--2430. PMLR.

\bibitem[{Black et~al.(2021)Black, Leo, Wang, Leahy, and Biderman}]{gpt-neo}
Sid Black, Gao Leo, Phil Wang, Connor Leahy, and Stella Biderman. 2021.
\newblock \href {https://doi.org/10.5281/zenodo.5297715} {{GPT-Neo: Large Scale
  Autoregressive Language Modeling with Mesh-Tensorflow}}.

\bibitem[{Carlini et~al.(2022)Carlini, Ippolito, Jagielski, Lee, Tramer, and
  Zhang}]{carlini2022quantifying}
Nicholas Carlini, Daphne Ippolito, Matthew Jagielski, Katherine Lee, Florian
  Tramer, and Chiyuan Zhang. 2022.
\newblock \href {https://openreview.net/forum?id=TatRHT_1cK} {Quantifying
  memorization across neural language models}.
\newblock In \emph{International Conference on Learning Representations}.

\bibitem[{Carlini et~al.(2021)Carlini, Tramer, Wallace, Jagielski,
  Herbert-Voss, Lee, Roberts, Brown, Song, Erlingsson
  et~al.}]{carlini2021extracting}
Nicholas Carlini, Florian Tramer, Eric Wallace, Matthew Jagielski, Ariel
  Herbert-Voss, Katherine Lee, Adam Roberts, Tom Brown, Dawn Song, Ulfar
  Erlingsson, et~al. 2021.
\newblock \href
  {https://www.usenix.org/system/files/sec21-carlini-extracting.pdf}
  {Extracting training data from large language models}.
\newblock In \emph{30th USENIX Security Symposium (USENIX Security 21)}, pages
  2633--2650.

\bibitem[{Chang et~al.(2023{\natexlab{a}})Chang, Cramer, Soni, and
  Bamman}]{chang2023speak}
Kent Chang, Mackenzie Cramer, Sandeep Soni, and David Bamman.
  2023{\natexlab{a}}.
\newblock \href {https://arxiv.org/abs/2305.00118} {Speak, memory: An
  archaeology of books known to {ChatGPT/GPT-4}}.
\newblock In \emph{Proceedings of the 2023 Conference on Empirical Methods in
  Natural Language Processing}, pages 7312--7327.

\bibitem[{Chang et~al.(2023{\natexlab{b}})Chang, Thomason, and
  Jia}]{chang2023localization}
Ting-Yun Chang, Jesse Thomason, and Robin Jia. 2023{\natexlab{b}}.
\newblock \href {https://arxiv.org/pdf/2311.09060.pdf} {Do localization methods
  actually localize memorized data in {LLMs}?}
\newblock \emph{arXiv preprint arXiv:2311.09060}.

\bibitem[{Chen et~al.(2024)Chen, Cao, Chen, Liu, and Zhao}]{chen2024journey}
Yuheng Chen, Pengfei Cao, Yubo Chen, Kang Liu, and Jun Zhao. 2024.
\newblock \href {https://arxiv.org/pdf/2308.13198} {Journey to the center of
  the knowledge neurons: Discoveries of language-independent knowledge neurons
  and degenerate knowledge neurons}.
\newblock In \emph{Proceedings of the AAAI Conference on Artificial
  Intelligence}, volume~38, pages 17817--17825.

\bibitem[{Clark et~al.(2019)Clark, Lee, Chang, Kwiatkowski, Collins, and
  Toutanova}]{clark2019boolq}
Christopher Clark, Kenton Lee, Ming-Wei Chang, Tom Kwiatkowski, Michael
  Collins, and Kristina Toutanova. 2019.
\newblock \href {https://aclanthology.org/N19-1300/} {{BoolQ}: Exploring the
  surprising difficulty of natural yes/no questions}.
\newblock In \emph{Proceedings of the 2019 Conference of the North American
  Chapter of the Association for Computational Linguistics: Human Language
  Technologies, Volume 1 (Long and Short Papers)}, pages 2924--2936.

\bibitem[{Cohen et~al.(2018)Cohen, Sapiro, and Giryes}]{cohen2018dnn}
Gilad Cohen, Guillermo Sapiro, and Raja Giryes. 2018.
\newblock \href {https://arxiv.org/pdf/1805.06822} {{DNN or k-NN}: That is the
  generalize vs. memorize question}.
\newblock \emph{arXiv preprint arXiv:1805.06822}.

\bibitem[{Conneau et~al.(2018)Conneau, Kruszewski, Lample, Barrault, and
  Baroni}]{conneau-etal-2018-cram}
Alexis Conneau, German Kruszewski, Guillaume Lample, Lo{\"\i}c Barrault, and
  Marco Baroni. 2018.
\newblock \href {https://doi.org/10.18653/v1/P18-1198} {What you can cram into
  a single {\$}{\&}!{\#}* vector: Probing sentence embeddings for linguistic
  properties}.
\newblock In \emph{Proceedings of the 56th Annual Meeting of the Association
  for Computational Linguistics (Volume 1: Long Papers)}, pages 2126--2136.

\bibitem[{Dagan et~al.(2006)Dagan, Glickman, and Magnini}]{rte1}
Ido Dagan, Oren Glickman, and Bernardo Magnini. 2006.
\newblock \href {https://link.springer.com/chapter/10.1007/11736790_9} {The
  {PASCAL} recognising textual entailment challenge}.
\newblock In \emph{Machine Learning Challenges. Evaluating Predictive
  Uncertainty, Visual Object Classification, and Recognising Textual
  Entailment}. Springer.

\bibitem[{Dai et~al.(2022)Dai, Dong, Hao, Sui, Chang, and
  Wei}]{dai2022knowledge}
Damai Dai, Li~Dong, Yaru Hao, Zhifang Sui, Baobao Chang, and Furu Wei. 2022.
\newblock \href {https://arxiv.org/pdf/2104.08696} {Knowledge neurons in
  pretrained transformers}.
\newblock In \emph{Proceedings of the 60th Annual Meeting of the Association
  for Computational Linguistics (Volume 1: Long Papers)}, pages 8493--8502.

\bibitem[{Dankers et~al.(2023)Dankers, Titov, and
  Hupkes}]{dankers2023memorisation}
Verna Dankers, Ivan Titov, and Dieuwke Hupkes. 2023.
\newblock \href {https://arxiv.org/pdf/2311.05379} {Memorisation cartography:
  Mapping out the memorisation-generalisation continuum in neural machine
  translation}.
\newblock In \emph{Proceedings of the 2023 Conference on Empirical Methods in
  Natural Language Processing}, pages 8323--8343.

\bibitem[{De~Cao et~al.(2021)De~Cao, Aziz, and Titov}]{de2021editing}
Nicola De~Cao, Wilker Aziz, and Ivan Titov. 2021.
\newblock \href {https://arxiv.org/pdf/2104.08164} {Editing factual knowledge
  in language models}.
\newblock In \emph{Proceedings of the 2021 Conference on Empirical Methods in
  Natural Language Processing}, pages 6491--6506.

\bibitem[{de~Gibert et~al.(2018)de~Gibert, P{\'e}rez, Garc{\'\i}a-Pablos, and
  Cuadros}]{de2018hate}
Ona de~Gibert, Naiara P{\'e}rez, Aitor Garc{\'\i}a-Pablos, and Montse Cuadros.
  2018.
\newblock \href {https://arxiv.org/pdf/1809.04444} {Hate speech dataset from a
  white supremacy forum}.
\newblock In \emph{Proceedings of the 2nd Workshop on Abusive Language Online
  (ALW2)}, pages 11--20.

\bibitem[{Devlin et~al.(2019)Devlin, Chang, Lee, and
  Toutanova}]{devlin-etal-2019-bert}
Jacob Devlin, Ming-Wei Chang, Kenton Lee, and Kristina Toutanova. 2019.
\newblock \href {https://doi.org/10.18653/v1/N19-1423} {{BERT}: Pre-training of
  deep bidirectional transformers for language understanding}.
\newblock In \emph{Proceedings of the 2019 Conference of the North {A}merican
  Chapter of the Association for Computational Linguistics: Human Language
  Technologies, Volume 1 (Long and Short Papers)}, pages 4171--4186.

\bibitem[{Dolan and Brockett(2005)}]{dolan2005automatically}
Bill Dolan and Chris Brockett. 2005.
\newblock \href
  {https://www.microsoft.com/en-us/research/wp-content/uploads/2016/02/I05-50025B15D.pdf}
  {Automatically constructing a corpus of sentential paraphrases}.
\newblock In \emph{Third International Workshop on Paraphrasing (IWP2005)}.

\bibitem[{ElSherief et~al.(2021)ElSherief, Ziems, Muchlinski, Anupindi,
  Seybolt, De~Choudhury, and Yang}]{elsherief2021latent}
Mai ElSherief, Caleb Ziems, David Muchlinski, Vaishnavi Anupindi, Jordyn
  Seybolt, Munmun De~Choudhury, and Diyi Yang. 2021.
\newblock \href {https://arxiv.org/pdf/2109.05322} {Latent hatred: A benchmark
  for understanding implicit hate speech}.
\newblock In \emph{Proceedings of the 2021 Conference on Empirical Methods in
  Natural Language Processing}, pages 345--363.

\bibitem[{Feldman(2020)}]{feldman2020does}
Vitaly Feldman. 2020.
\newblock \href {https://dl.acm.org/doi/pdf/10.1145/3357713.3384290} {Does
  learning require memorization? {A} short tale about a long tail}.
\newblock In \emph{Proceedings of the 52nd Annual ACM SIGACT Symposium on
  Theory of Computing}, pages 954--959.

\bibitem[{Feldman and Zhang(2020)}]{feldman2020neural}
Vitaly Feldman and Chiyuan Zhang. 2020.
\newblock \href
  {https://proceedings.neurips.cc/paper_files/paper/2020/file/1e14bfe2714193e7af5abc64ecbd6b46-Paper.pdf}
  {What neural networks memorize and why: Discovering the long tail via
  influence estimation}.
\newblock \emph{Advances in Neural Information Processing Systems},
  33:2881--2891.

\bibitem[{Gao et~al.(2020)Gao, Biderman, Black, Golding, Hoppe, Foster, Phang,
  He, Thite, Nabeshima et~al.}]{gao2020pile}
Leo Gao, Stella Biderman, Sid Black, Laurence Golding, Travis Hoppe, Charles
  Foster, Jason Phang, Horace He, Anish Thite, Noa Nabeshima, et~al. 2020.
\newblock \href {https://arxiv.org/pdf/2101.00027} {The pile: An 800gb dataset
  of diverse text for language modeling}.
\newblock \emph{arXiv preprint arXiv:2101.00027}.

\bibitem[{Geva et~al.(2023)Geva, Bastings, Filippova, and
  Globerson}]{geva2023dissecting}
Mor Geva, Jasmijn Bastings, Katja Filippova, and Amir Globerson. 2023.
\newblock \href {https://aclanthology.org/2023.emnlp-main.751/} {Dissecting
  recall of factual associations in auto-regressive language models}.
\newblock In \emph{Proceedings of the 2023 Conference on Empirical Methods in
  Natural Language Processing}, pages 12216--12235.

\bibitem[{Geva et~al.(2021)Geva, Schuster, Berant, and
  Levy}]{geva2021transformer}
Mor Geva, Roei Schuster, Jonathan Berant, and Omer Levy. 2021.
\newblock \href {https://arxiv.org/pdf/2012.14913} {Transformer feed-forward
  layers are key-value memories}.
\newblock In \emph{Proceedings of the 2021 Conference on Empirical Methods in
  Natural Language Processing}, pages 5484--5495.

\bibitem[{Hase et~al.(2024)Hase, Bansal, Kim, and Ghandeharioun}]{hase2023does}
Peter Hase, Mohit Bansal, Been Kim, and Asma Ghandeharioun. 2024.
\newblock \href {https://arxiv.org/abs/2301.04213} {Does localization inform
  editing? {S}urprising differences in causality-based localization vs.
  knowledge editing in language models}.
\newblock \emph{Advances in Neural Information Processing Systems},
  36:17643--17668.

\bibitem[{Haviv et~al.(2023)Haviv, Cohen, Gidron, Schuster, Goldberg, and
  Geva}]{haviv2023understanding}
Adi Haviv, Ido Cohen, Jacob Gidron, Roei Schuster, Yoav Goldberg, and Mor Geva.
  2023.
\newblock \href {https://arxiv.org/pdf/2210.03588} {Understanding transformer
  memorization recall through idioms}.
\newblock In \emph{Proceedings of the 17th Conference of the European Chapter
  of the Association for Computational Linguistics}, pages 248--264.

\bibitem[{Hovy et~al.(2001)Hovy, Gerber, Hermjakob, Lin, and
  Ravichandran}]{hovy-etal-2001-toward}
Eduard Hovy, Laurie Gerber, Ulf Hermjakob, Chin-Yew Lin, and Deepak
  Ravichandran. 2001.
\newblock \href {https://www.aclweb.org/anthology/H01-1069} {Toward
  semantics-based answer pinpointing}.
\newblock In \emph{Proceedings of the First International Conference on Human
  Language Technology Research}.

\bibitem[{Jiang et~al.(2021)Jiang, Zhang, Talwar, and
  Mozer}]{jiang2021characterizing}
Ziheng Jiang, Chiyuan Zhang, Kunal Talwar, and Michael~C Mozer. 2021.
\newblock \href {https://arxiv.org/pdf/2002.03206} {Characterizing structural
  regularities of labeled data in overparameterized models}.
\newblock In \emph{International Conference on Machine Learning}, pages
  5034--5044. PMLR.

\bibitem[{Li and Roth(2002)}]{li-roth-2002-learning}
Xin Li and Dan Roth. 2002.
\newblock \href {https://www.aclweb.org/anthology/C02-1150} {Learning question
  classifiers}.
\newblock In \emph{{COLING} 2002: The 19th International Conference on
  Computational Linguistics}.

\bibitem[{Maini et~al.(2023)Maini, Mozer, Sedghi, Lipton, Kolter, and
  Zhang}]{maini2023can}
Pratyush Maini, Michael~Curtis Mozer, Hanie Sedghi, Zachary~Chase Lipton,
  J~Zico Kolter, and Chiyuan Zhang. 2023.
\newblock \href {https://arxiv.org/pdf/2307.09542.pdf} {Can neural network
  memorization be localized?}
\newblock In \emph{International Conference on Machine Learning}, pages
  23536--23557. PMLR.

\bibitem[{Meng et~al.(2022{\natexlab{a}})Meng, Bau, Andonian, and
  Belinkov}]{meng2022locating}
Kevin Meng, David Bau, Alex Andonian, and Yonatan Belinkov. 2022{\natexlab{a}}.
\newblock \href
  {https://proceedings.neurips.cc/paper_files/paper/2022/file/6f1d43d5a82a37e89b0665b33bf3a182-Paper-Conference.pdf}
  {Locating and editing factual associations in {GPT}}.
\newblock \emph{Advances in Neural Information Processing Systems},
  35:17359--17372.

\bibitem[{Meng et~al.(2022{\natexlab{b}})Meng, Sharma, Andonian, Belinkov, and
  Bau}]{meng2022mass}
Kevin Meng, Arnab~Sen Sharma, Alex~J Andonian, Yonatan Belinkov, and David Bau.
  2022{\natexlab{b}}.
\newblock \href {https://openreview.net/forum?id=MkbcAHIYgyS} {Mass-editing
  memory in a transformer}.
\newblock In \emph{International Conference on Learning Representations}.

\bibitem[{Mireshghallah et~al.(2022)Mireshghallah, Uniyal, Wang, Evans, and
  Berg-Kirkpatrick}]{mireshghallah2022empirical}
Fatemehsadat Mireshghallah, Archit Uniyal, Tianhao Wang, David~K Evans, and
  Taylor Berg-Kirkpatrick. 2022.
\newblock \href {https://aclanthology.org/2022.emnlp-main.119.pdf} {An
  empirical analysis of memorization in fine-tuned autoregressive language
  models}.
\newblock In \emph{Proceedings of the 2022 Conference on Empirical Methods in
  Natural Language Processing}, pages 1816--1826.

\bibitem[{Morcos et~al.(2018)Morcos, Raghu, and Bengio}]{morcos2018insights}
Ari Morcos, Maithra Raghu, and Samy Bengio. 2018.
\newblock \href
  {https://proceedings.neurips.cc/paper/2018/file/a7a3d70c6d17a73140918996d03c014f-Paper.pdf}
  {Insights on representational similarity in neural networks with canonical
  correlation}.
\newblock \emph{Advances in neural information processing systems}, 31.

\bibitem[{M{\"u}ller-Eberstein et~al.(2023)M{\"u}ller-Eberstein, Van Der~Goot,
  Plank, and Titov}]{muller2023subspace}
Max M{\"u}ller-Eberstein, Rob Van Der~Goot, Barbara Plank, and Ivan Titov.
  2023.
\newblock \href {https://arxiv.org/pdf/2310.16484} {Subspace chronicles: How
  linguistic information emerges, shifts and interacts during language model
  training}.
\newblock In \emph{Findings of the Association for Computational Linguistics:
  EMNLP 2023}, pages 13190--13208.

\bibitem[{Nasr et~al.(2023)Nasr, Carlini, Hayase, Jagielski, Cooper, Ippolito,
  Choquette-Choo, Wallace, Tram{\`e}r, and Lee}]{nasr2023scalable}
Milad Nasr, Nicholas Carlini, Jonathan Hayase, Matthew Jagielski, A~Feder
  Cooper, Daphne Ippolito, Christopher~A Choquette-Choo, Eric Wallace, Florian
  Tram{\`e}r, and Katherine Lee. 2023.
\newblock \href {https://arxiv.org/abs/2311.17035} {Scalable extraction of
  training data from (production) language models}.
\newblock \emph{arXiv preprint arXiv:2311.17035}.

\bibitem[{Niu et~al.(2024)Niu, Liu, Zhu, and Penn}]{Niu2024WhatDT}
Jingcheng Niu, Andrew Liu, Zining Zhu, and Gerald Penn. 2024.
\newblock \href {https://arxiv.org/pdf/2405.02421} {What does the knowledge
  neuron thesis have to do with knowledge?}
\newblock In \emph{International Conference on Learning Representations}.

\bibitem[{Ortu et~al.(2024)Ortu, Jin, Doimo, Sachan, Cazzaniga, and
  Sch{\"o}lkopf}]{ortu2024competition}
Francesco Ortu, Zhijing Jin, Diego Doimo, Mrinmaya Sachan, Alberto Cazzaniga,
  and Bernhard Sch{\"o}lkopf. 2024.
\newblock \href {https://arxiv.org/pdf/2402.11655} {Competition of mechanisms:
  Tracing how language models handle facts and counterfactuals}.
\newblock \emph{arXiv preprint arXiv:2402.11655}.

\bibitem[{Pilehvar and Camacho-Collados(2019)}]{pilehvar2019wic}
Mohammad~Taher Pilehvar and Jose Camacho-Collados. 2019.
\newblock \href {https://arxiv.org/pdf/1808.09121} {{WiC}: the word-in-context
  dataset for evaluating context-sensitive meaning representations}.
\newblock In \emph{Proceedings of the 2019 Conference of the North American
  Chapter of the Association for Computational Linguistics: Human Language
  Technologies, Volume 1 (Long and Short Papers)}, pages 1267--1273.

\bibitem[{Saravia et~al.(2018)Saravia, Liu, Huang, Wu, and
  Chen}]{saravia2018carer}
Elvis Saravia, Hsien-Chi~Toby Liu, Yen-Hao Huang, Junlin Wu, and Yi-Shin Chen.
  2018.
\newblock \href {https://doi.org/10.18653/v1/D18-1404} {{CARER}: Contextualized
  affect representations for emotion recognition}.
\newblock In \emph{Proceedings of the 2018 Conference on Empirical Methods in
  Natural Language Processing}, pages 3687--3697.

\bibitem[{Sharma et~al.(2024)Sharma, Atkinson, and Bau}]{sharma2024locating}
Arnab~Sen Sharma, David Atkinson, and David Bau. 2024.
\newblock \href {http://www.arxiv.org/pdf/2404.03646} {Locating and editing
  factual associations in {M}amba}.
\newblock \emph{arXiv preprint arXiv:2404.03646}.

\bibitem[{Shi et~al.(2023)Shi, Ajith, Xia, Huang, Liu, Blevins, Chen, and
  Zettlemoyer}]{shi2023detecting}
Weijia Shi, Anirudh Ajith, Mengzhou Xia, Yangsibo Huang, Daogao Liu, Terra
  Blevins, Danqi Chen, and Luke Zettlemoyer. 2023.
\newblock \href {https://arxiv.org/abs/2310.16789} {Detecting pretraining data
  from large language models}.
\newblock In \emph{NeurIPS 2023 Workshop on Regulatable ML}.

\bibitem[{Socher et~al.(2013)Socher, Perelygin, Wu, Chuang, Manning, Ng, and
  Potts}]{socher2013recursive}
Richard Socher, Alex Perelygin, Jean Wu, Jason Chuang, Christopher~D Manning,
  Andrew~Y Ng, and Christopher Potts. 2013.
\newblock \href {https://aclanthology.org/D13-1170.pdf} {Recursive deep models
  for semantic compositionality over a sentiment treebank}.
\newblock In \emph{Proceedings of the 2013 conference on empirical methods in
  natural language processing}, pages 1631--1642.

\bibitem[{Stephenson et~al.(2021)Stephenson, Padhy, Ganesh, Hui, Tang, and
  Chung}]{stephenson2021geometry}
Cory Stephenson, Suchismita Padhy, Abhinav Ganesh, Yue Hui, Hanlin Tang, and
  Sue~Yeon Chung. 2021.
\newblock \href
  {https://nyuscholars.nyu.edu/en/publications/on-the-geometry-of-generalization-and-memorization-in-deep-neural}
  {On the geometry of generalization and memorization in deep neural networks}.
\newblock In \emph{International Conference on Learning Representations}.

\bibitem[{Stoehr et~al.(2024)Stoehr, Gordon, Zhang, and
  Lewis}]{stoehr2024localizing}
Niklas Stoehr, Mitchell Gordon, Chiyuan Zhang, and Owen Lewis. 2024.
\newblock \href {https://arxiv.org/pdf/2403.19851} {Localizing paragraph
  memorization in language models}.
\newblock \emph{arXiv preprint arXiv:2403.19851}.

\bibitem[{T{\"a}nzer et~al.(2022)T{\"a}nzer, Ruder, and
  Rei}]{tanzer2022memorisation}
Michael T{\"a}nzer, Sebastian Ruder, and Marek Rei. 2022.
\newblock \href {https://arxiv.org/pdf/2105.00828} {Memorisation versus
  generalisation in pre-trained language models}.
\newblock In \emph{Proceedings of the 60th Annual Meeting of the Association
  for Computational Linguistics (Volume 1: Long Papers)}, pages 7564--7578.

\bibitem[{Tenney et~al.(2019)Tenney, Das, and Pavlick}]{tenney2019bert}
Ian Tenney, Dipanjan Das, and Ellie Pavlick. 2019.
\newblock \href {https://aclanthology.org/P19-1452.pdf} {{BERT} rediscovers the
  classical {NLP} pipeline}.
\newblock In \emph{Proceedings of the 57th Annual Meeting of the Association
  for Computational Linguistics}, pages 4593--4601.

\bibitem[{Wang et~al.(2019)Wang, Pruksachatkun, Nangia, Singh, Michael, Hill,
  Levy, and Bowman}]{wang2019superglue}
Alex Wang, Yada Pruksachatkun, Nikita Nangia, Amanpreet Singh, Julian Michael,
  Felix Hill, Omer Levy, and Samuel Bowman. 2019.
\newblock \href
  {https://proceedings.neurips.cc/paper_files/paper/2019/file/4496bf24afe7fab6f046bf4923da8de6-Paper.pdf}
  {Super{GLUE}: A stickier benchmark for general-purpose language understanding
  systems}.
\newblock \emph{Advances in neural information processing systems}, 32.

\bibitem[{Wang et~al.(2018)Wang, Singh, Michael, Hill, Levy, and
  Bowman}]{wang2018glue}
Alex Wang, Amanpreet Singh, Julian Michael, Felix Hill, Omer Levy, and Samuel
  Bowman. 2018.
\newblock \href {https://arxiv.org/pdf/1804.07461} {{GLUE}: A multi-task
  benchmark and analysis platform for natural language understanding}.
\newblock In \emph{Proceedings of the 2018 EMNLP Workshop BlackboxNLP:
  Analyzing and Interpreting Neural Networks for NLP}. Association for
  Computational Linguistics.

\bibitem[{Warstadt et~al.(2019)Warstadt, Singh, and
  Bowman}]{warstadt2019neural}
Alex Warstadt, Amanpreet Singh, and Samuel~R Bowman. 2019.
\newblock \href
  {https://direct.mit.edu/tacl/article/doi/10.1162/tacl_a_00290/43528} {Neural
  network acceptability judgments}.
\newblock \emph{Transactions of the Association for Computational Linguistics},
  7:625--641.

\bibitem[{Zeng et~al.(2023)Zeng, Li, Ren, Liu, Xu, He, Xing, Wang, Tang, and
  Yin}]{zeng2023exploring}
Shenglai Zeng, Yaxin Li, Jie Ren, Yiding Liu, Han Xu, Pengfei He, Yue Xing,
  Shuaiqiang Wang, Jiliang Tang, and Dawei Yin. 2023.
\newblock \href {https://arxiv.org/abs/2310.06714} {Exploring memorization in
  fine-tuned language models}.
\newblock \emph{arXiv preprint arXiv:2310.06714}.

\bibitem[{Zhan et~al.(2023)Zhan, Fang, Bindu, Gupta, Hashimoto, and
  Kang}]{zhan2023removing}
Qiusi Zhan, Richard Fang, Rohan Bindu, Akul Gupta, Tatsunori Hashimoto, and
  Daniel Kang. 2023.
\newblock \href {https://arxiv.org/pdf/2311.05553} {Removing {RLHF} protections
  in {GPT-4} via fine-tuning}.
\newblock \emph{arXiv preprint arXiv:2311.05553}.

\bibitem[{Zhang et~al.(2023)Zhang, Ippolito, Lee, Jagielski, Tram{\`e}r, and
  Carlini}]{zhang2023counterfactual}
Chiyuan Zhang, Daphne Ippolito, Katherine Lee, Matthew Jagielski, Florian
  Tram{\`e}r, and Nicholas Carlini. 2023.
\newblock \href
  {https://proceedings.neurips.cc/paper_files/paper/2023/file/7bc4f74e35bcfe8cfe43b0a860786d6a-Paper-Conference.pdf}
  {Counterfactual memorization in neural language models}.
\newblock \emph{Advances in Neural Information Processing Systems},
  36:39321--39362.

\bibitem[{Zhang et~al.(2022)Zhang, Roller, Goyal, Artetxe, Chen, Chen, Dewan,
  Diab, Li, Lin et~al.}]{zhang2022opt}
Susan Zhang, Stephen Roller, Naman Goyal, Mikel Artetxe, Moya Chen, Shuohui
  Chen, Christopher Dewan, Mona Diab, Xian Li, Xi~Victoria Lin, et~al. 2022.
\newblock \href
  {https://arxiv.org/pdf/2205.01068.pdf?fbclid=IwAR1_0YiQKgxIsy8unzoLvL9E2OA41_kze-H0YvhoCzIQUp_gk-MR9dUs2ZE}
  {{OPT}: Open pre-trained transformer language models}.
\newblock \emph{arXiv preprint arXiv:2205.01068}.

\bibitem[{Zhao et~al.(2024)Zhao, Yoshinaga, and Oba}]{zhao2024tracing}
Wayne~Xin Zhao, Naoki Yoshinaga, and Daisuke Oba. 2024.
\newblock \href {https://aclanthology.org/2024.eacl-long.127.pdf} {Tracing the
  roots of facts in multilingual language models: Independent, shared, and
  transferred knowledge}.
\newblock In \emph{Proceedings of the 18th Conference of the European Chapter
  of the Association for Computational Linguistics (Volume 1: Long Papers)},
  pages 2088--2102.

\bibitem[{Zheng and Jiang(2022)}]{zheng2022empirical}
Xiaosen Zheng and Jing Jiang. 2022.
\newblock \href {https://arxiv.org/pdf/2203.12171} {An empirical study of
  memorization in {NLP}}.
\newblock In \emph{Proceedings of the 60th Annual Meeting of the Association
  for Computational Linguistics (Volume 1: Long Papers)}, pages 6265--6278.

\end{thebibliography}
\clearpage

\appendix
\onecolumn

\section{Extended results}
\subsection{Main results (\S\ref{sec:results})}
\label{ap:additional_results}
\begin{figure}[!h]\centering
\includegraphics[width=0.5\textwidth]{figures/legend.pdf}
\begin{subfigure}[b]{\textwidth}
    \caption*{\small\hspace{1.5cm}NLU tasks\hspace{1.5cm}Sentiment tasks\hspace{1.2cm}Hate speech tasks\hspace{1.1cm}Topic classification\hspace{1.5cm}}
    \hfill\includegraphics[width=.265\textwidth]{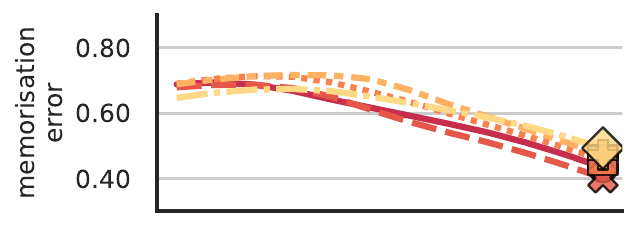}
    \includegraphics[width=.205\textwidth]{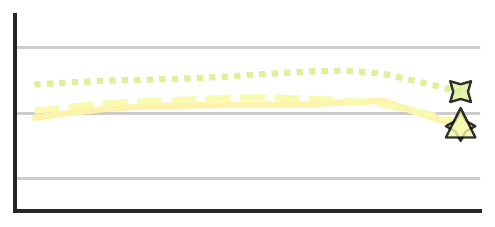}
    \includegraphics[width=.205\textwidth]{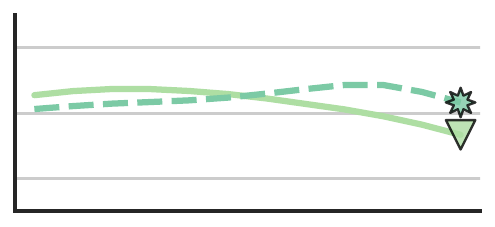}
    \includegraphics[width=.205\textwidth]{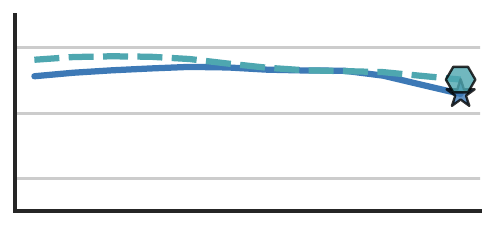}\hspace{.9cm}

    \hfill\includegraphics[width=.262\textwidth]{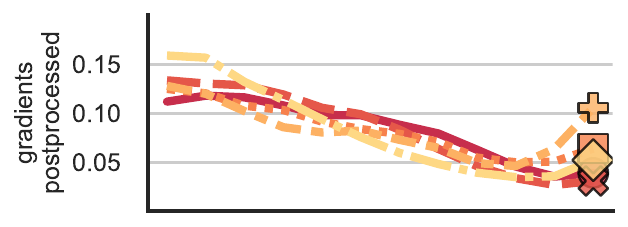}
    \includegraphics[width=.205\textwidth]{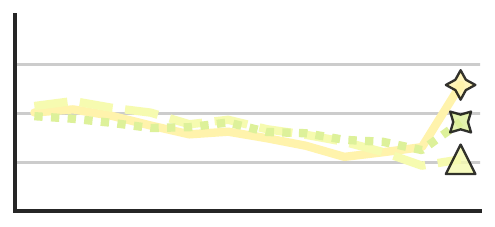}
    \includegraphics[width=.205\textwidth]{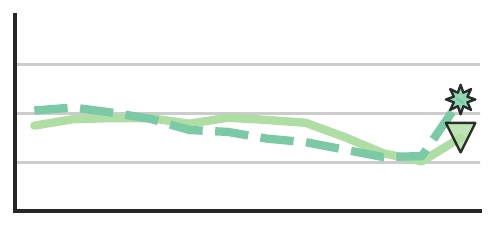}
    \includegraphics[width=.205\textwidth]{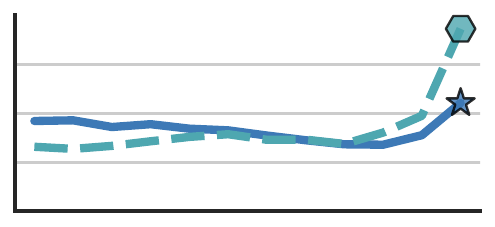}\hspace{.9cm}

    \hfill\includegraphics[width=.255\textwidth]{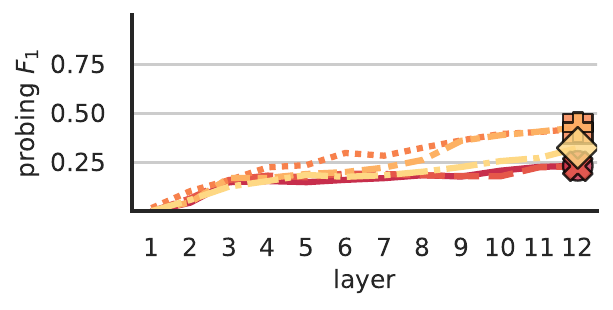}
    \includegraphics[width=.205\textwidth]{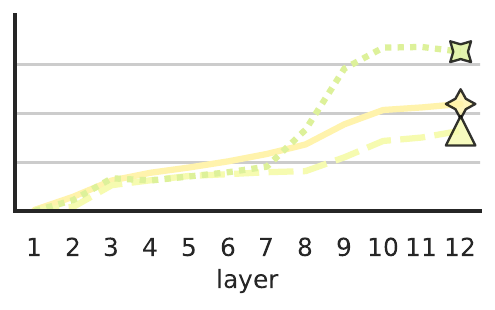}
    \includegraphics[width=.205\textwidth]{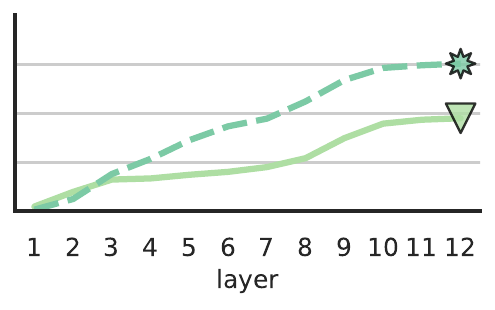}
    \includegraphics[width=.205\textwidth]{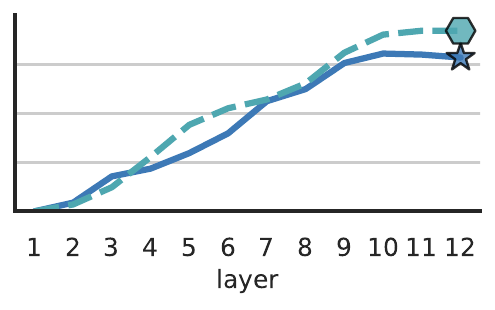}\hspace{0.9cm}
    
    \caption{\texttt{BERT}}\label{fig:main_results_bert}
\end{subfigure}

\begin{subfigure}[b]{\textwidth}
    \hfill\includegraphics[width=.265\textwidth]{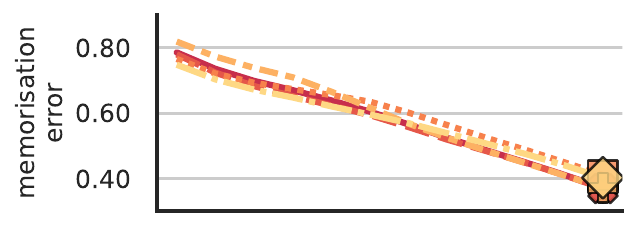}
    \includegraphics[width=.205\textwidth]{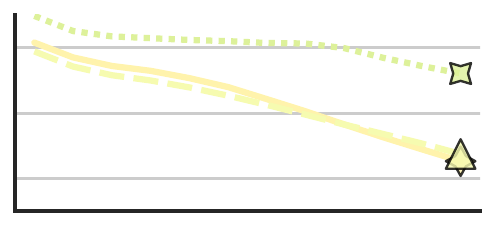}
    \includegraphics[width=.205\textwidth]{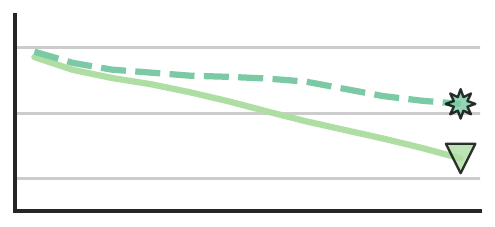}
    \includegraphics[width=.205\textwidth]{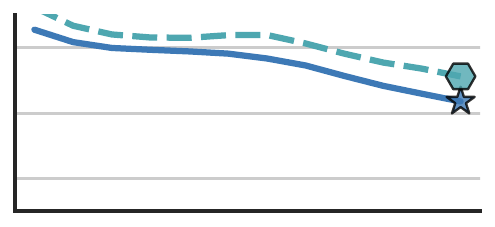}\hspace{.9cm}

    \hfill\includegraphics[width=.262\textwidth]{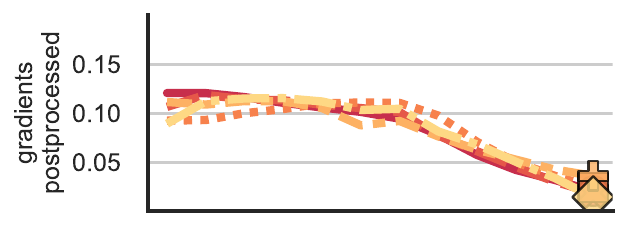}
    \includegraphics[width=.205\textwidth]{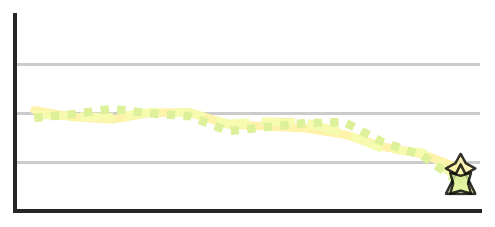}
    \includegraphics[width=.205\textwidth]{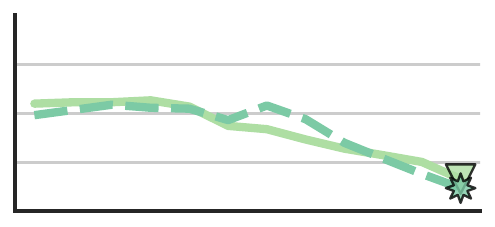}
    \includegraphics[width=.205\textwidth]{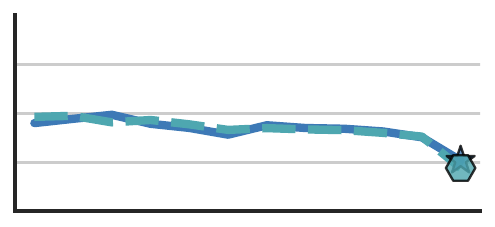}\hspace{.9cm}

    \hfill\includegraphics[width=.255\textwidth]{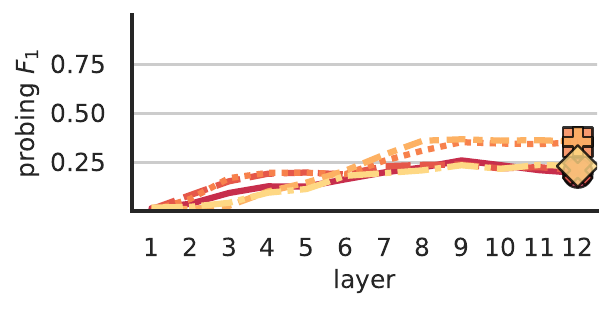}
    \includegraphics[width=.205\textwidth]{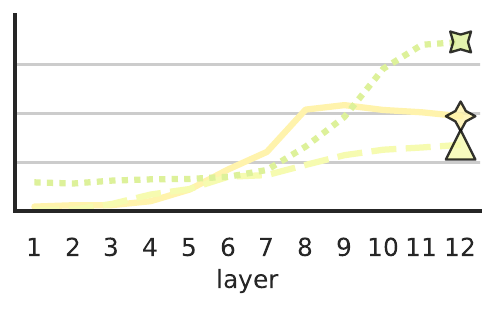}
    \includegraphics[width=.205\textwidth]{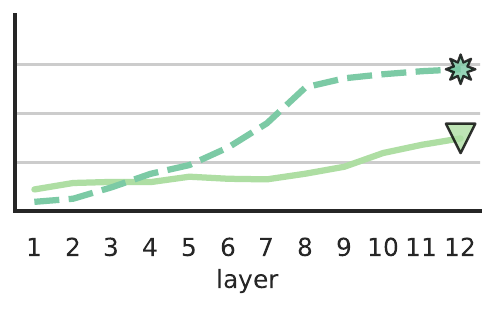}
    \includegraphics[width=.205\textwidth]{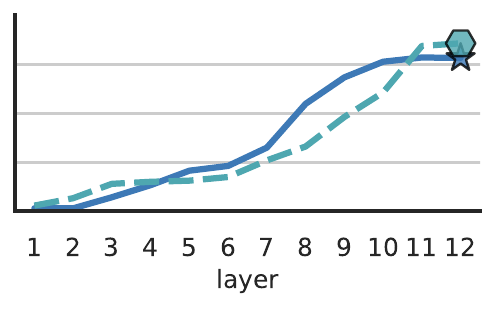}\hspace{0.9cm}
    
    \caption{\texttt{GPT-N}}\label{fig:main_results_gptn}
\end{subfigure}

\begin{subfigure}[b]{\textwidth}
    \hfill\includegraphics[width=.265\textwidth]{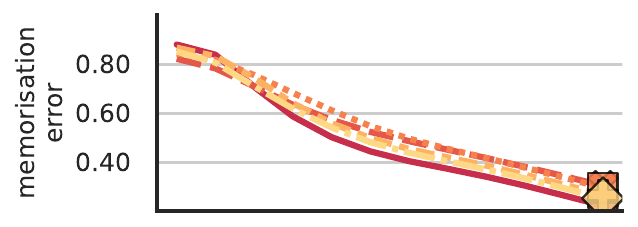}
    \includegraphics[width=.205\textwidth]{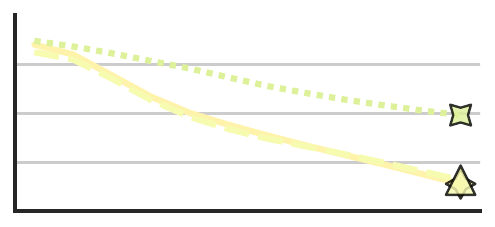}
    \includegraphics[width=.205\textwidth]{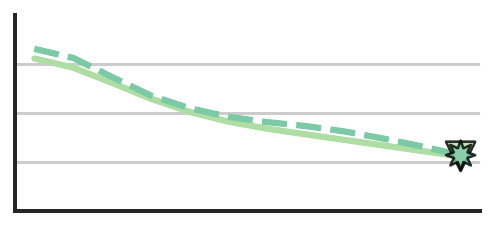}
    \includegraphics[width=.205\textwidth]{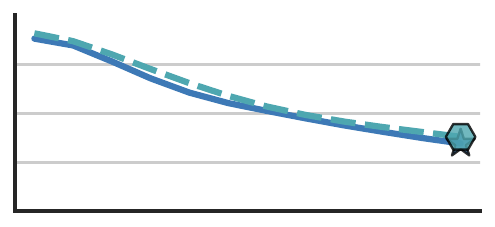}\hspace{.9cm}

    \hfill\includegraphics[width=.262\textwidth]{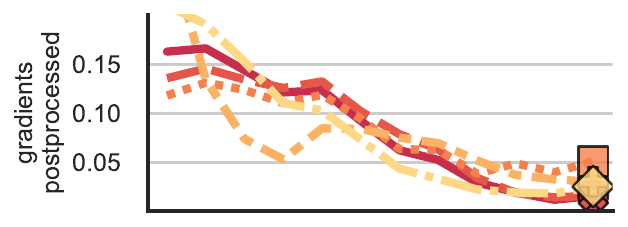}
    \includegraphics[width=.205\textwidth]{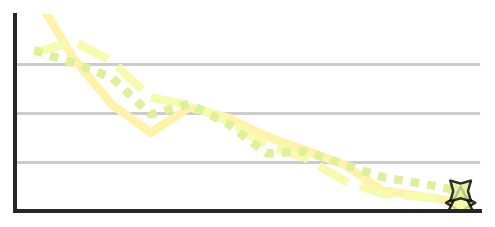}
    \includegraphics[width=.205\textwidth]{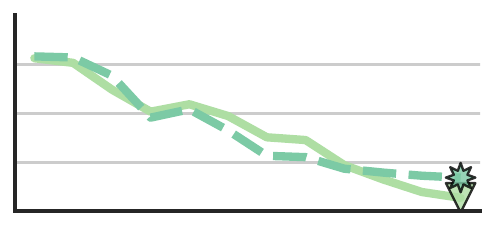}
    \includegraphics[width=.205\textwidth]{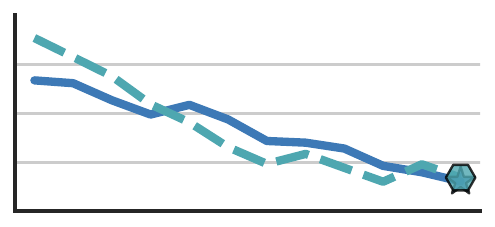}\hspace{.9cm}

    \hfill\includegraphics[width=.255\textwidth]{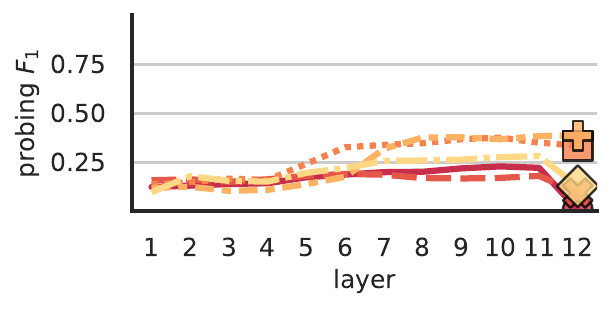}
    \includegraphics[width=.205\textwidth]{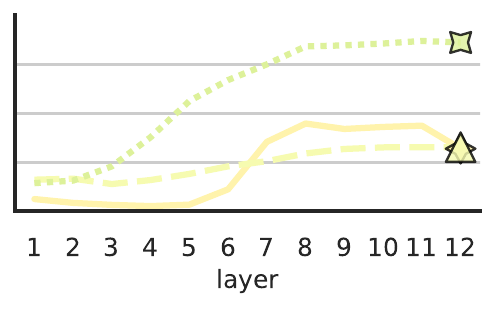}
    \includegraphics[width=.205\textwidth]{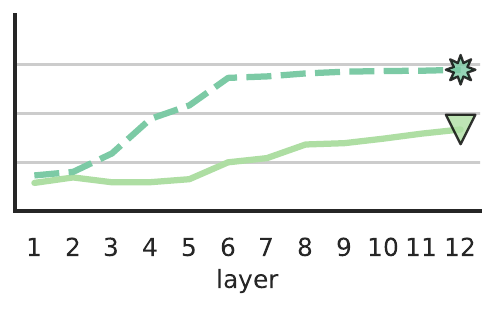}
    \includegraphics[width=.205\textwidth]{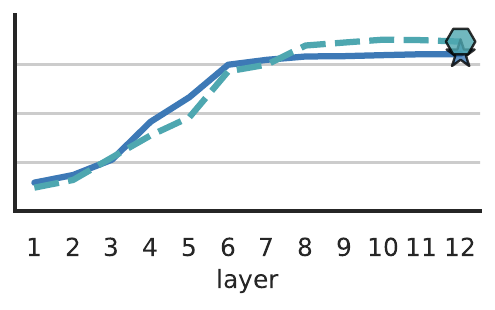}\hspace{0.9cm}
    
    \caption{\texttt{Pythia}}\label{fig:main_results_pythia}
\end{subfigure}
\caption{Memorisation localisation techniques: (1) the top row provides the memorisation error when swapping layers (inserting non-memorisation layers into a memorisation model), higher numbers suggest higher relevance. (2) the middle row indicates (postprocessed) gradient norms, higher numbers suggest higher relevance. (3) the bottom row provides probing $F_1$-scores when training probes to predict whether an example is a noisy one, where the increase between layers suggests layer relevance.}

\end{figure}

We provide the same visualisation of results as shown for \texttt{OPT} in \S\ref{sec:results} in Figures~\ref{fig:main_results_bert}, for \texttt{BERT}, \ref{fig:main_results_gptn}, for \texttt{GPT-N}, and in \ref{fig:main_results_pythia} for \texttt{Pythia}. We omit layer retraining, that correlates very strongly with layer swapping (shown in the top rows of each subfigure). 

\clearpage
\subsection{Probing to consolidate centroid analysis (\S\ref{sec:centroid_analysis})}
\label{ap:probing}
\begin{figure}[h]
    \centering
    \begin{subfigure}[b]{0.195\columnwidth}
            \includegraphics[width=0.97\textwidth]{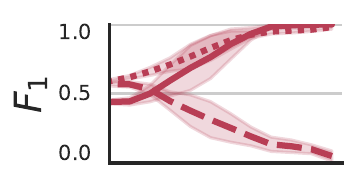}
            \caption{\wic}
    \end{subfigure}
    \begin{subfigure}[b]{0.145\columnwidth}
            \includegraphics[width=\textwidth]{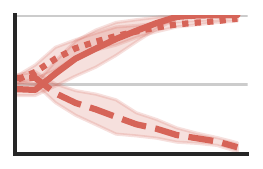}
            \caption{\rte}
    \end{subfigure}
    \begin{subfigure}[b]{0.145\columnwidth}
            \includegraphics[width=\textwidth]{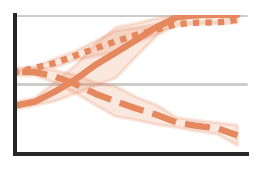}
            \caption{\mrpc}
    \end{subfigure}
    \begin{subfigure}[b]{0.145\columnwidth}
            \includegraphics[width=\textwidth]{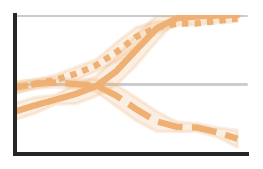}
            \caption{\cola}
    \end{subfigure}
    \begin{subfigure}[b]{0.145\columnwidth}
            \includegraphics[width=\textwidth]{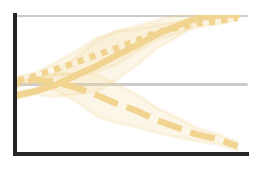}
            \caption{\boolq}
    \end{subfigure}
    \begin{subfigure}[b]{0.145\columnwidth}
            \includegraphics[width=\textwidth]{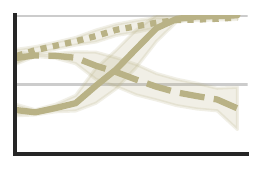}
            \caption{\sstt}
    \end{subfigure}
    
    \begin{subfigure}[b]{0.195\columnwidth}
            \includegraphics[width=0.97\textwidth]{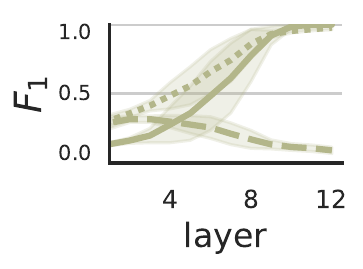}
            \caption{\sstf}
    \end{subfigure}
    \begin{subfigure}[b]{0.145\columnwidth}
            \includegraphics[width=\textwidth]{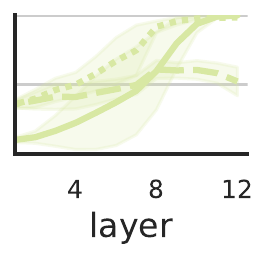}
            \caption{\emotion}
    \end{subfigure}
    \begin{subfigure}[b]{0.145\columnwidth}
            \includegraphics[width=\textwidth]{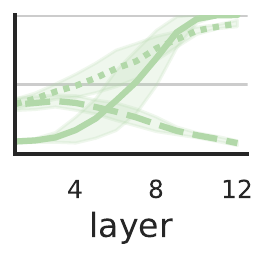}
            \caption{\textcolor{black!10!ih}{\bf\texttt{ImplicitH.}\textsubscript{\includegraphics[width=.2cm]{figures/symbols/symbol_implicithate.pdf}}}}
    \end{subfigure}
    \begin{subfigure}[b]{0.145\columnwidth}
            \includegraphics[width=\textwidth]{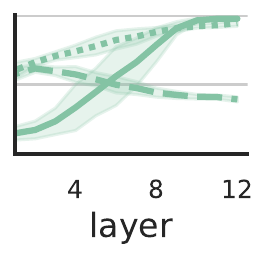}
            \caption{\stormfront}
    \end{subfigure}
    \begin{subfigure}[b]{0.145\columnwidth}
            \includegraphics[width=\textwidth]{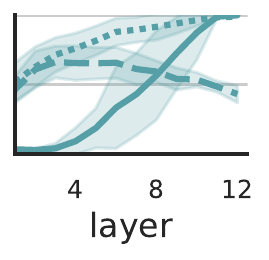}
            \caption{\reuters}
    \end{subfigure}
    \begin{subfigure}[b]{0.145\columnwidth}
            \includegraphics[width=\textwidth]{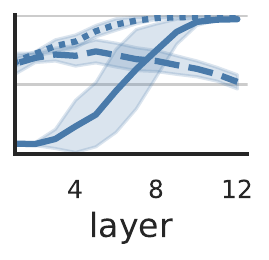}
            \caption{\trec}
    \end{subfigure}
    \caption{We train probes to predict the noisy label (solid line, shown for noisy examples) or the original label (dashed line for noisy examples, dotted line for clean examples).}
    \label{fig:probes_cleanvsnoisy}
\end{figure}

In \S\ref{sec:centroid_analysis}, we trained probes to predict an example's class based on its hidden state, using either original or perturbed labels. Figure~\ref{fig:probes_cleanvsnoisy} shows a) test $F_1$-scores of the noisy examples for the original label (dashed line), b) test $F_1$-scores of the noisy examples for the perturbed label (solid line), and c) the performance on clean examples when training with the original labels (dotted line). Tasks vary widely in terms of when the $F_1$-score for noisy labels exceeds that of the original labels. This happens early on for \wic\ and \rte, but for other tasks (e.g.\ \sstt, \trec\ and \reuters), the probe is better at predicting the original label before it can predict the noisy one. We summarised this in the main paper using the `memorisation >> generalisation' event. 

\subsection{Centroid analysis (\S\ref{sec:centroid_analysis})}
\label{ap:centroid_analysis}

In \S\ref{sec:centroid_analysis} we introduced the centroid analysis as a way to visualise what happens to noisy examples as they move through the different processing layers. 
Figure~\ref{fig:all_centroids} now includes visualisations for all four models, and for \texttt{OPT} with either the bottom or the top six layers swapped with those from $\theta_O$.
Visual inspection leads to the following observations:
\begin{itemize}
    \item \textbf{Gradual vs. relatively localised tasks:} For nearly all setups, the noisy examples gradually move from one centroid to another, confirming that memorisation is a process in which many layers are involved. Still, there are relative differences to be observed, e.g. compare \texttt{TREC} (gradual) to \texttt{MRPC} (relatively localised) for \texttt{Pythia}, or compare \texttt{RTE} (relatively localised) to \texttt{ImplicitHate} (gradual) for \texttt{BERT}.
    \item \textbf{Classification initiation task variation:} Tasks do not always have a consistent classification initiation across models: some tasks are relatively stable in terms of when the two distributions of $a$ and $b$ stop overlapping (e.g. \texttt{WiC}), others show great variation across the four models (e.g. \texttt{Reuters}, \texttt{Stormfront}, \texttt{Emotion}).
    \item \textbf{Swapping bottom layers most successful:} Inspecting the swapping centroid visualisations (final two columns) demonstrates that swapping out the bottom six layers of a memorisation model with $\theta_O$ can prevent the `crossing' (see \S\ref{sec:centroid_analysis}) from happening. Swapping out the top six layers, on the other hand, is less successful since, for some datasets, the memorised examples have already `crossed' (e.g.\ see \texttt{WiC}, \texttt{RTE} and \texttt{MRPC}).
    \item \textbf{Centroid analysis is a simplification:} There are a few dataset-model combinations that show that the centroid analysis is not always a straightforward way to explain the model's behaviour. For instance, for \texttt{Pythia}, \texttt{CoLA} leads to unintuitive results, mostly due to the fact that the two centroids nearly overlap, making the relative distance between them less meaningful in the early layers.
\end{itemize}

\begin{table}[!h]\small\centering
\setlength{\tabcolsep}{0.5pt}
\begin{tabular}{lcccccccc}
     & \texttt{BERT} & \texttt{GPT-N} & \texttt{Pythia} & \texttt{OPT} & \texttt{OPT}, bottom swap & \texttt{OPT}, top swap \\
    \texttt{WiC} & 
    \raisebox{-0.4\totalheight}{\includegraphics[width=0.15\textwidth]{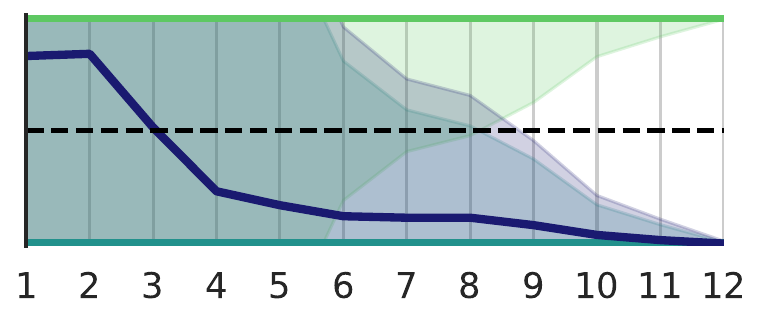}} &
    \raisebox{-0.4\totalheight}{\includegraphics[width=0.15\textwidth]{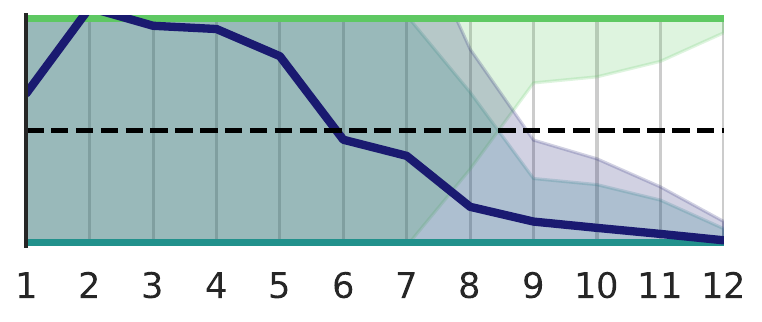}} &
    \raisebox{-0.4\totalheight}{\includegraphics[width=0.15\textwidth]{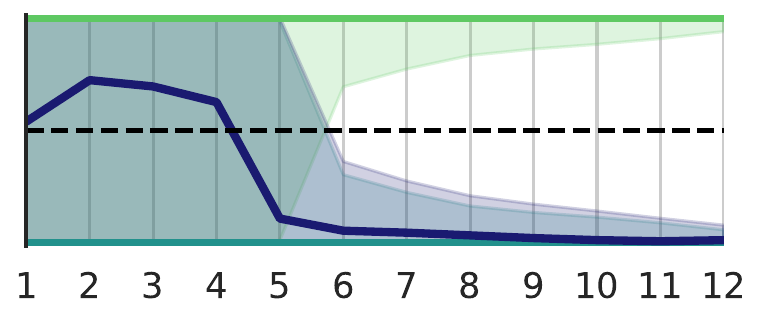}} &
    \raisebox{-0.4\totalheight}{\includegraphics[width=0.15\textwidth]{appendix_figures/centroid_analysis/wic_OPT.pdf}} &
    \raisebox{-0.4\totalheight}{\includegraphics[width=0.15\textwidth]{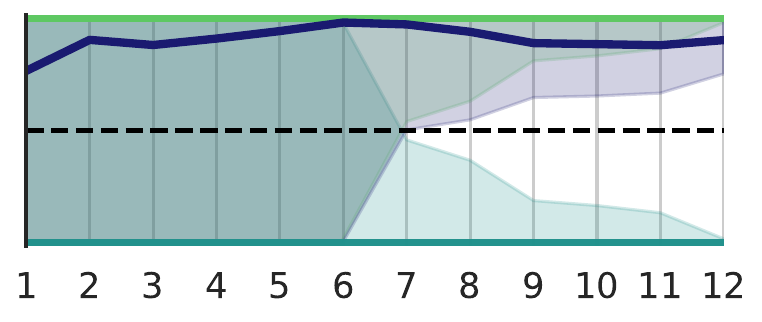}} &
    \raisebox{-0.4\totalheight}{\includegraphics[width=0.15\textwidth]{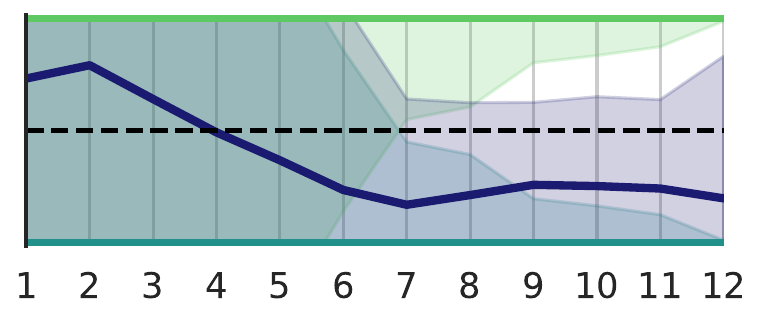}} \\
    \texttt{RTE} & 
    \raisebox{-0.4\totalheight}{\includegraphics[width=0.15\textwidth]{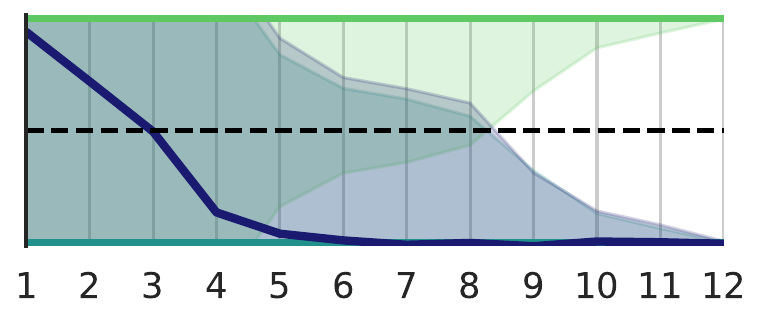}} &
    \raisebox{-0.4\totalheight}{\includegraphics[width=0.15\textwidth]{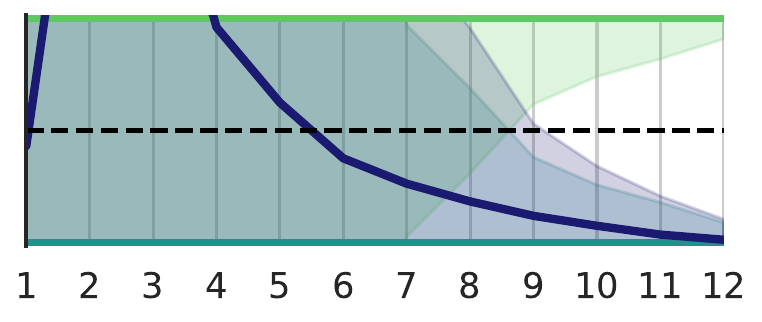}} &
    \raisebox{-0.4\totalheight}{\includegraphics[width=0.15\textwidth]{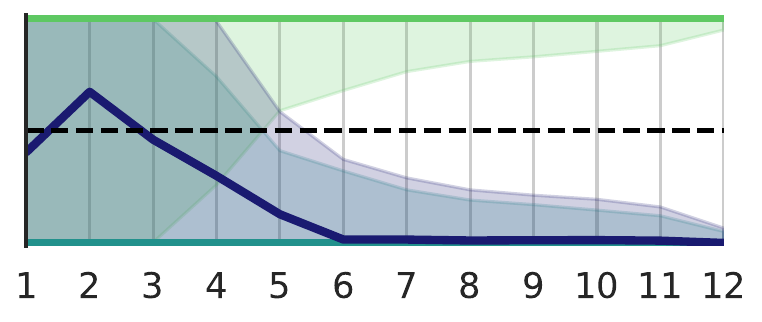}} &
    \raisebox{-0.4\totalheight}{\includegraphics[width=0.15\textwidth]{appendix_figures/centroid_analysis/rte_OPT.pdf}} &
    \raisebox{-0.4\totalheight}{\includegraphics[width=0.15\textwidth]{appendix_figures/centroid_analysis/rte_OPT_mixb.pdf}} &
    \raisebox{-0.4\totalheight}{\includegraphics[width=0.15\textwidth]{appendix_figures/centroid_analysis/rte_OPT_mixt.pdf}} \\
    \texttt{CoLA} & 
    \raisebox{-0.4\totalheight}{\includegraphics[width=0.15\textwidth]{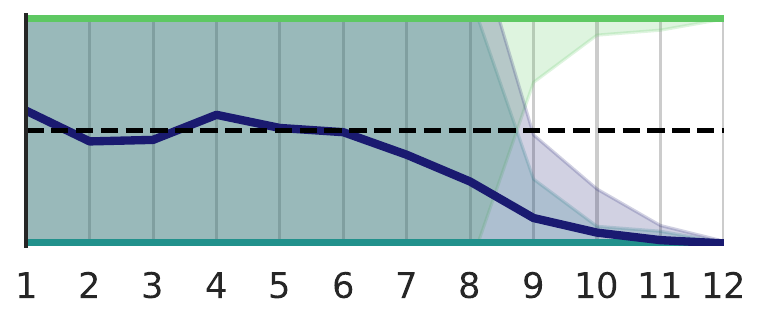}} &
    \raisebox{-0.4\totalheight}{\includegraphics[width=0.15\textwidth]{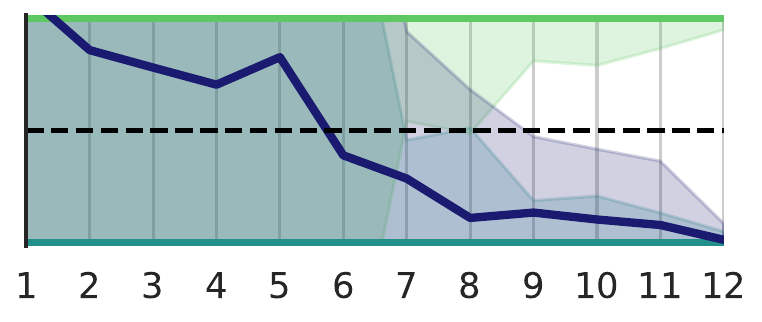}} &
    \raisebox{-0.4\totalheight}{\includegraphics[width=0.15\textwidth]{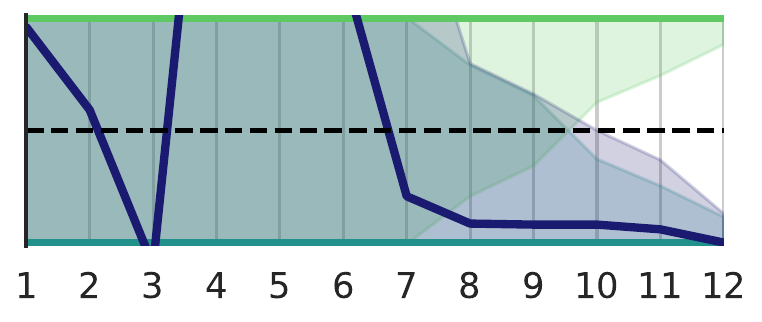}} &
    \raisebox{-0.4\totalheight}{\includegraphics[width=0.15\textwidth]{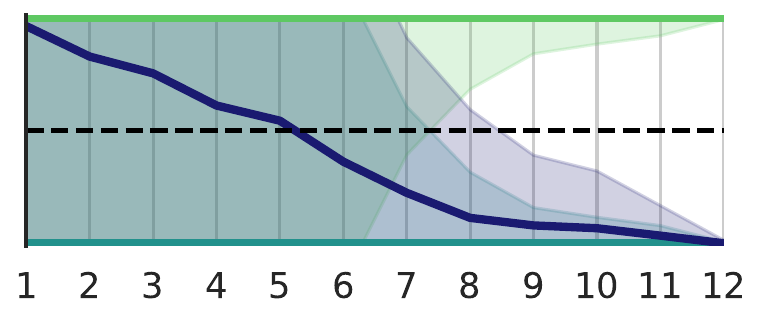}} &
    \raisebox{-0.4\totalheight}{\includegraphics[width=0.15\textwidth]{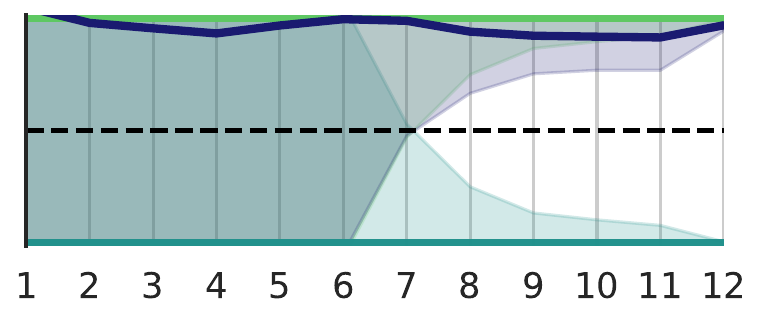}} &
    \raisebox{-0.4\totalheight}{\includegraphics[width=0.15\textwidth]{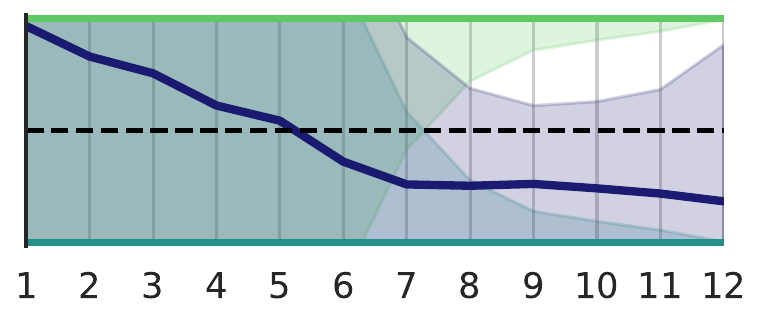}} \\
    \texttt{BoolQ} & 
    \raisebox{-0.4\totalheight}{\includegraphics[width=0.15\textwidth]{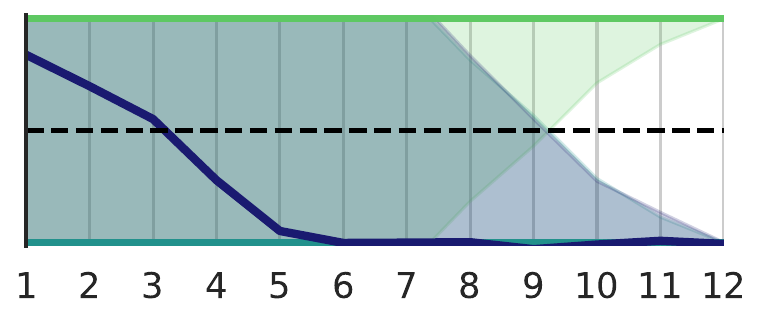}} &
    \raisebox{-0.4\totalheight}{\includegraphics[width=0.15\textwidth]{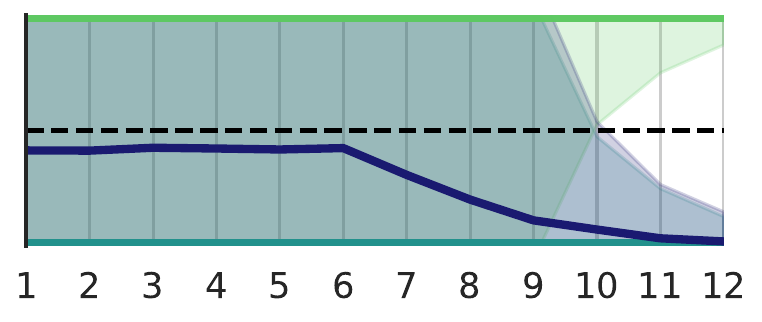}} &
    \raisebox{-0.4\totalheight}{\includegraphics[width=0.15\textwidth]{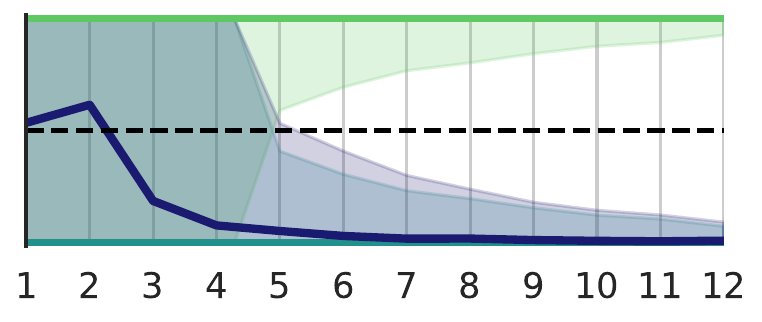}} &
    \raisebox{-0.4\totalheight}{\includegraphics[width=0.15\textwidth]{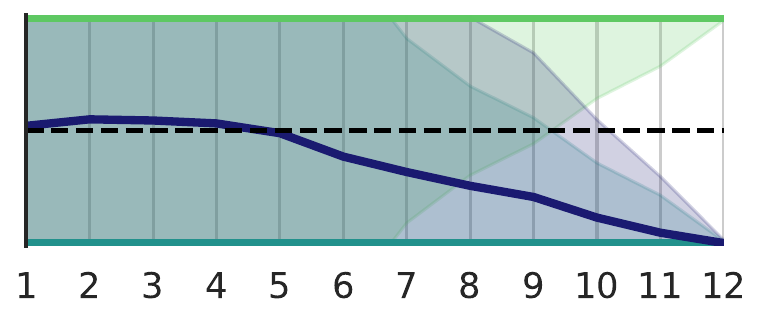}} &
    \raisebox{-0.4\totalheight}{\includegraphics[width=0.15\textwidth]{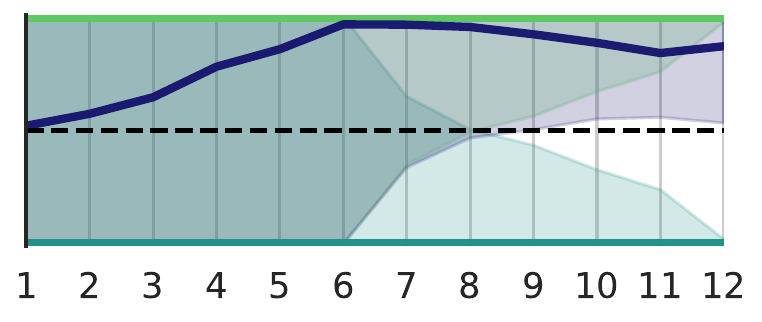}} &
    \raisebox{-0.4\totalheight}{\includegraphics[width=0.15\textwidth]{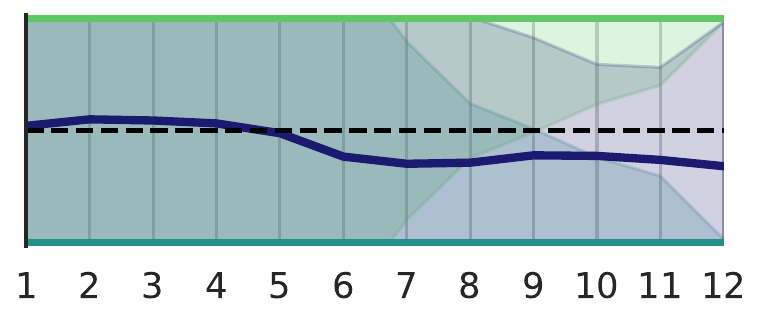}} \\
    \texttt{MRPC} & 
    \raisebox{-0.4\totalheight}{\includegraphics[width=0.15\textwidth]{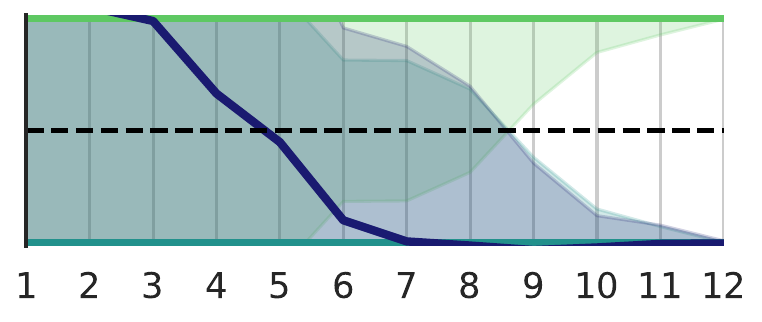}} &
    \raisebox{-0.4\totalheight}{\includegraphics[width=0.15\textwidth]{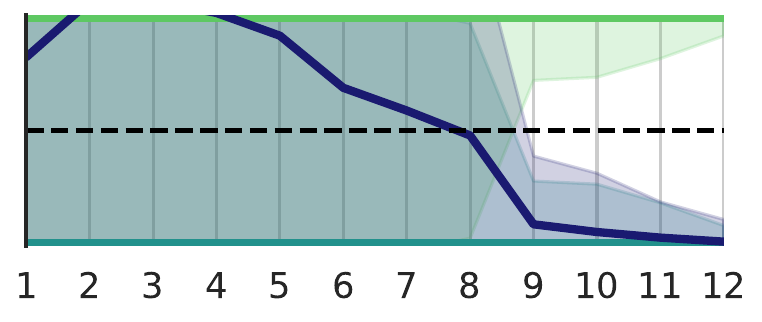}} &
    \raisebox{-0.4\totalheight}{\includegraphics[width=0.15\textwidth]{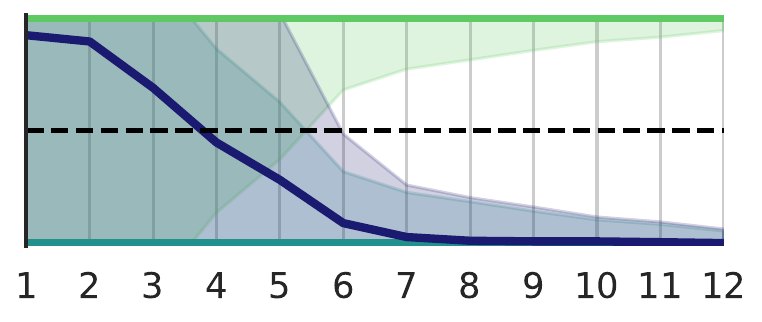}} &
    \raisebox{-0.4\totalheight}{\includegraphics[width=0.15\textwidth]{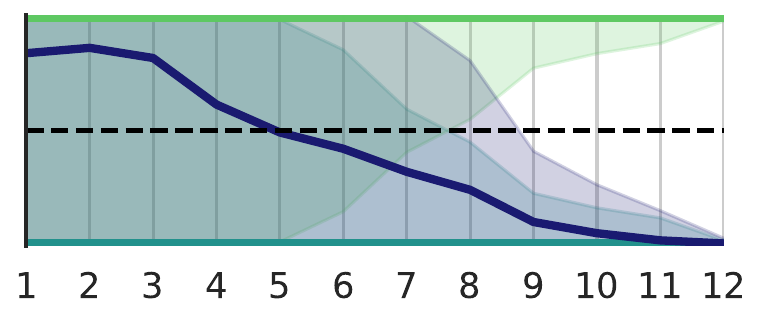}} &
    \raisebox{-0.4\totalheight}{\includegraphics[width=0.15\textwidth]{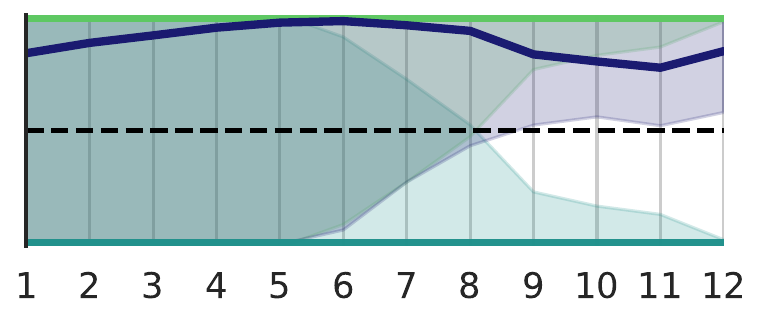}} &
    \raisebox{-0.4\totalheight}{\includegraphics[width=0.15\textwidth]{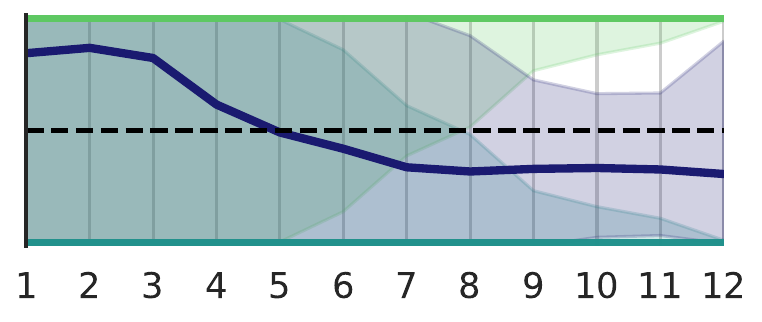}} \\
    \texttt{SST-2} & 
    \raisebox{-0.4\totalheight}{\includegraphics[width=0.15\textwidth]{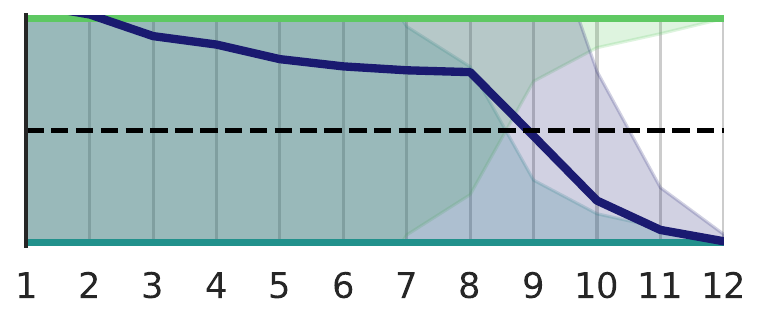}} &
    \raisebox{-0.4\totalheight}{\includegraphics[width=0.15\textwidth]{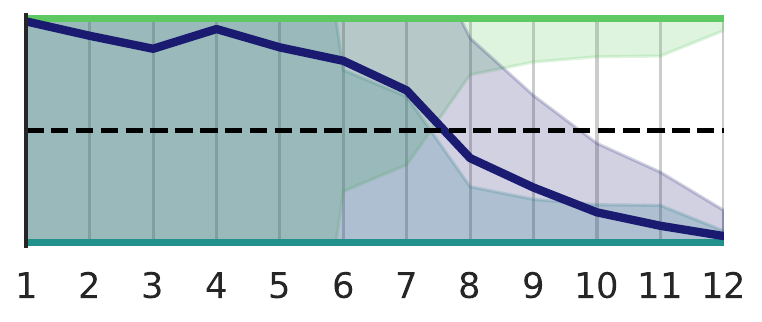}} &
    \raisebox{-0.4\totalheight}{\includegraphics[width=0.15\textwidth]{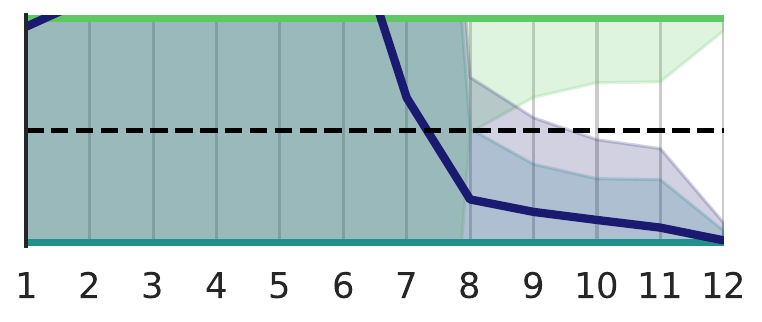}} &
    \raisebox{-0.4\totalheight}{\includegraphics[width=0.15\textwidth]{appendix_figures/centroid_analysis/sst2_OPT.pdf}} &
    \raisebox{-0.4\totalheight}{\includegraphics[width=0.15\textwidth]{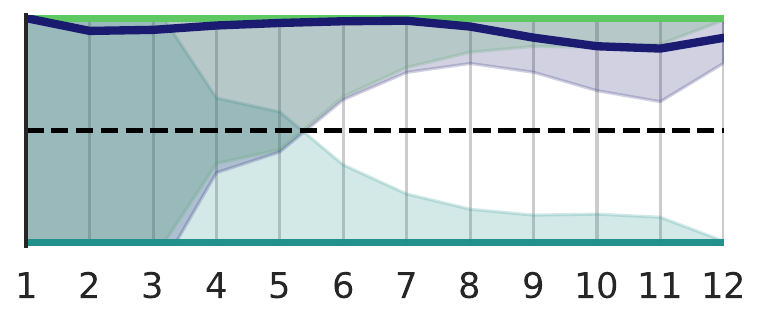}} &
    \raisebox{-0.4\totalheight}{\includegraphics[width=0.15\textwidth]{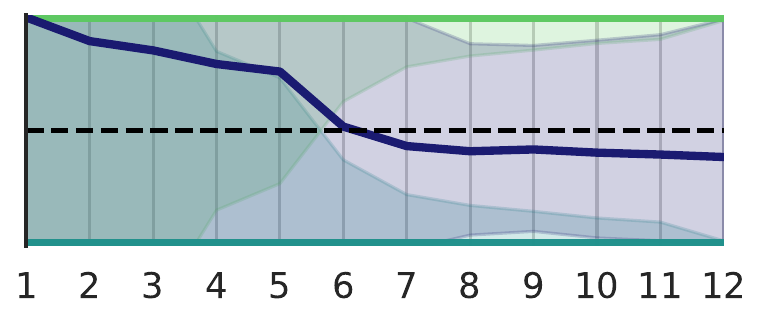}} \\
    \texttt{SST-5} & 
    \raisebox{-0.4\totalheight}{\includegraphics[width=0.15\textwidth]{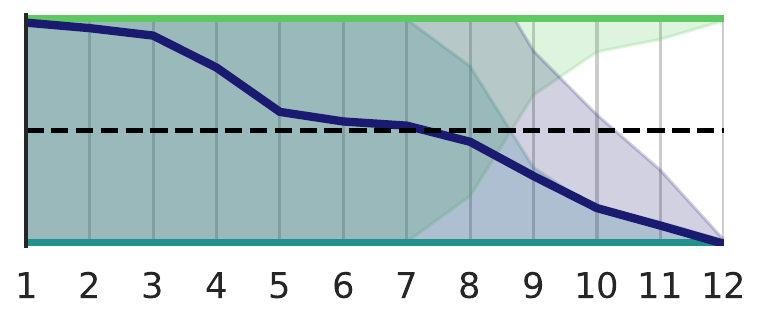}} &
    \raisebox{-0.4\totalheight}{\includegraphics[width=0.15\textwidth]{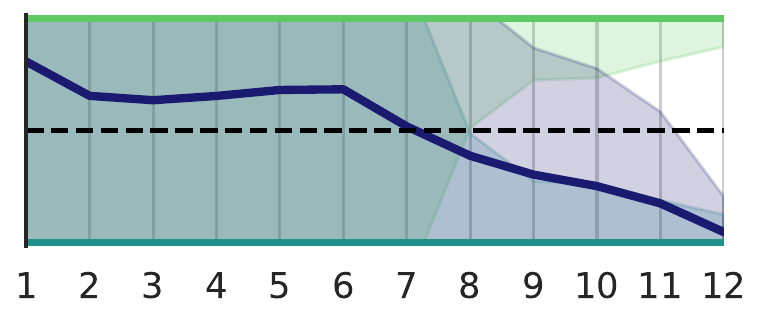}} &
    \raisebox{-0.4\totalheight}{\includegraphics[width=0.15\textwidth]{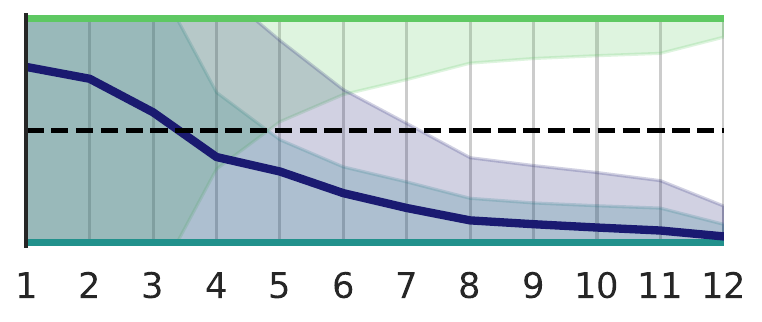}} &
    \raisebox{-0.4\totalheight}{\includegraphics[width=0.15\textwidth]{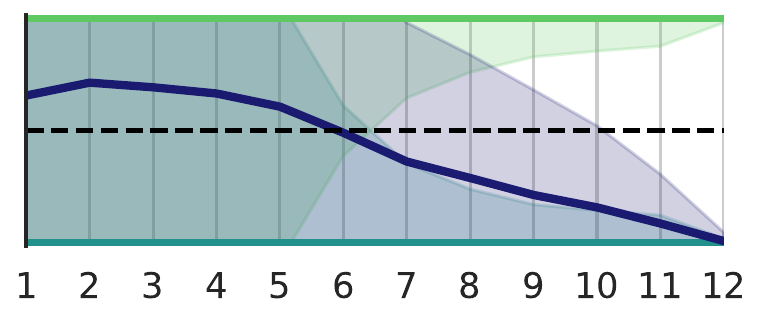}} &
    \raisebox{-0.4\totalheight}{\includegraphics[width=0.15\textwidth]{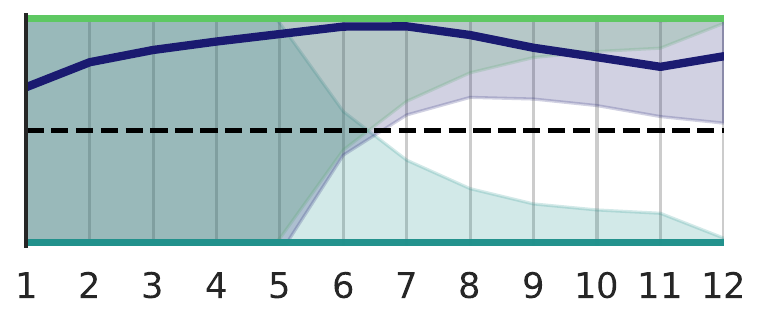}} &
    \raisebox{-0.4\totalheight}{\includegraphics[width=0.15\textwidth]{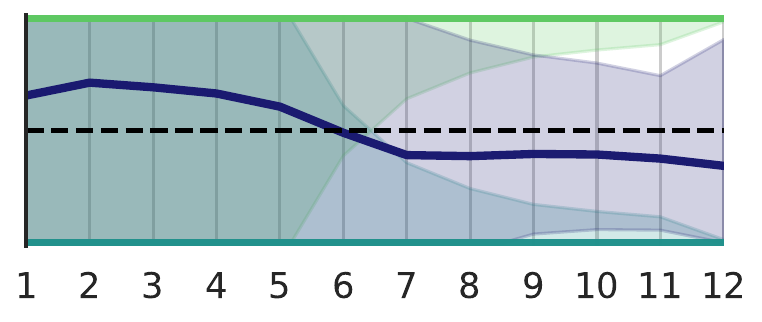}} \\
    \texttt{Emotion} & 
    \raisebox{-0.4\totalheight}{\includegraphics[width=0.15\textwidth]{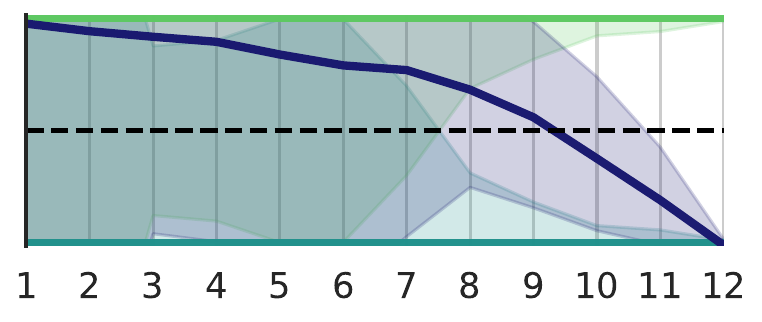}} &
    \raisebox{-0.4\totalheight}{\includegraphics[width=0.15\textwidth]{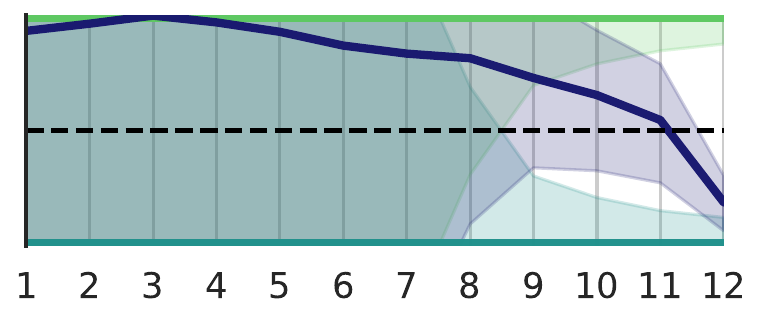}} &
    \raisebox{-0.4\totalheight}{\includegraphics[width=0.15\textwidth]{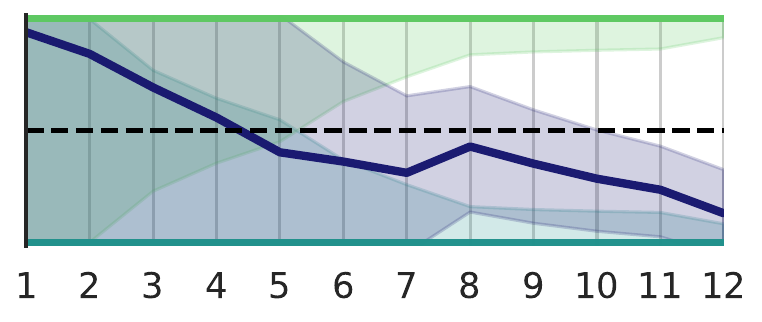}} &
    \raisebox{-0.4\totalheight}{\includegraphics[width=0.15\textwidth]{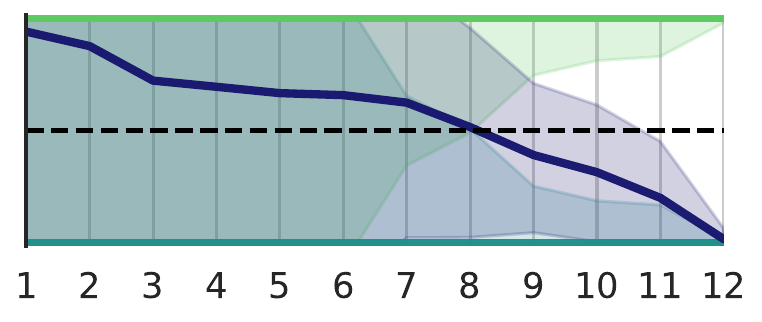}} &
    \raisebox{-0.4\totalheight}{\includegraphics[width=0.15\textwidth]{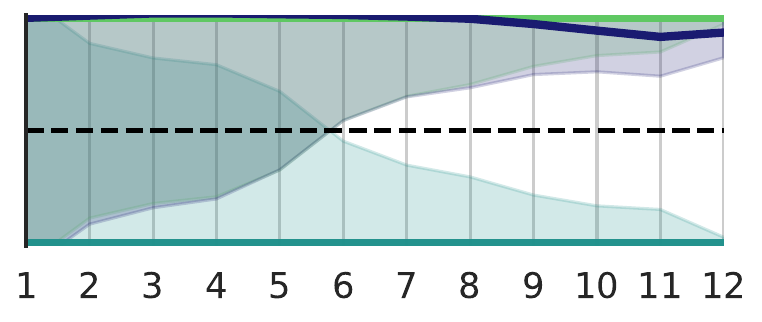}} &
    \raisebox{-0.4\totalheight}{\includegraphics[width=0.15\textwidth]{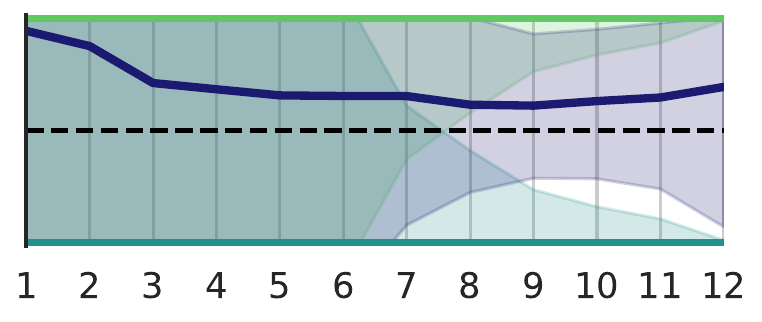}} \\
    \texttt{Stormf.} & 
    \raisebox{-0.4\totalheight}{\includegraphics[width=0.15\textwidth]{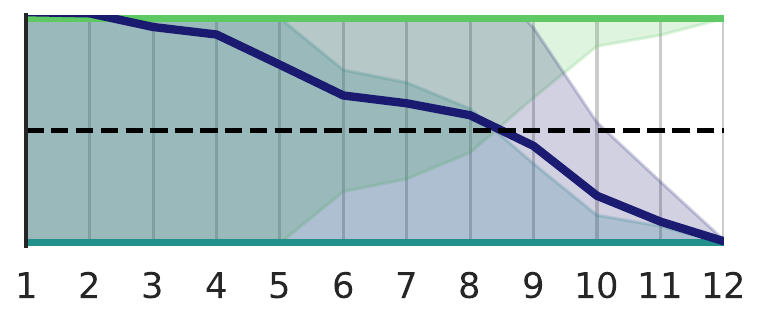}} &
    \raisebox{-0.4\totalheight}{\includegraphics[width=0.15\textwidth]{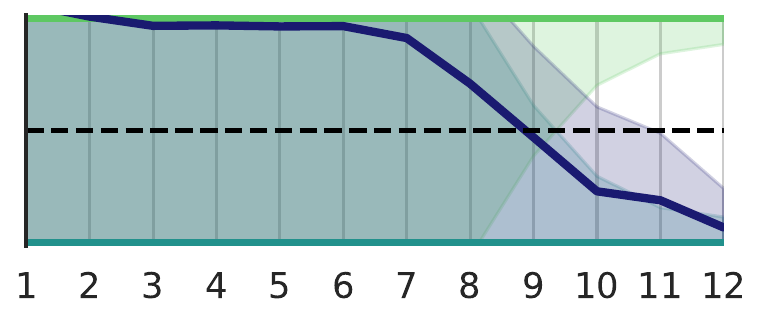}} &
    \raisebox{-0.4\totalheight}{\includegraphics[width=0.15\textwidth]{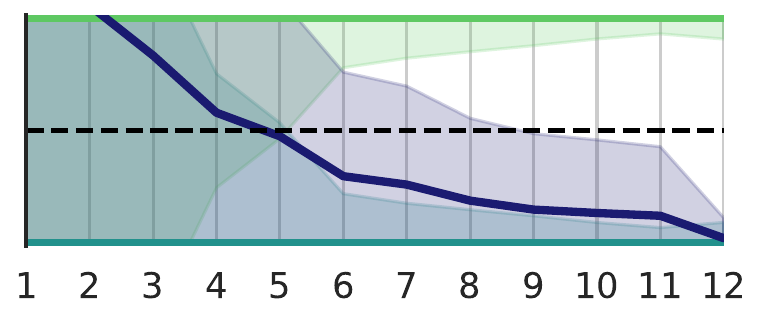}} &
    \raisebox{-0.4\totalheight}{\includegraphics[width=0.15\textwidth]{appendix_figures/centroid_analysis/stormfront_OPT.pdf}} &
    \raisebox{-0.4\totalheight}{\includegraphics[width=0.15\textwidth]{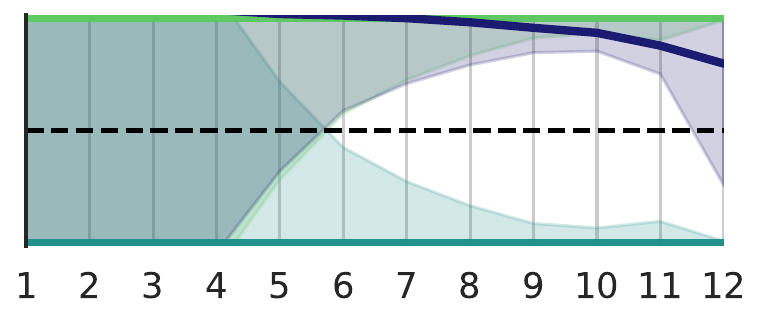}} &
    \raisebox{-0.4\totalheight}{\includegraphics[width=0.15\textwidth]{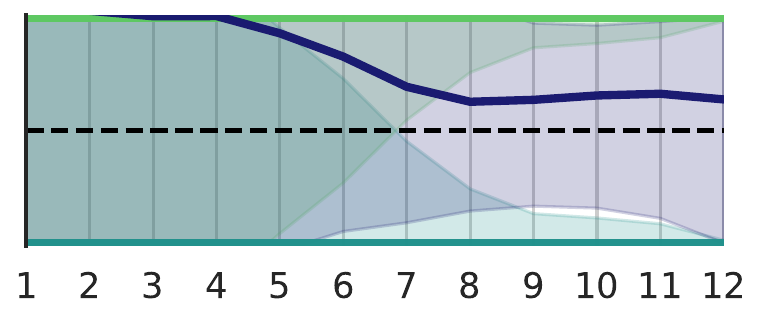}} \\ 
    \texttt{IH} & 
    \raisebox{-0.4\totalheight}{\includegraphics[width=0.15\textwidth]{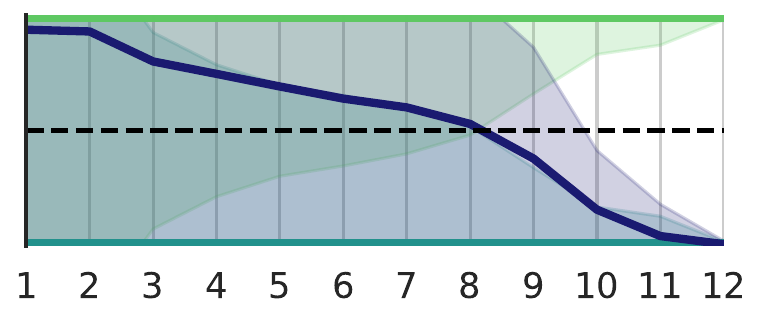}} &
    \raisebox{-0.4\totalheight}{\includegraphics[width=0.15\textwidth]{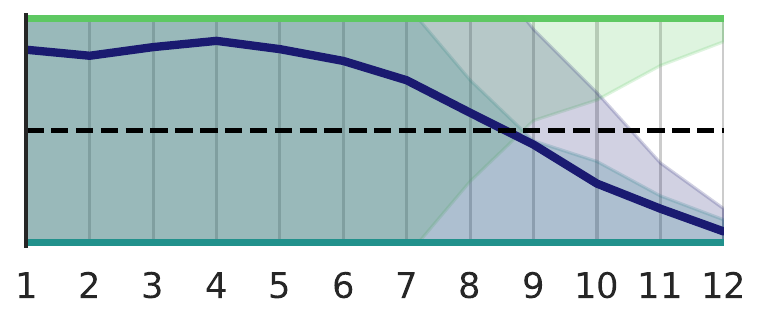}} &
    \raisebox{-0.4\totalheight}{\includegraphics[width=0.15\textwidth]{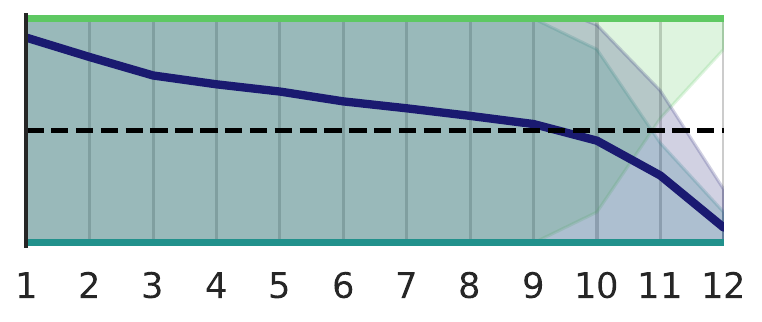}} &
    \raisebox{-0.4\totalheight}{\includegraphics[width=0.15\textwidth]{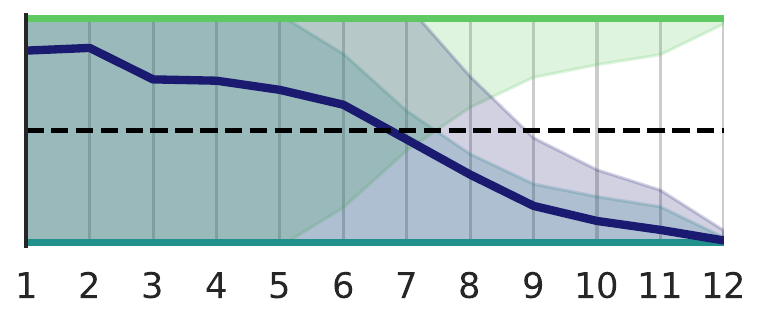}} &
    \raisebox{-0.4\totalheight}{\includegraphics[width=0.15\textwidth]{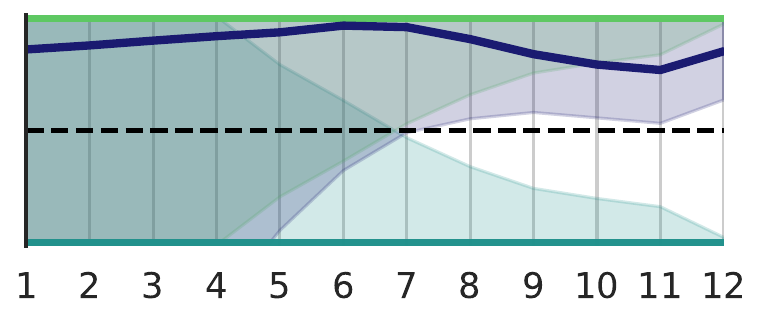}} &
    \raisebox{-0.4\totalheight}{\includegraphics[width=0.15\textwidth]{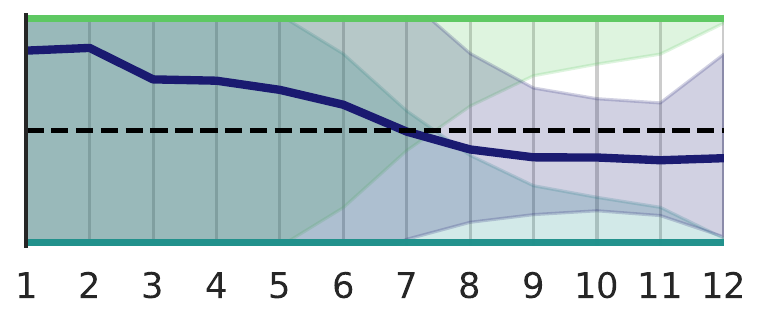}} \\ 
    \texttt{Reuters} & 
    \raisebox{-0.4\totalheight}{\includegraphics[width=0.15\textwidth]{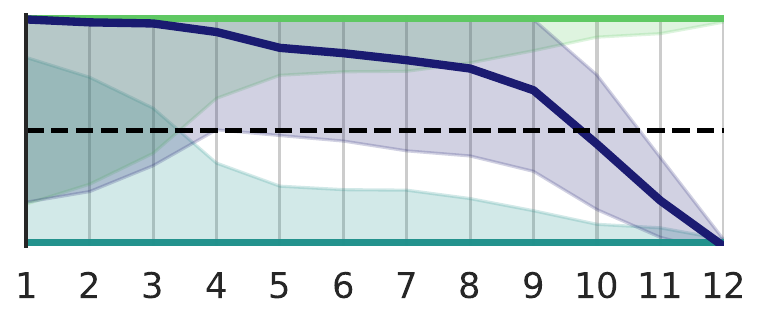}} &
    \raisebox{-0.4\totalheight}{\includegraphics[width=0.15\textwidth]{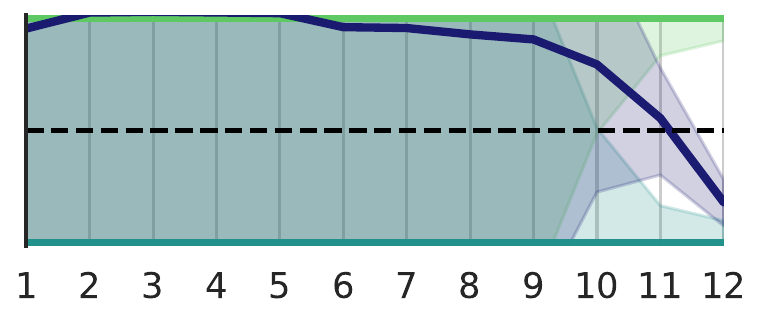}} &
    \raisebox{-0.4\totalheight}{\includegraphics[width=0.15\textwidth]{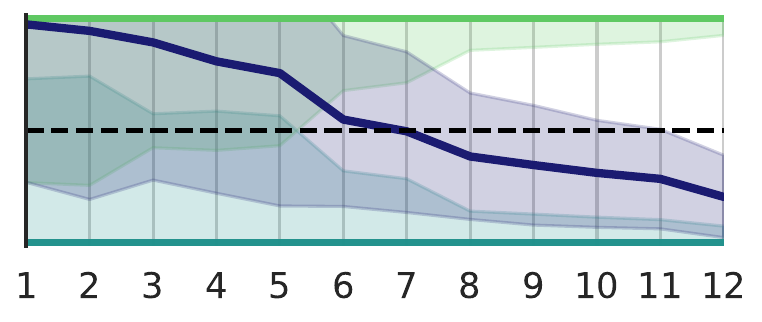}} &
    \raisebox{-0.4\totalheight}{\includegraphics[width=0.15\textwidth]{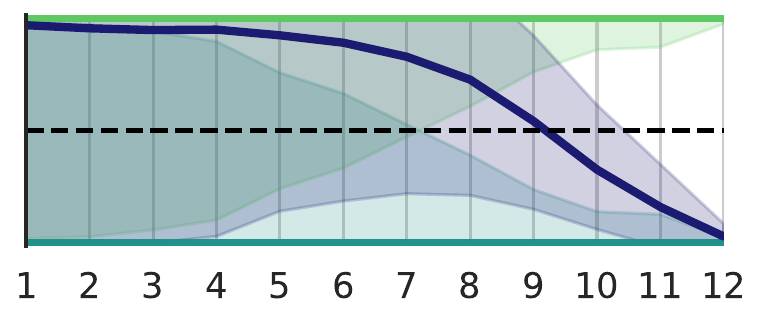}} &
    \raisebox{-0.4\totalheight}{\includegraphics[width=0.15\textwidth]{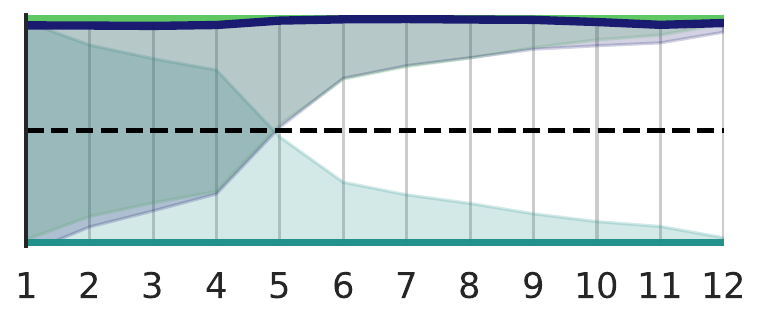}} &
    \raisebox{-0.4\totalheight}{\includegraphics[width=0.15\textwidth]{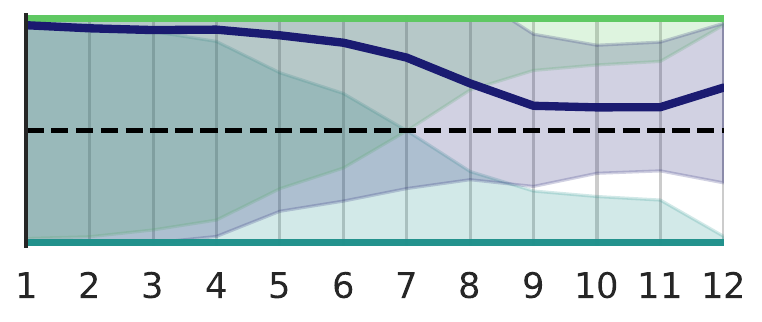}} \\ 
    \texttt{TREC} & 
    \raisebox{-0.4\totalheight}{\includegraphics[width=0.15\textwidth]{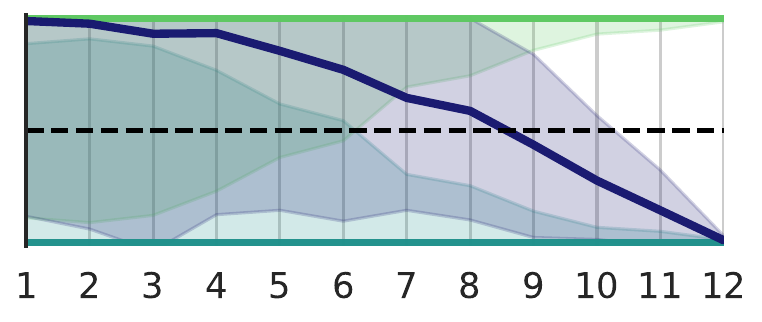}} &
    \raisebox{-0.4\totalheight}{\includegraphics[width=0.15\textwidth]{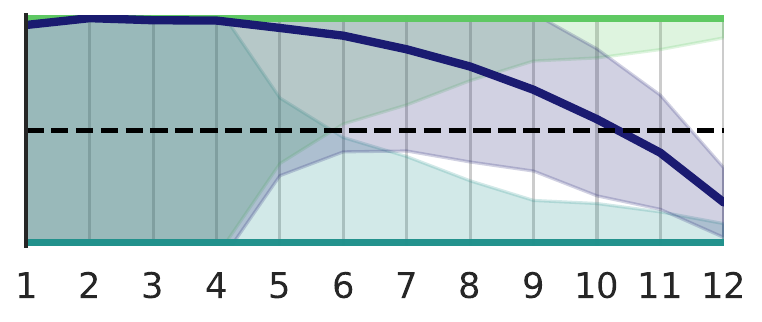}} &
    \raisebox{-0.4\totalheight}{\includegraphics[width=0.15\textwidth]{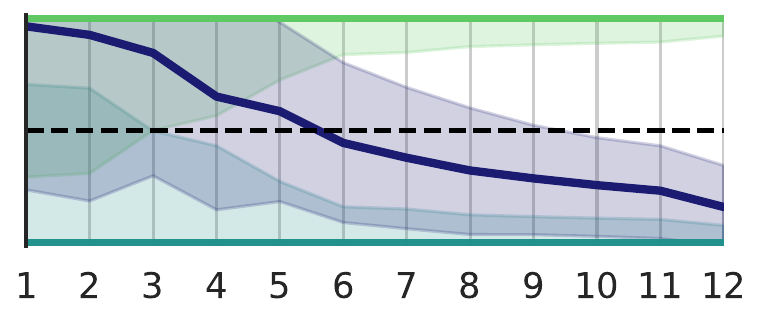}} &
    \raisebox{-0.4\totalheight}{\includegraphics[width=0.15\textwidth]{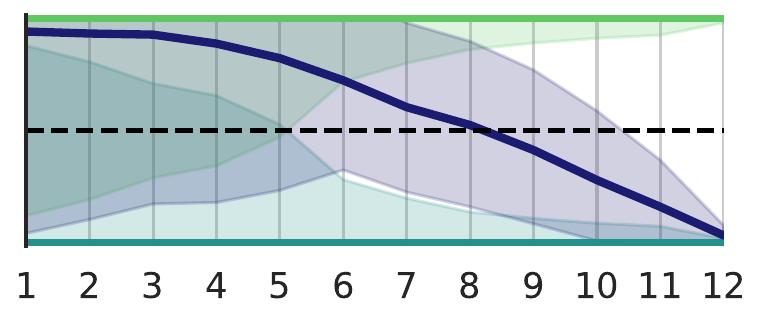}} &
    \raisebox{-0.4\totalheight}{\includegraphics[width=0.15\textwidth]{appendix_figures/centroid_analysis/trec_OPT_mixb.pdf}} &
    \raisebox{-0.4\totalheight}{\includegraphics[width=0.15\textwidth]{appendix_figures/centroid_analysis/trec_OPT_mixt.pdf}} \\ 
\end{tabular}
\captionof{figure}{Visualisations of the centroid analyses for all four models, including additional visualisations for \texttt{OPT} where the model either is assigned a `regular' bottom 6 layers, or a `regular' top 6 layers (regular meaning the layers come from $\theta_{O}$, trained on the original labels).}
\label{fig:all_centroids}
\end{table}

\newpage

\section{Increasing model size}
\label{ap:opt_big}
In the main paper, we have discussed results for four 12-layer architectures, observing generally strong agreement across those models, in spite of the fact that they were trained with varying corpora and for varying numbers of updates. To examine to what extent the results observed were specific to \textit{12}-layer architectures, we apply layer swapping to the 1.3B variant of \texttt{OPT}, containing 24 layers and ten times the number of parameters of the other models we considered.

\begin{figure}[!htb]
\minipage{0.7\textwidth}
    \begin{subfigure}[b]{0.32\textwidth}\includegraphics[width=\textwidth]{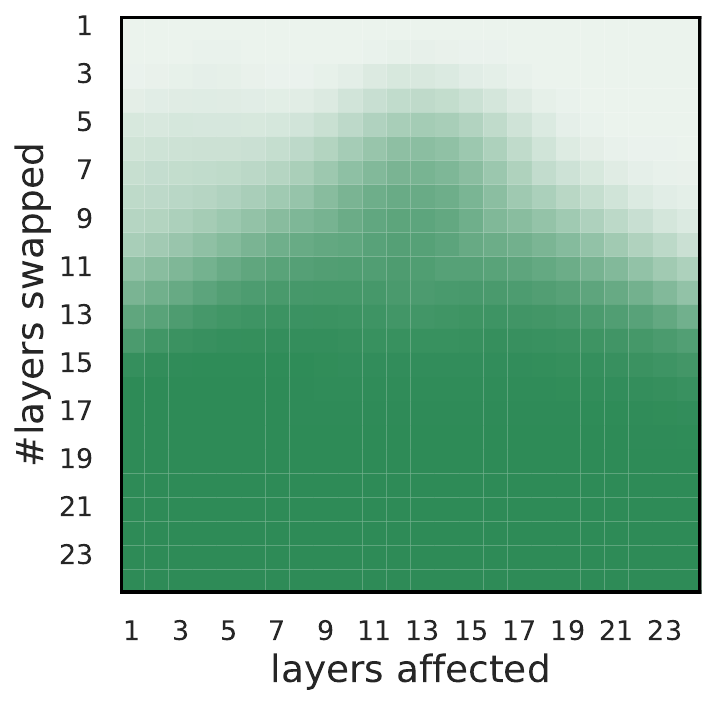}\caption{\wicb}\end{subfigure}
    \begin{subfigure}[b]{0.32\textwidth}\includegraphics[width=\textwidth]{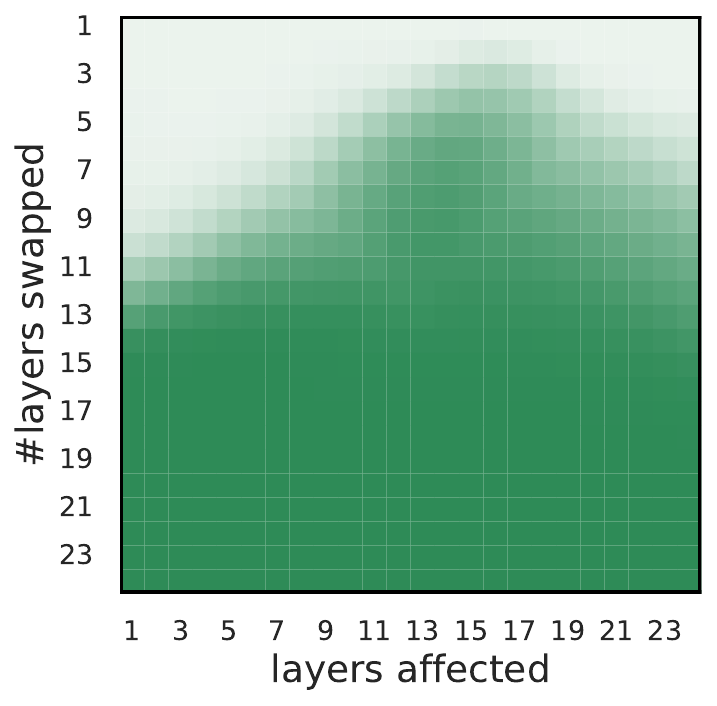}\caption{\ssttb}\end{subfigure}
    \begin{subfigure}[b]{0.32\textwidth}\includegraphics[width=\textwidth]{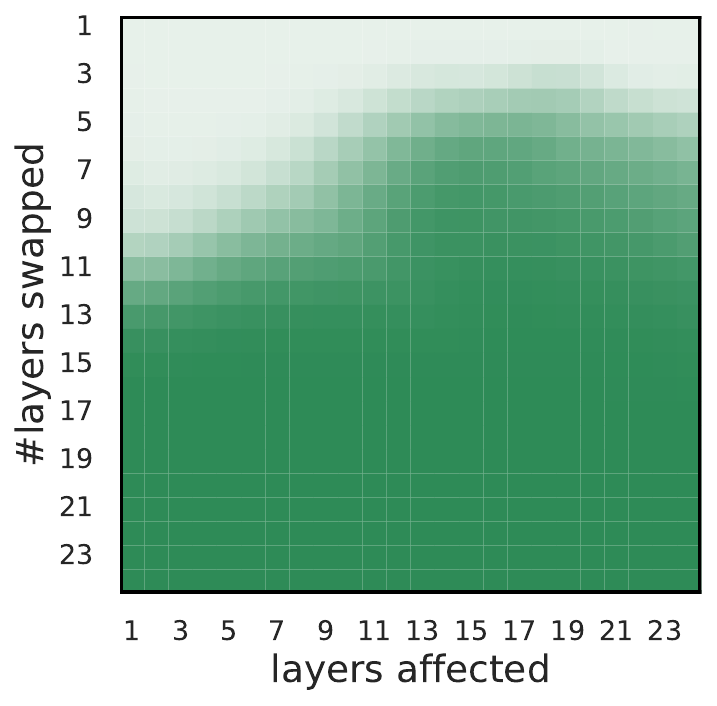}\caption{\trecb}\end{subfigure}
    \caption{Layer swapping results for three datasets, for \texttt{OPT-1.3B} containing 24 layers. The graphs show the error rate for noisy examples, that goes from 0\% when swapping only 1 layer to 100\% when swapping all layers.}
    \label{fig:swapping_1.3b}
\endminipage\hfill
\minipage{0.25\textwidth}
    \includegraphics[width=\textwidth]{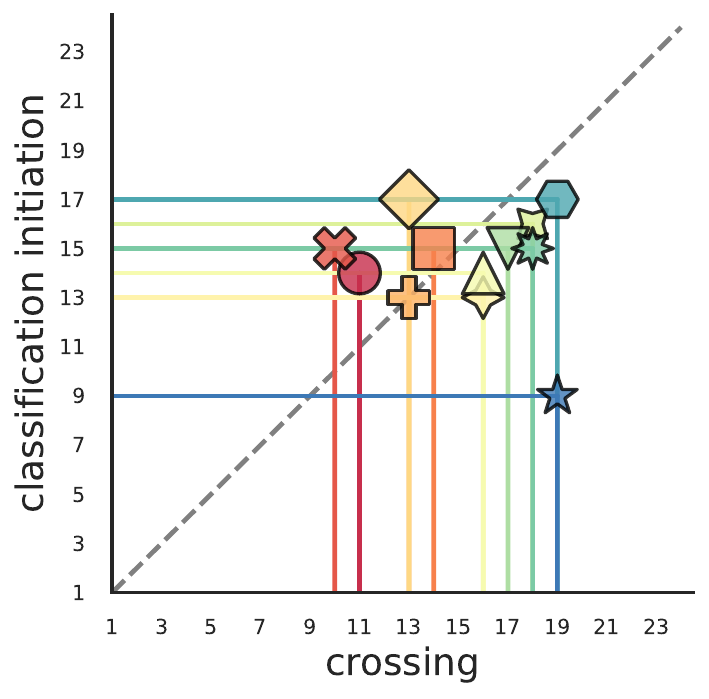}
  \caption{Summary of the memorisation and classification onsets for \texttt{OPT-1.3B}.}\label{fig:events_big}
\endminipage
\end{figure}

\begin{figure}[!htb]
    \begin{subfigure}[b]{0.24\textwidth}
        \includegraphics[width=\textwidth]{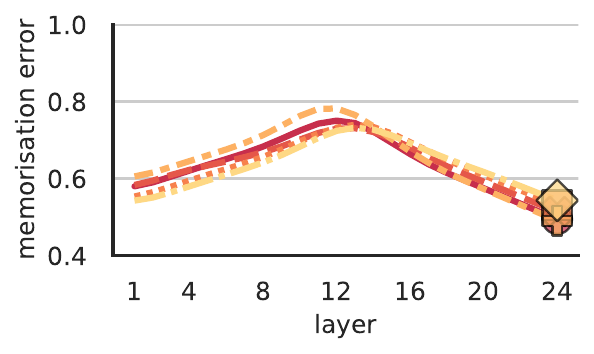}
        \caption{NLU tasks}
    \end{subfigure}
    \begin{subfigure}[b]{0.24\textwidth}
        \includegraphics[width=\textwidth]{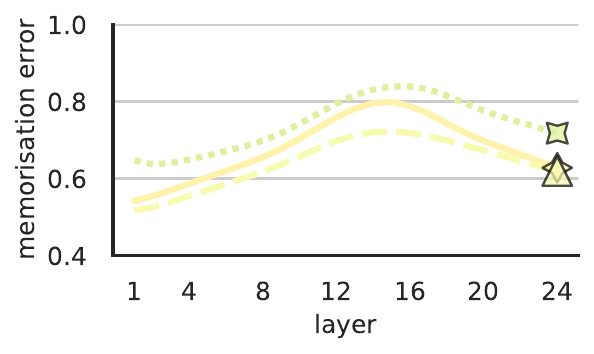}
        \caption{Sentiment tasks}
    \end{subfigure}
    \begin{subfigure}[b]{0.24\textwidth}
        \includegraphics[width=\textwidth]{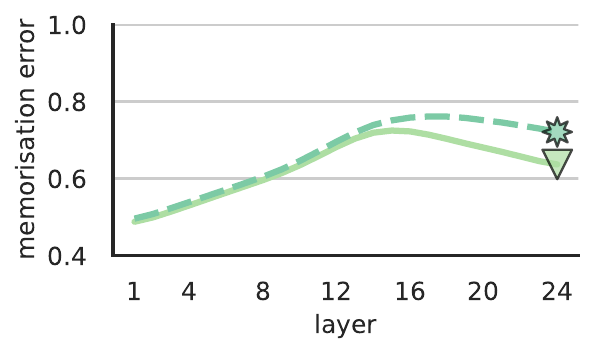}
        \caption{Hate speech tasks}
    \end{subfigure}
    \begin{subfigure}[b]{0.24\textwidth}
        \includegraphics[width=\textwidth]{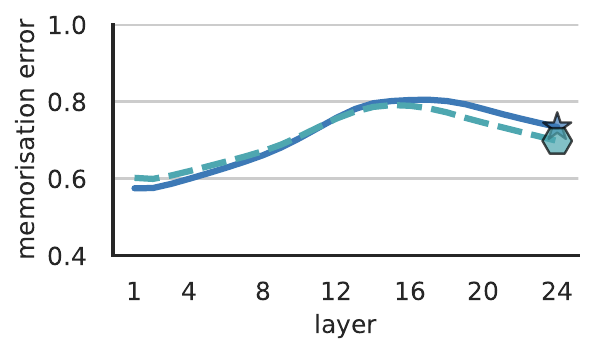}
        \caption{Topic classification}
    \end{subfigure}
    \caption{Per-layer memorisation error rate, averaged over all window sizes during layer swapping for \texttt{OPT-1.3B}. A higher error rate suggests higher relevance for memorisation.}
    \label{fig:swapping_1.3b-2}
    \vspace{-0.5cm}
\end{figure}

Figure~\ref{fig:swapping_1.3b} firstly provides three example matrices, similar to the ones discussed in \S\ref{sec:swapping_retraining}. For all three datasets shown, swapping the middle layers most effectively reverts memorisation when considering the smaller window sizes, but there are clear distinctions between the three datasets, too: for \wicb\ only the middle layers appear most relevant, whereas for \ssttb\ and \trecb\ the upper layers are more relevant than for \wicb. Figure~\ref{fig:swapping_1.3b-2} averages the rows from the matrices to summarise results across the 12 datasets, displaying a pattern similar to the main paper, with NLU tasks relying more heavily on (relatively speaking) lower layers than the remaining tasks. The agreement is also reflected in Spearman's $\rho$ between the M-CoG coefficients from the main paper for layer swapping and the M-CoG coefficients computed using Figure~\ref{fig:swapping_1.3b-2}: $\rho=0.73$ for \texttt{Pythia}, $\rho=0.84$ for \texttt{GPT-N}, $\rho=0.75$ for \texttt{BERT} and $\rho=0.87$ for \texttt{OPT} (small).
At the same time, there is a difference to the results discussed in the main paper since the lowest layers (in absolute terms) appear much less relevant.

When we execute the centroid analysis and summarise the results using the crossing and classification initiation events (Figure~\ref{fig:events_big}), we similarly observe that the crossings correlate very strongly with the crossings from the four models ($\rho=0.94$), although the classification initiations correlate very weakly with \texttt{OPT-1.3B} ($\rho=0.15$, but $p>0.05$).

\newpage
\section{Binarisation of tasks}
\label{ap:binarised_tasks}

In the main paper, we used 12 varied NLP classification tasks in our memorisation localisation endeavours. Having identified that tasks differ in terms of the layers that matter most for memorisation, we should also note that the tasks with the highest M-CoG coefficients and the highest crossings in \S\ref{sec:centroid_analysis} also happen to be the tasks that do not have a binary label set -- e.g.\ consider Figure~\ref{fig:events_averaged_centroid}, where among the six highest crossings, there are five from multi-class tasks. To ensure that the effect observed is not specific to tasks with a large label set size, we now change the multi-class tasks (\sstfb, \emotionb, \ihb, \trecb, \reutersb) into binary classification and repeat layer swapping and the centroid analysis. We do this by taking the most frequent two classes for a task, and training models again with 15\% of the labels perturbed, using one model seed only. We now compare these models to the same model seed trained on the multi-class variant of the same tasks.

\begin{figure}[!htb]
\minipage{0.42\textwidth}
    \begin{subfigure}[b]{\textwidth}\includegraphics[width=\textwidth]{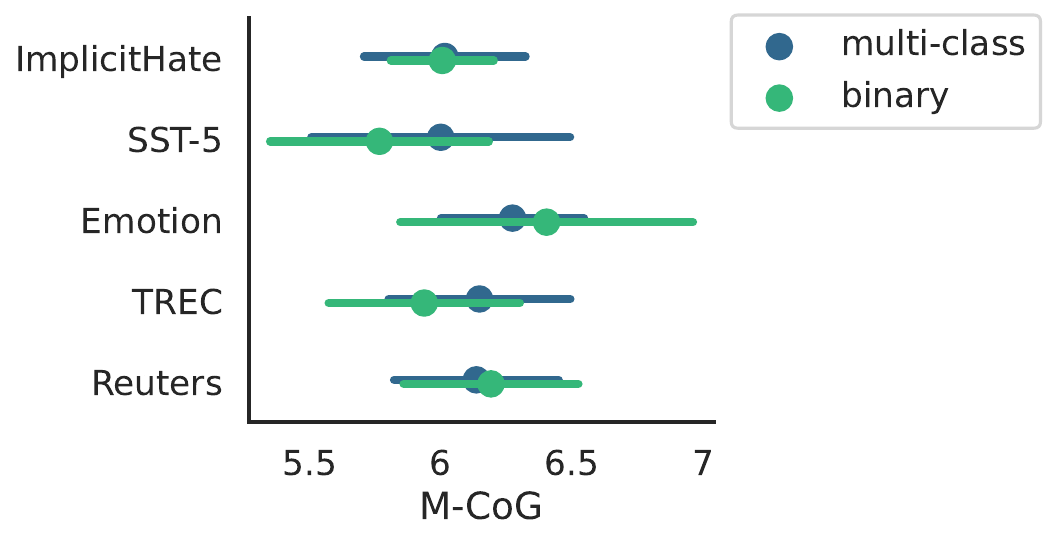}\end{subfigure}
    \caption{M-CoG coefficients for layer swapping, comparing multi-class to binarised tasks. Error bars show standard deviations over models.}
    \label{fig:binarised_mcog}
\endminipage\hfill
\minipage{0.47\textwidth}
    \begin{subfigure}[b]{0.45\textwidth}\includegraphics[width=\textwidth]{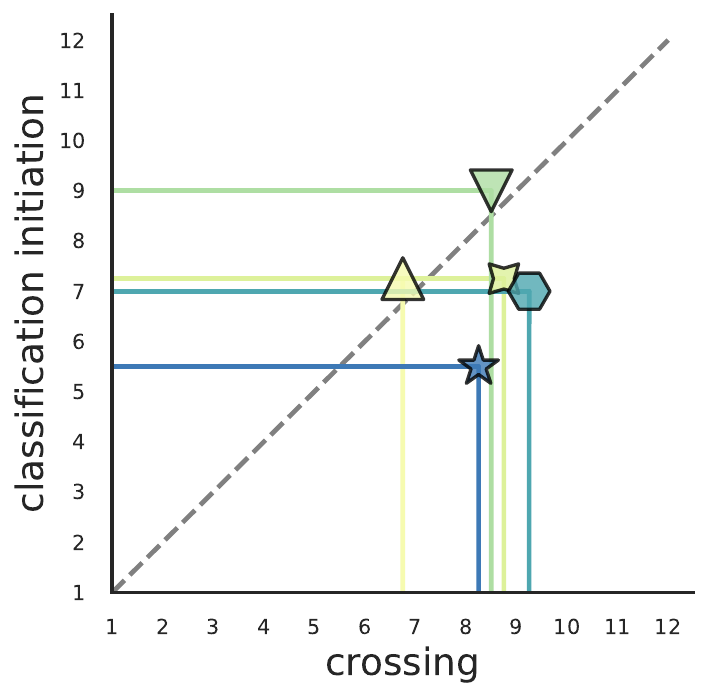}\caption{Multi-class tasks}\end{subfigure}
    \begin{subfigure}[b]{0.45\textwidth}\includegraphics[width=\textwidth]{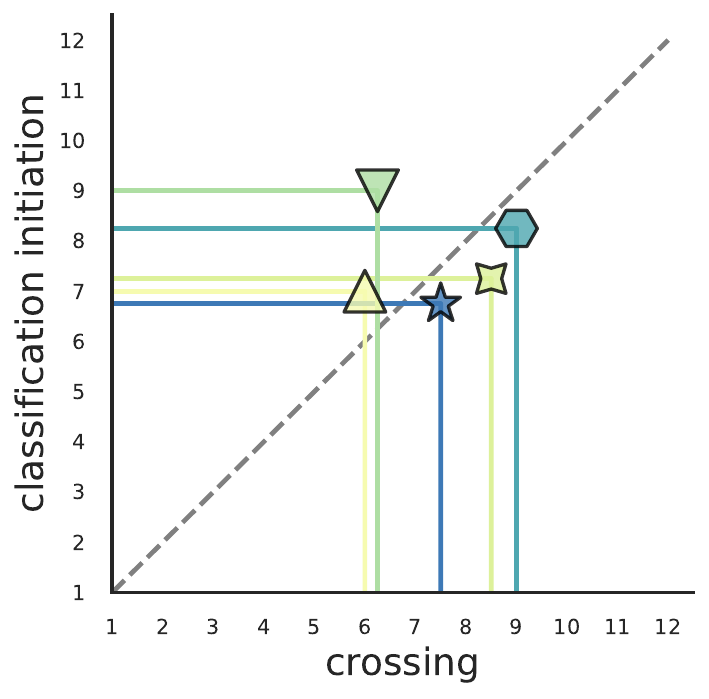}\caption{Binarised tasks}\end{subfigure}
    \caption{Summary of the memorisation and classification onsets for the binarised multi-class tasks.}
    \label{fig:binarised_events}
\endminipage
\end{figure}

For layer swapping, the M-CoG of the multi-class and binary setups correlate with Spearman's $\rho=0.84$, combining data points from all four models (see Figure~\ref{fig:binarised_mcog}); those same coefficients have a mean difference of -0.05 and a mean absolute difference of 0.16, meaning that overall, the coefficients differ only slightly.

When we repeat the centroid analysis and compute the crossing and classification initiation events, those similarly correlate strongly before and after binarisation ($\rho=0.90$ with $p<0.05$ for the crossing and $\rho=0.67$ with $p>0.05$ for the classification initiation). Figure~\ref{fig:binarised_events} shows the events when averaged over models. And when we look at the absolute numbers obtained for these two events, the crossing is an average of 0.85 layers earlier, and the initiation is an average of 0.45 layers later, meaning that although the binarised tasks yield slightly different results, they still starkly differ from the results obtained for the group of NLU tasks.

\clearpage
\section{Technical setup and model/data details}
\label{ap:setup}

\paragraph{Technical setup} We ran the experiments for the 12-layer models on NVIDIA GeForce RTX 1080/2080 Ti GPUs.
We train the small models using a batch size of 8 (due to GPU restrictions, or 4 in the few cases where we still get memory errors, which happens for \reutersb, in particular) and an initial learning rate of 1e-5 for 50 epochs, capping sequences at 512 tokens. 50 epochs is beyond the point of convergence since the aim is to investigate memorisation rather than optimise models for their generalisation capabilities.
For the models from \S\ref{sec:control_setup} where the main task can only modify two layers at a time, we rerun training with an increased learning rate if the training accuracy does not exceed .99.
For every model trained, we store checkpoint $\theta_{M_1}$ when the training accuracy exceeds .993, and store checkpoint $\theta_{M_2}$ at the end of training.
The most time-consuming experiments are model training and layer retraining:
\begin{itemize}
    \item \S\ref{sec:control_setup}: 11 datasets $\times$ 3 control setups to obtain $\theta_M$ + 11 datasets $\times$ 3 control setups to obtain $\theta_O$ + 11 datasets $\times$ 1 frozen model = 77 setups trained for each of the 4 models, taking 1 - 6 hours each
    \item \S\ref{sec:results}: 12 datasets $\times$ 3 seeds for $\theta_M$ + 12 datasets $\times$ 3 seeds for $\theta_O$ + 12 datasets $\times$ 1 frozen model  = 84 setups trained for each of the 4 models, taking 1 - 6 hours each \\ Layer retraining: 12 datasets $\times$ 3 seeds $\theta_M$ $\times$ 78 windows = 2808 setups trained for each of the 4 models, taking 3 to 45 minutes each
\end{itemize}

The experiments discussed in Appendix \ref{ap:opt_big} are ran on NVIDIA A100-SXM80GB GPUs. \texttt{OPT-1.3B} is trained with an initial learning rate of $5e-6$ and a batch size of 32 or 16. We train two models per dataset ($\theta_M$ and $\theta_O$), and individual training runs take 45 minutes to 6 hours, depending on the dataset. Visit our codebase here: \url{https://github.com/vernadankers/memorisation_localisation}.

We use the \texttt{transformers} library\footnote{\url{https://huggingface.co/docs/transformers}} to obtain the models/tokenisers and train them, implementing the remaining analyses ourselves.

\paragraph{Model licenses}
The licenses of all models, which are \texttt{Apache 2.0} (\texttt{BERT}), a custom license for \texttt{OPT} models\footnote{\url{https://github.com/facebookresearch/metaseq/blob/main/projects/OPT/MODEL_LICENSE.md}} and the MIT License (\texttt{Pythia}, \texttt{GPT-N}) allow non-commercial use for research purposes.

\paragraph{Dataset licenses}
The datasets contained in GLUE and SuperGlue are available under licences that allow use and redistribution for research purposes \citep{wang2018glue,wang2019superglue}. \stormfrontb\ is available under \texttt{CC-by-SA-3.0}; \ihb\ is not explicitly assigned a license, but the corresponding repository is available under the \texttt{MIT} license; \reutersb\ is available under the \texttt{CC-BY-4.0} license; for \texttt{TREC} the license is unknown, and \emotionb\ should be used for educational and research purposes only, and has no license, otherwise\footnote{\url{https://github.com/dair-ai/emotion_dataset}}.

\begin{table}[!h]\small\centering
    \begin{tabular}{llccccc}
        \toprule
        \textbf{Model}               & \textbf{Corpus}            & \textbf{Tokens} & \textbf{Steps} & \textbf{Params} & \textbf{Layers} & \textbf{Model dim} \\\midrule
        \href{https://huggingface.co/bert-base-cased}{\texttt{BERT-base}} & BooksCorpus, Wikipedia     & 3.3B & 1M           & 85M & 12 & 768 \\
        \href{https://huggingface.co/EleutherAI/pythia-160m-deduped}{\texttt{Pythia-160m}} & The Pile                   & 300B & 143k         & 85M & 12 & 768\\
        \href{https://huggingface.co/EleutherAI/gpt-neo-125m}{\texttt{GPT-Neo-125m}}        & The Pile                   & 300B & 572k          & 85M    & 12 & 768\\
        \href{https://huggingface.co/facebook/opt-125m}{\texttt{OPT-125m}} & BookCorpus, CC-Stories,    & 180B & ?        & 85M    & 12 & 768\\
                                                                          & The Pile, Reddit, CCNewsV2 & \\
        \href{https://huggingface.co/facebook/opt-1.3b}{\texttt{OPT-1.3B}} & idem    & 180B & ?       & 1.2B    & 24 & 2048\\

        \bottomrule
    \end{tabular}
    \caption{Overview of models, along with their pre-training corpora, the number of tokens the model has seen during training, the number of training steps, the number of non-embedding parameters, and the number of layers and hidden dimensionality.}
    \label{tab:my_label}
\end{table}


\section{Postprocessing gradients}
\label{ap:hypestimation}
As described in \S\ref{sec:methods}, forgetting gradients are one of the signals we examine to perform memorisation localisation. We average them over all noisy examples, or over a similar amount of clean examples. Preliminary experiments indicated that, taken at face value, the gradients do not necessarily pinpoint the correct layers in a control setup. Using two validation tasks (\texttt{MRPC} and \texttt{TREC}), we consider taking the $L_1$-norm or the $L_2$-norm over gradients and applying two ways of normalising the forgetting gradients of the noisy examples: i) subtract the forgetting gradients of clean examples, ii) normalise the per-layer norm by the norm obtained using a frozen model. The final post-processing step applied afterwards is that the weights of the 12 layers are normalised to sum to 1 to allow for the computation of the M-CoG coefficients, and to reduce variation among tasks.

\begin{figure}[t]\centering
\begin{subfigure}[b]{0.24\textwidth}
    \centering
    \includegraphics[width=\textwidth]{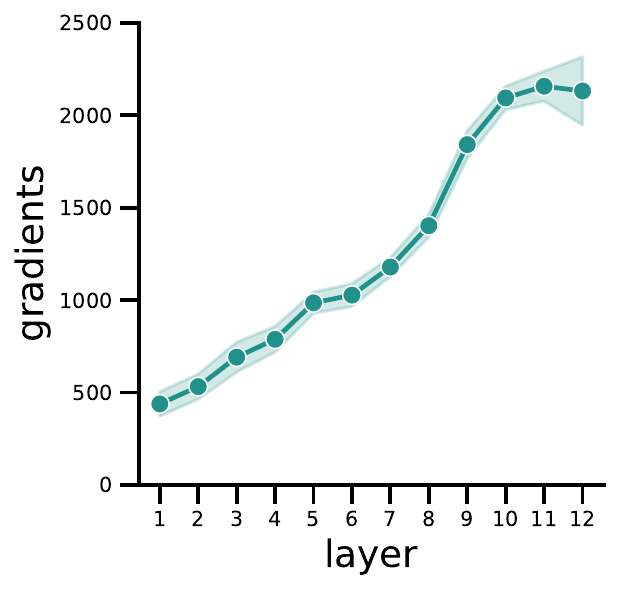}
    \caption{Frozen model gradients}
    \label{fig:validate_gradients_frozen}
\end{subfigure}
\begin{subfigure}[b]{0.24\textwidth}
    \centering
    \includegraphics[width=\textwidth]{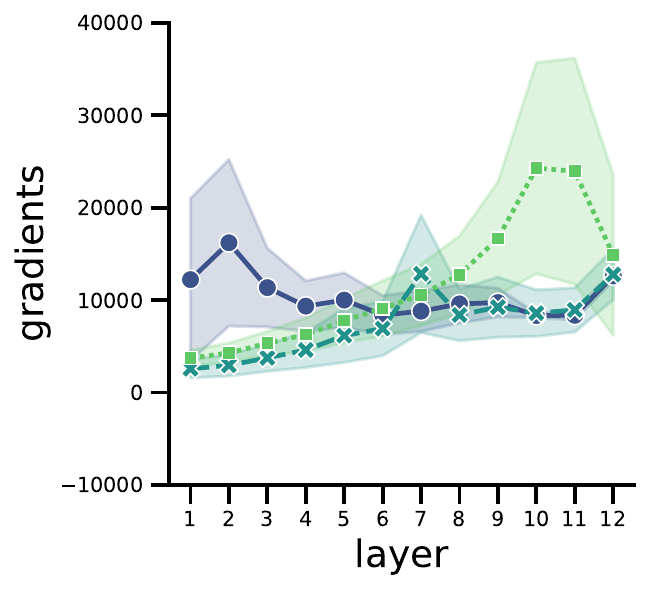}
    \caption{`Clean' gradients}
    \label{fig:validate_gradients_clean}
\end{subfigure}
\begin{subfigure}[b]{0.24\textwidth}
    \centering
    \includegraphics[width=\textwidth]{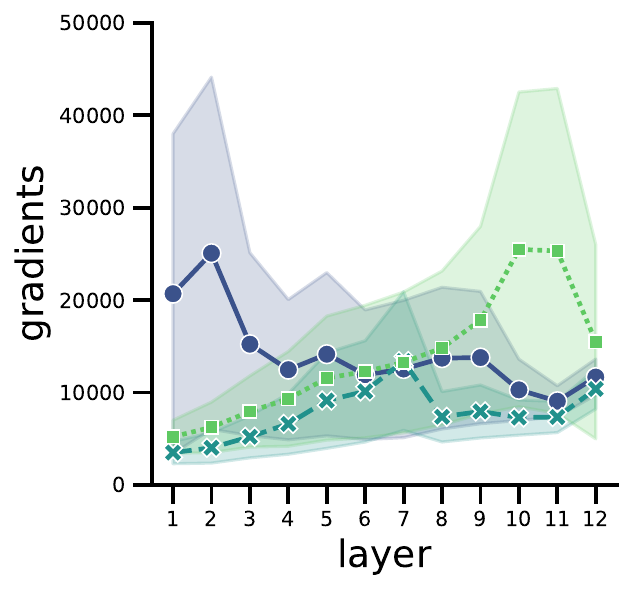}
    \caption{`Noisy' gradients}
    \label{fig:validate_gradients_noisy}
\end{subfigure}

\begin{subfigure}[b]{0.24\textwidth}
    \centering
    \includegraphics[width=\textwidth]{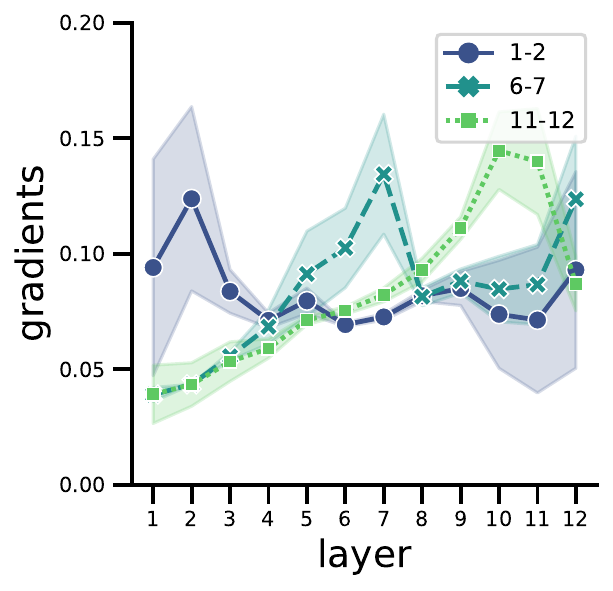}
    \caption{`Noisy' gradients, norm.}
    \label{fig:validate_gradients_a_bert}
\end{subfigure}
\begin{subfigure}[b]{0.24\textwidth}
    \centering
    \includegraphics[width=\textwidth]{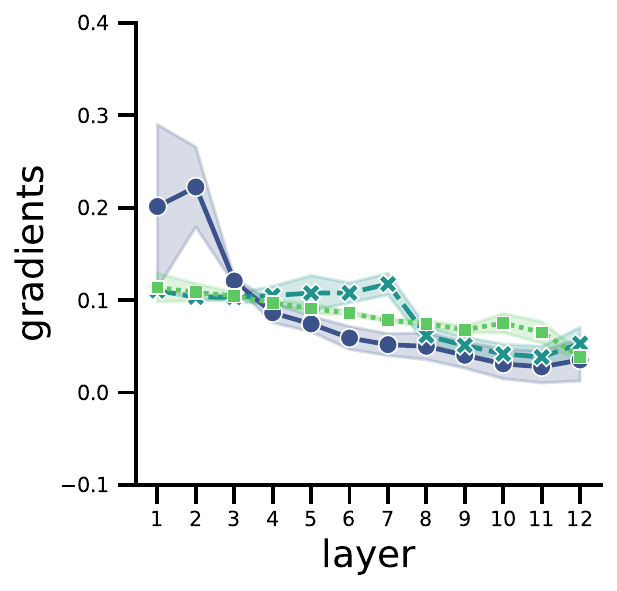}
    \caption{Divide by frozen, norm.}
    \label{fig:validate_gradients_b_bert}
\end{subfigure}
\begin{subfigure}[b]{0.24\textwidth}
    \centering
    \includegraphics[width=\textwidth]{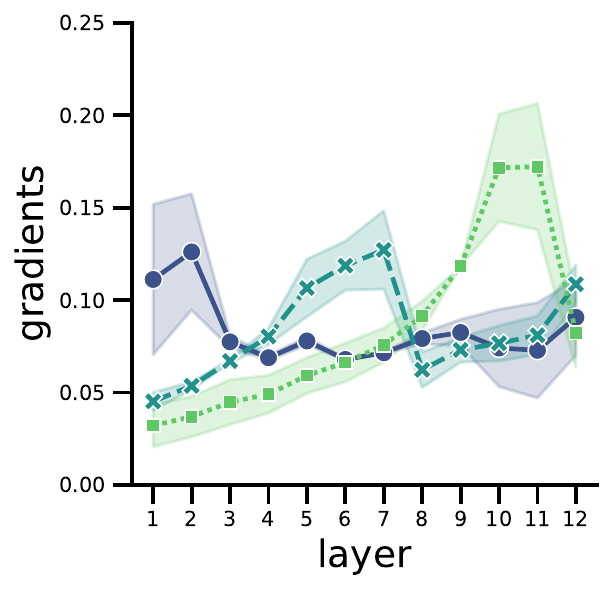}
    \caption{Subtract clean, norm.}
    \label{fig:validate_gradients_c_bert}
\end{subfigure}
\begin{subfigure}[b]{0.24\textwidth}
    \centering
    \includegraphics[width=\textwidth]{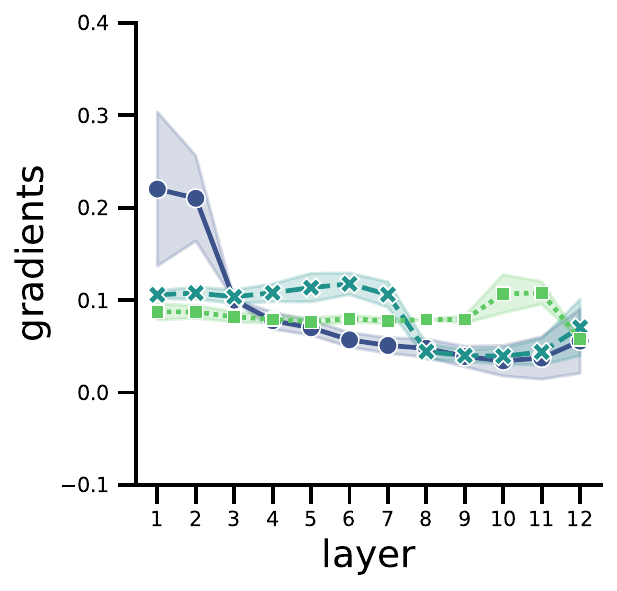}
    \caption{Divide, subtract, norm.}
    \label{fig:validate_gradients_d_bert}
\end{subfigure}    
\caption{Effect of the gradient analysis postprocessing steps on the \mrpcb\ and \trecb\ tasks for the \texttt{BERT} model when using the $L_1$-norm.}
\label{fig:validate_gradients}
\end{figure}

Figure~\ref{fig:validate_gradients_frozen} illustrates the $L_1$-norm for `forgetting' gradients for a frozen \texttt{BERT}, that tend to point to the final layers; Figure~\ref{fig:validate_gradients_clean} and Figure~\ref{fig:validate_gradients_noisy} demonstrate forgetting gradients for clean and noisy examples in the control setup. Both point to similar layers, but the norms are higher for noisy examples.

Figure~\ref{fig:validate_gradients_a_bert}-\ref{fig:validate_gradients_d_bert} \textit{do} apply the within-dataset normalisation that normalises layer weights to sum to one. Figure~\ref{fig:validate_gradients_a_bert} again demonstrates for noisy examples that without further post-processing, the gradients overestimate the relevance of later layers in \texttt{BERT}. Both post-processing steps i) and ii) dampen that.
When measuring the success of the post-processing steps using the accuracy metric, included in Table~\ref{tab:validate_gradients}, the combination of both is most successful at recovering the layers in which memorisation had taken place in the control setups, and the $L_1$-norm leads to more accurate results than the $L_2$-norm.

These post-processing steps improve the accuracy for all models but \texttt{Pythia}. Across the board, applying both steps i) and ii) and using the $L_1$-norm yields the highest accuracy, so we apply both of these steps in the main paper.

\begin{table}[!h]
    \centering\small
    \begin{tabular}{llcccccccc}
    \toprule
    & & \multicolumn{2}{c}{\texttt{Pythia}} & \multicolumn{2}{c}{\texttt{GPT-N}} & \multicolumn{2}{c}{\texttt{BERT}} & \multicolumn{2}{c}{\texttt{OPT}} \\
    \textbf{subtracing clean} & \textbf{normalising frozen} &  $L_1$ & $L_2$ &  $L_1$ & $L_2$ &  $L_1$ & $L_2$   &  $L_1$ & $L_2$  \\
    \midrule
    $\times$ & $\times$         & 0.08 & 0.08 & 0.58 & 0.25 & 0.58 & 0.50 & 0.58 & 0.17 \\
    $\times$ & $\checkmark$     & 0.08 & 0.08 & 0.67 & 0.42 & 0.50 & 0.50 & 0.50 & 0.42 \\
    $\checkmark$ & $\times$     & 0.08 & 0.08 & 0.58 & 0.25 & 0.58 & 0.50 & 0.67 & 0.33 \\
    $\checkmark$ & $\checkmark$ & 0.08 & 0.00 & 0.75 & 0.25 & 0.75 & 0.58 & 0.58 & 0.50 \\
    \bottomrule
    \end{tabular}
    \caption{Effect of the gradient analysis postprocessing steps on the \mrpcb\ and \trecb\ tasks, measured as the average accuracy of the highest scoring layers.}
    \label{tab:validate_gradients}
\end{table}

\end{document}